\documentclass[]{article}

\usepackage{pdfpages}
\usepackage{algorithm}
\usepackage{algpseudocode}
\usepackage{url}
\usepackage{graphicx}
\usepackage{stmaryrd}

\usepackage{amssymb}
\usepackage{xspace}
\usepackage{color}
\usepackage{subcaption}
\usepackage{graphicx}
\usepackage{amssymb}

\usepackage{amsmath,mathtools}

\usepackage{tabu}
\usepackage{tabularx}
\usepackage{booktabs}
\usepackage{adjustbox}
\usepackage{array}
\newcommand{\dashedline}{\tabucline[0.3pt on 3pt]\\}
\usepackage{xcolor}
\usepackage{color,colortbl}
\usepackage{arydshln}

\usepackage{mathabx}
\usepackage{multirow}
\usepackage{soul}

\usepackage{listings}
\usepackage{relsize}
\usepackage{amsthm}
\usepackage{bm}

\newcommand{\vDf}[1]{\ensuremath{\bm{#1}}}

\newcommand{\vDvx}{\vDf{x}}

\newcommand{\vvx}{\vDvx}

\newcommand{\vp}[1]{\ensuremath{( #1 )}}
\newcommand{\vt}[1]{\ensuremath{#1^{\mathrm{T}}}}

\newcommand{\dd}{\ensuremath{\mathrm{d}}}

\newcommand{\ev}{\ensuremath{\mathbb{E}}}
\newcommand{\vv}{\ensuremath{\mathrm{var}}}
\newcommand{\cv}{\ensuremath{\mathrm{cov}}}

\newcommand{\Data}{\ensuremath{\mathcal{D}}}

\newcommand{\dbern}{\ensuremath{\mathrm{Bern}}}

\newcommand{\dbeta}{\ensuremath{\mathrm{Beta}}}

\usepackage{colonequals}

\newcommand{\tuple}[1]{\ensuremath{\langle #1 \rangle}}
\usepackage{bm}

\newcommand{\slop}[1]{\ensuremath{\omega_{#1}}}
\newcommand{\slbel}[1]{\ensuremath{b_{#1}}}
\newcommand{\sldis}[1]{\ensuremath{d_{#1}}}
\newcommand{\slunc}[1]{\ensuremath{u_{#1}}}
\newcommand{\slbase}[1]{\ensuremath{a_{#1}}}
\newcommand{\sloptuple}[1]{\ensuremath{\tuple{\slbel{#1}, \sldis{#1}, \slunc{#1}, \slbase{#1}}}}
\newcommand{\variance}[1]{\ensuremath{\sigma^2_{#1}}}
\newcommand{\mean}[1]{\ensuremath{\ev[#1]}}
\newcommand{\var}[1]{\vv[#1]}

\newcommand{\semiring}{\ensuremath{\mathcal{S}}}
\newcommand{\semiringp}{\ensuremath{\semiring_p}}
\newcommand{\semiringsl}{\ensuremath{\semiring_{\text{SL}}}}
\newcommand{\semiringbeta}{\ensuremath{\semiring^{\beta}}}

\definecolor{cBetaProblog}{HTML}{0000FF}
\definecolor{cHonolulu}{HTML}{000000}
\definecolor{cJosang}{HTML}{FF0000}
\definecolor{cSBN}{HTML}{BF00BF}
\definecolor{cGBT}{HTML}{008000}
\definecolor{cCredal}{HTML}{BFBF00}
\definecolor{cMonteCarlo}{HTML}{B35F00}

\newcommand{\mtt}[1]{\ensuremath{\texttt{#1}}}

\newcommand{\netone}{Net1}
\newcommand{\nettwo}{Net2}
\newcommand{\netthree}{Net3}
\newcommand{\best}[1]{\fbox{#1}}

\newcommand{\amccond}{\ensuremath{\text{}}}

\newcommand{\phspace}{}%\vphantom{$\semiringbeta\semiringsl$}}
\newcommand{\lbetaprobcov}{\ensuremath{\color{cBetaProblog}\fbox{CPB\phspace}}}
\newcommand{\lBetaProblog}{\lbetaprobcov}
\newcommand{\lHonolulu}{\ensuremath{\color{cHonolulu}\fbox{\amccond \semiringbeta\phspace}}}
\newcommand{\lJosang}{\ensuremath{\color{cJosang}\fbox{\amccond \semiringsl\phspace}}}
\newcommand{\lSBN}{\ensuremath{\color{cSBN}\fbox{SBN\phspace}}}
\newcommand{\lGBT}{\ensuremath{\color{cGBT}\fbox{GBT\phspace}}}
\newcommand{\lCredal}{\ensuremath{\color{cCredal}\fbox{Credal\phspace}}}
\newcommand{\lMonteCarlo}{\ensuremath{\color{cMonteCarlo}\fbox{MC\phspace}}}

\usepackage{bm}

\newfloat{Algorithm}{thpb}{lop}
\floatname{Algorithm}{Algorithm}

\usepackage{todonotes}
\presetkeys%
    {todonotes}%
    {inline,backgroundcolor=gray}{}

\newcommand{\NNF}{{\tt NNF}}

\newcommand{\dDNNF}{{\tt d-DNNF}}

\newcommand{\labelfun}{\ensuremath{\rho}}

\newcommand{\set}[1]{\ensuremath{\{#1\}}}

\newcommand{\prior}[1]{\ensuremath{#1^{0}}}
\usepackage{siunitx}

\definecolor{shadow}{HTML}{DDDDDD}
\newcommand{\noderv}[1]{\ensuremath{X_{#1}}}

\newcommand{\sh}[1]{\ensuremath{\widehat{#1}}}
\usepackage[]{authblk}

\begin{document}

\title{Handling Epistemic and Aleatory Uncertainties in Probabilistic Circuits}

\author[1]{Federico Cerutti}
\author[2]{Lance M. Kaplan}
\author[3]{Angelika Kimmig}
\author[4]{Murat \c{S}ensoy}

\affil[1]{Department of Information Engineering, Universit\`a degli Studi di Brescia, Brescia, Italy\\
            Cardiff University, Crime and Security Research Institute, Cardiff, UK\\
            \texttt{federico.cerutti@unibs.it}}

\affil[2]{CCDC Army Research Laboratory, Adelphi, MD, USA\\
            \texttt{lance.m.kaplan.civ@mail.mil}}
\affil[3]{Department of Computer Science, KU Leuven, Belgium\\
            \texttt{angelika.kimmig@kuleuven.be}}
\affil[4]{Blue Prism AI Labs, London, UK
            \\
            Department of Computer Science, Ozyegin University, Istanbul, Turkey\\
            \texttt{murat.sensoy@ozyegin.edu.tr}}

\date{Submitted to MACH: Under Review}
          
\maketitle

\begin{abstract}
When collaborating with an AI system, we need to assess when to trust its recommendations. If we mistakenly trust it in regions where it is likely to err, catastrophic failures may occur, hence the need for Bayesian approaches for probabilistic reasoning in order to determine the confidence (or epistemic uncertainty) in the probabilities in light of the training data.
We propose an approach to overcome the independence assumption behind most of the approaches dealing with a large class of probabilistic reasoning that includes Bayesian networks as well as several instances of probabilistic logic. We  provide an algorithm for Bayesian learning from sparse, albeit complete, observations, and for deriving inferences and their confidences keeping track of the dependencies between variables when they are manipulated within the unifying computational formalism provided by probabilistic circuits. Each leaf of such circuits is labelled with a beta-distributed random variable that provides us with an elegant framework for representing uncertain probabilities.
We achieve better estimation of epistemic uncertainty than state-of-the-art approaches, including highly engineered ones, while being able to handle general circuits and with just a modest increase in the computational effort compared to using point probabilities.
\end{abstract}

\section{Introduction}
Even in simple collaboration scenarios---like those in which an artificial intelligence (AI) system assists a human operator with predictions---the success of the team hinges on the human correctly deciding when to follow the recommendations of the AI system and when to override them \cite{DBLP:conf/aaai/BansalNKWLH19}. 
Extracting benefits from collaboration with the AI system depends on the human developing insights (i.e., a mental model) of when to trust the AI system with its recommendations \cite{DBLP:conf/aaai/BansalNKWLH19}. If the human mistakenly trusts the AI system in regions where it is likely to err, catastrophic failures may occur. This is a strong argument in favour of Bayesian approaches to probabilistic reasoning: research in the intersection of AI and HCI has found that interaction improves when setting expectations right about what the system can do and how well it performs \cite{Kocielnik2019,bansal2019beyond}. Guidelines have been produced \cite{Amershi2019}, and they recommend to \emph{Make clear what the system can do} (\textbf{G1}), and \emph{Make clear how well the system can do what it can
do} (\textbf{G2}).

To identify such regions where the AI system is likely to err, we need to distinguish between (at least) two different sources of uncertainty: \emph{aleatory} (or \emph{aleatoric}), and \emph{epistemic}  uncertainty \cite{HORA1996217,hllermeier2019aleatoric}. Aleatory uncertainty refers to the variability in the outcome of an experiment which is due to inherently random effects (e.g. flipping a fair coin): no additional source of information but Laplace's daemon\footnote{``An
intelligence that, at a given instant, could comprehend all the forces by
which nature is animated and the respective situation of the beings that
make it up'' \cite[p.2]{Laplace-prob}.} can reduce such a variability. Epistemic uncertainty refers to the epistemic state of the agent using the model, hence its lack of knowledge that---in principle---can be reduced on the basis of additional data samples.
Particularly when considering sparse data, the epistemic uncertainty
around the learnt model can significantly affect decision making \cite{anderson2016using,antonucci2014decision}, for instance when used for computing an expected utility \cite{von2007theory}.

In this paper, we propose an approach to probabilistic reasoning
that manipulates distributions of probabilities without assuming independence and without resorting to sampling approaches
within the unifying computational formalism provided by arithmetic circuits \cite{VonzurGathen1988}, sometimes named \emph{probabilistic circuits} when manipulating probabilities, or simply \emph{circuits}. 
This is clearly a novel contribution as the few approaches \cite{Rashwan2016,Jaini2016,DBLP:conf/aaai/CeruttiKKS19} resorting to distribution estimation via moment matching like we also propose, still assume statistical independence in particular when manipulating distributions within the circuit.
Instead, we provide an algorithm for Bayesian learning from sparse---albeit complete---observations, and for probabilistic inferences that keep track of the dependencies between variables when they are manipulated within the circuit.
In particular, we focus on the large class of approaches to probabilistic reasoning that rely upon algebraic model counting (AMC) \cite{kimmig2017-algebraic} (Section \ref{sec:amc}), which has been proven to encompass probabilistic inferences under \cite{sato:iclp95}'s semantics, thus covering not only Bayesian networks \cite{Sang2005}, but also probabilistic logic programming approaches such as ProbLog \cite{Fierens2015}, and others as discussed by \cite{DBLP:journals/ijar/CeruttiT19}.
As  AMC is defined in
terms of the set of models of a propositional logic theory, we can
exploit the results of \cite{Darwiche2020knowledge} (Section \ref{sec:circuits}) who studied the  succinctness relations between various
types of circuits and thus their applicability to model counting.
To stress the applicability of this setting, circuit
compilation techniques  \cite{Choi2013a,Darwiche2004,Oztok2015} are behind state-of-the-art algorithms for (1) exact and approximate inference in discrete probabilistic graphical
models \cite{Chavira2008,Kisa2014,Friedman2018}; 
and (2) probabilistic programs \cite{Fierens2015,Bellodi2013}. Also, learning tractable circuits is the current method of choice for discrete
density estimation \cite{Gens2013,Rooshenas2014,Vergari2019,Vergari2015,Liang2017}. Finally, \cite{Xu2018} also used circuits to enforce logical constraints on deep neural networks.

In this paper, we label each leaf of the circuit with a beta-distributed random variable (Section \ref{sec:betasl}). The beta distribution is a well-defined theoretical framework that specifies a distribution of probabilities representing all the possible values of a probability when the exact value is unknown. In this way, the expected value of a beta-distributed random variable relates to the aleatory uncertainty of the phenomenon, and the variance to the epistemic uncertainty: the higher the variance, the \emph{less certain} the machine is, thus targeting directly \cite[\textbf{G1} and \textbf{G2}]{Amershi2019}.
In previous work \cite{DBLP:conf/aaai/CeruttiKKS19} we provided operators for manipulating beta-distributed random variables under strong independence assumptions (Section \ref{sec:parametrisation}). This paper significantly extends and improves our previous approach by eliminating the independence assumption in manipulating beta-distributed random variables within a circuit.

Indeed, our main contribution (Section \ref{sec:contribution}) is an algorithm for reasoning over a circuit whose leaves are labelled with beta-distributed random variables, with the additional piece of information describing which of those are actually independent (Section \ref{sec:preprocessing}). This is the input to an algorithm that \emph{shadows} the circuit by superimposing a second circuit for computing the probability of a query conditioned on a set of pieces of evidence (Section \ref{sec:shadowing}) in a single feed forward. While this at first might seems unnecessary, it is actually essential when inspecting the main algorithm that evaluates such a shadowed circuit (Section \ref{sec:mainalgorithm}), where a covariance matrix plays an essential role by keeping track of the dependencies between random variables while they are manipulated within the circuit. We also include discussions on memory management of the covariance matrix in Section \ref{sec:memory-performance}.

We evaluate our approach against a set of competing approaches in an extensive set of experiments detailed in Section \ref{sec:experiment}, comparing against leading approaches to dealing with uncertain probabilities, notably: (1) Monte Carlo sampling; (2) our previous proposal \cite{DBLP:conf/aaai/CeruttiKKS19} taken as representative of the class of approaches using moment matching with strong independence assumptions; (3) Subjective Logic \cite{Josang2016-SL}, that provides an alternative  representation of beta distributions as well as a calculus for manipulating them applied already in a variety of domains, e.g. \cite{josang2006trust,MOGLIA2012180,Sensoy2018}; (4) Subjective Bayesian Network (SBN) on circuits derived from singly-connected Bayesian networks \cite{ivanovska.15,kaplan.16.fusion,KAPLAN2018132}, that already showed higher performance against other traditional approaches dealing with uncertain probabilities, such as (5) Dempster-Shafer Theory of Evidence \cite{DEMPSTER68,Smets2005}, and (6) replacing single probability values with closed intervals representing the possible range of probability values \cite{credal98}. 
We achieve better estimation of epistemic uncertainty than state-of-the-art approaches, including highly engineered ones for a narrow domain such as SBN, while being able to handle general circuits and with just a modest increase in the computational effort compared to using point probabilities.

\section{Background}
\label{sec:background}

\subsection{Algebraic Model Counting}
\label{sec:amc}

\cite{kimmig2017-algebraic} introduce the task of
\emph{algebraic model counting (AMC)}.  AMC generalises weighted
model counting (WMC) to the semiring setting and supports various
types of labels,  including numerical ones as used in WMC,  but also sets,
polynomials, Boolean formulae, and many more. The underlying mathematical structure is that of a
commutative semiring. 

A \emph{semiring} is a structure $(\mathcal{A}, \oplus, \otimes, 
e^{\oplus}, e^{\otimes})$, where \emph{addition}~$\oplus$ and
\emph{multiplication}~$\otimes$ are associative binary operations over
the set~$\mathcal{A}$, $\oplus$~is commutative, $\otimes$~distributes
over~$\oplus$, $e^{\oplus}\in\mathcal{A}$ is the neutral element of~$\oplus$, $e^{\otimes}\in\mathcal{A}$ that of~$\otimes$,
and for all $a\in \mathcal{A}$, $e^{\oplus}\otimes a = a \otimes
e^{\oplus} = e^{\oplus}$. In a \emph{commutative semiring}, $\otimes$~is
commutative as well.

Algebraic model counting is now defined as follows. Given:
\begin{itemize}
  \item a \emph{propositional logic theory} $T$ over a set of
    variables $\mathcal{V}$, 
  \item a \emph{commutative semiring} $(\mathcal{A},\oplus,\otimes,   e^{\oplus},e^{\otimes})$, and 
 \item a \emph{labelling function} $\labelfun : \mathcal{L} \rightarrow \mathcal{A}$, mapping literals $\mathcal{L}$ of the variables in $\mathcal{V}$ to elements of the semiring set $\mathcal{A}$, 
\end{itemize}
compute
\begin{equation} \label{eq:amc}
\operatorname{\mathbf{A}}(T)  = \bigoplus_{I\in \mathcal{M}(T)} \bigotimes_{l \in I} \labelfun(l),
\end{equation}
where $\mathcal{M}(T)$ denotes the set of models of~$T$.

Among others, AMC generalises the task of
probabilistic inference according to \cite{sato:iclp95}'s semantics (\textbf{PROB}), \cite[Thm. 1]{kimmig2017-algebraic}, \cite{goodman-1999-semiring,eisner-2002-parameter,Bacchus2009,Baras2010,DBLP:conf/aaai/KimmigBR11}.

A \emph{query} $q$ is a finite set of algebraic literals  $q\subseteq \mathcal{L}$. We denote the set of interpretations where the query is true by $\mathcal{I}(q)$,
\begin{equation}
\mathcal{I}(q) = \{I~|~I\in \mathcal{M}(T) ~\land~ q \subseteq I\}
\end{equation}
The label of query $q$ is defined as the label of $\mathcal{I}(q)$,
\begin{equation}
\operatorname{\mathbf{A}}(q)  =  \operatorname{\mathbf{A}}(\mathcal{I}(q)) = \bigoplus_{I\in \mathcal{I}(q)}\bigotimes_{l\in I}\rho(l).\label{eq:q_int}
\end{equation}
As both operators are commutative and associative, the label is independent of the order of both literals and interpretations.

In the context of this paper, we extend AMC for handling \textbf{PROB} of queries with evidence by introducing an additional division operator $\oslash$ that defines the conditional label of a query as follows:
\begin{equation}
\label{eq:fusion}
\displaystyle{\mathbf{A}(q|\bm{E}=\bm{e})} = \mathbf{A}(\mathcal{I}(q ~\land~ \bm{E}=\bm{e})) ~\oslash~\mathbf{A}(\mathcal{I}(\bm{E}=\bm{e}))
\end{equation}

\noindent
where $\mathbf{A}(\mathcal{I}(q ~\land~ \bm{E}=\bm{e})) ~\oslash~ \mathbf{A}(\mathcal{I}(\bm{E}=\bm{e}))$ returns the label of $q ~\land~ \bm{E}=\bm{e}$ given the label of a set of pieces of evidence $\bm{E} = \bm{e}$.

In the case of probabilities as labels, i.e. $\rho(\cdot) \in [0,1]$, \eqref{eq:semiringprobability} presents the AMC-conditioning parametrisation $\semiringp$ for handling \textbf{PROB} of (conditioned) queries:

\begin{equation}
    \label{eq:semiringprobability}
    \begin{array}{l}
         \mathcal{A} = \mathbb{R}_{\geq 0}\\
         a ~\oplus~ b = a + b\\
         a ~\otimes~ b = a \cdot b\\
         e^\oplus = 0\\
         e^{\otimes} = 1\\
         \rho(f) \in [0,1]\\
         \rho(\lnot f) = 1 - \rho(f)\\
        \displaystyle{ a ~\oslash~ b = \frac{a}{b}}
    \end{array}
\end{equation}

A na\"ive implementation of \eqref{eq:fusion} is clearly exponential: \cite{Darwiche2004} introduced the first method for deriving tractable circuits (\dDNNF s) that allow polytime algorithms for clausal
entailment, model counting and enumeration.

\subsection{Probabilistic Circuits}
\label{sec:circuits}

As  AMC is defined in
terms of the set of models of a propositional logic theory, we can
exploit the succinctness results of 
the knowledge compilation map of~\cite{Darwiche2020knowledge}. 
The
restriction to two-valued variables allows us to directly compile AMC
tasks to circuits without adding constraints on legal
variable assignments to the theory.

In their knowledge compilation map, \cite{Darwiche2020knowledge}
provide an overview of succinctness relationships between various
types of  circuits. Instead of focusing on classical, flat target compilation languages based on conjunctive or disjunctive normal forms, \cite{Darwiche2020knowledge} 
consider a richer, nested class based on representing propositional sentences using directed acyclic
graphs: \NNF s.
A sentence in \emph{negation normal form} (\NNF) over a set of propositional variables $\mathcal{V}$
is a rooted, directed acyclic graph where each leaf node is labeled
with true ($\top$), false ($\bot$), or a literal of a variable in~$\mathcal{V}$, and
each internal node with disjunction ($\vee$) or conjunction ($\wedge$).

An \NNF\ is \emph{decomposable} if for each conjunction
node~$\bigwedge_{i=1}^n\phi_i$, no two children~$\phi_i$ and~$\phi_j$
share any variable. 

An \NNF\ is \emph{deterministic} if for each disjunction
node~$\bigvee_{i=1}^n\phi_i$, each pair of different 
children~$\phi_i$ and~$\phi_j$ is logically contradictory, that is $\phi_i \land \phi_j \models \bot$ for $i \neq j$. In other terms, only one child can be true at any time.\footnote{In the case $\phi_i$ and $\phi_j$ are seen as events in a sample space, the determinism can be equivalently rewritten as $\phi_i \cap \phi_j = \emptyset$ and hence $P(\phi_i \cap \phi_j) = 0$.}

The function \textsc{Eval} specified in Algorithm~\ref{alg:eval} \emph{evaluates} an \NNF\ circuit for a commutative semiring $(\mathcal{A},\oplus,\otimes, e^{\oplus},e^{\otimes})$ and labelling function $\labelfun$.
Evaluating an \NNF\ representation~$N_T$ of a propositional theory~$T$
for a  semiring $(\mathcal{A},\oplus,\otimes,e^{\oplus},e^{\otimes})$
and labelling function~$\labelfun$   is a
\emph{sound AMC computation} iff \textsc{Eval}$(N_T,\oplus,\otimes,e^{\oplus},e^{\otimes},\labelfun) =\operatorname{\mathbf{A}}(T)$.

\begin{algorithm}[t]
  \caption[\textsc{Label}]{Evaluating an \NNF\ circuit $N$ for a
    commutative semiring $(\mathcal{A},\oplus,\otimes,
    e^{\oplus},e^{\otimes})$ and labelling function $\labelfun$.}
\label{alg:eval}
\begin{algorithmic}[1]
\Procedure{\textsc{Eval}}{$N,\oplus,\otimes,e^{\oplus},e^{\otimes},\labelfun$}
\State \textbf{if} $N$ is a true node $\top$ \textbf{then} \textbf{return} $e^{\otimes}$
\State \textbf{if} $N$ is a false node $\bot$ \textbf{then} \textbf{return} $e^{\oplus}$
\State \textbf{if} $N$ is a literal node $l$ \textbf{then} \textbf{return} $\labelfun(l)$
\If{$N$ is a disjunction $\bigvee_{i=1}^mN_i$} \State \textbf{return} $\bigoplus_{i=1}^m$ \textsc{Eval}($N_i,\oplus,\otimes,e^{\oplus},e^{\otimes},\labelfun$) \EndIf
\If{$N$ is a conjunction $\bigwedge_{i=1}^mN_i$} \State \textbf{return} $\bigotimes_{i=1}^m$ \textsc{Eval}($N_i,\oplus,\otimes,e^{\oplus},e^{\otimes},\labelfun$) \EndIf
\EndProcedure
\end{algorithmic}
\end{algorithm}

In particular, \cite[Theorem 4]{kimmig2017-algebraic} shows that 
evaluating a \dDNNF\ representation of the propositional theory~$T$
for a semiring and labelling function with  neutral~$(\oplus,\labelfun)$
is a sound AMC computation. A semiring addition and labelling function pair~$(\oplus,\labelfun)$ is 
\emph{neutral} iff $\forall v \in \mathcal{V}: \labelfun(v) \oplus \labelfun(\neg v) =  e^{\otimes}$.

Unless specified otherwise, in the following we will refer to \dDNNF\ circuits labelled with probabilities or distributions of probability simply as circuits, and any addition and labelling function pair $(\oplus, \labelfun)$ are neutral. Also, we extend the definition of the labelling function such that it also operates on $\{\bot, \top \}$, i.e. $\labelfun(\bot) = e^{\oplus}$ and $\labelfun(\top) = e^{\otimes}$.

\begin{figure}
    \centering
    \includegraphics[width=\textwidth]{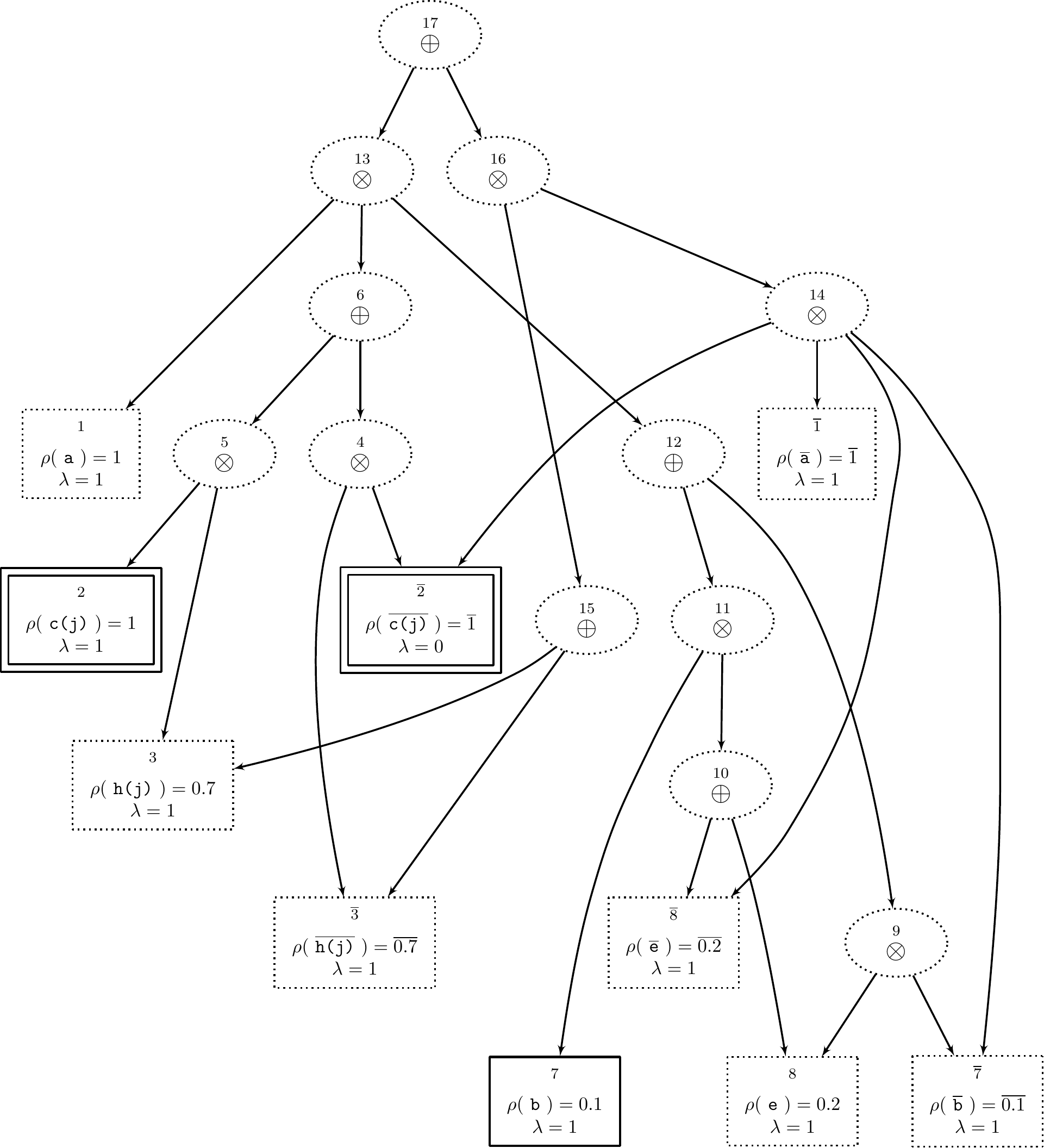}
    \caption{Circuit computing $p(\mtt{calls(john)}) $ for the Burglary example (Listing \ref{lst:problogburglary}). Solid box for query, double box for evidence.}
    \label{fig:burglarycircuit}
\end{figure}

Let us now introduce a graphical notation for circuits in this paper: Figure \ref{fig:burglarycircuit} illustrates a \dDNNF\ circuit where each node has a unique integer (positive or negative) identifier. Moreover, circled nodes are labelled either with $\oplus$ for disjunction (a.k.a. $\oplus$-gates) or with $\otimes$ for conjunction (a.k.a. $\otimes$-gates). Leaves nodes are marked with a squared box and they are labelled with the literal, $\top$, or $\bot$, as well as its label via the labelling function $\labelfun$. 

Unless specified otherwise, in the following we will slightly abuse the notation by defining an $\overline{\cdot}$ operator both for variables and $\top$, $\bot$, i.e. for $x \in \mathcal{V} \cup \{\bot, \top \}$, 
\begin{equation}
    \overline{x} = \left\{
    \begin{array}{l l}
    \lnot x & \text{if } x\in \mathcal{V}\\
    \bot & \text{if } x = \top\\
    \top & \text{if } x = \bot\\
    \end{array}
    \right.
\end{equation}

\noindent
and for elements of the set $\mathcal{A}$ of labels, s.t. $\overline{\rho(x)} = \rho(~\overline{x}~)$.

Finally, each leaf node presents an additional parameter $\lambda$---i.e. the indicator variable cf. \cite{Fierens2015}---that assumes values 0 or 1, and we will be using it for reusing the same circuit for different purposes.

In the following, we will make use of a running example based upon the burglary example as presented in \cite[Example 6]{Fierens2015}. In this way, we hope to convey better to the reader the value of our approach as the circuit derived from it using \cite{Darwiche2004} will have a clear, intuitive meaning behind. However, our approach is independent from the system that employs circuit compilation for its reasoning process, as long as it can make use of \dDNNF s circuits. The \dDNNF\ circuit for our running example is depicted in Figure \ref{fig:burglarycircuit} and has been derived by compiling \cite{Darwiche2004} in a \dDNNF\  the ProbLog \cite{Fierens2015} code listed in Listing \ref{lst:problogburglary} \cite[Example 6]{Fierens2015}. For compactness, in the graph each literal of the program is represented only by the initials, i.e. \texttt{burglary} becomes \texttt{b}, \texttt{hears\_alarm(john)} becomes \texttt{h(j)}. ProbLog is an approach to augment\footnote{We refer readers interested in probabilistic augmentation of logical theories in general to \cite{DBLP:journals/ijar/CeruttiT19}.} prolog programs \cite{Kowalski:1988,Bratko:2001} annotating facts\footnote{Albeit ProbLog allows for rules to be annotated with probabilities: rules of the form \texttt{p::h :- b} are translated into \texttt{h :- b,t} with \texttt{t} a new fact of the form \texttt{p::t}.} with probabilities: see Appendix \ref{sec:aproblog} for an introduction. As discussed in \cite{Fierens2015}, the prolog language admits a propositional representation of its semantics. For the example the propositional representation of Listing \ref{lst:problogburglary} is:
\begin{equation}
\label{eq:propburglary}
\begin{array}{c}
    \texttt{alarm} \leftrightarrow \texttt{burglary} \lor \texttt{earthquake}\\
    \texttt{calls(john)} \leftrightarrow \texttt{alarm} \land \texttt{hears\_alarm(john)}\\
    \texttt{calls(john)}
    \end{array}
\end{equation}

Figure \ref{fig:burglarycircuit} thus shows the result of the compilation of \eqref{eq:propburglary} in a circuit, annotated with a unique id that is either a number $x$ or $\overline{x}$ to indicate the node that represent the negation of the variable represented by node $x$; and with weights (probabilities) as per Listing \ref{lst:problogburglary}.

To enforce that we know \texttt{calls(john)} is true (see line 7 of Listing \ref{lst:problogburglary}). This translates in having $\lambda = 1$ for the double boxed node with index $2$ in Figure \ref{fig:burglarycircuit}---that indeed is labelled with the shorthand for \texttt{calls(john)}, i.e. \texttt{c(j)}---while $\lambda = 0$ for the double boxed node with index $\overline{2}$ that is instead labelled with the shorthand for $\overline{\texttt{calls(john)}}$, i.e. $\overline{\texttt{c(j)}}$.

\begin{algorithm}[t]
  \caption[\textsc{Label}]{Evaluating an \NNF\ circuit $N$ for a
    commutative semiring $(\mathcal{A},\oplus,\otimes,
    e^{\oplus},e^{\otimes})$ and labelling function $\labelfun$, considering indicators $\lambda$.}
\label{alg:evallambda}
\begin{algorithmic}[1]
\Procedure{\textsc{Eval}}{$N,\oplus,\otimes,e^{\oplus},e^{\otimes},\labelfun$}
\If{$N$ is a true node $\top$}
    \State \textbf{if} $\lambda = 1$ \textbf{then} \textbf{return} $e^{\otimes}$
    \State \textbf{else} \textbf{return} $e^{\oplus}$
\EndIf
\State \textbf{if} $N$ is a false node $\bot$ \textbf{then} \textbf{return} $e^{\oplus}$
\If{$N$ is a literal node $l$}
    \State \textbf{if} $\lambda = 1$ \textbf{then} \textbf{return} $\labelfun(l)$
    \State \textbf{else} \textbf{return} $e^{\oplus}$
\EndIf
\If{$N$ is a disjunction $\bigvee_{i=1}^mN_i$} \State \textbf{return} $\bigoplus_{i=1}^m$ \textsc{Eval}($N_i,\oplus,\otimes,e^{\oplus},e^{\otimes},\labelfun$) \EndIf
\If{$N$ is a conjunction $\bigwedge_{i=1}^mN_i$} \State \textbf{return} $\bigotimes_{i=1}^m$ \textsc{Eval}($N_i,\oplus,\otimes,e^{\oplus},e^{\otimes},\labelfun$) \EndIf
\EndProcedure
\end{algorithmic}
\end{algorithm}

The $\lambda$ indicators modify the execution of the function \textsc{Eval} (Alg. \ref{alg:eval}) in the way illustrated by Algorithm \ref{alg:evallambda}: note that Algorithm \ref{alg:evallambda} is analogous to Algorithm \ref{alg:eval} when all $\lambda=1$. Hence, in the following, when considering the function \textsc{Eval}, we will be referring to the one defined in Algorithm \ref{alg:evallambda}.

\begin{lstlisting}[columns=fullflexible,caption={Problog code for the Burglary example, originally Example 6 in  \protect\cite{Fierens2015}.},label={lst:problogburglary},captionpos=b,float,numbers=left,
    stepnumber=1]
0.1::burglary.
0.2::earthquake.
0.7::hears_alarm(john).
alarm :- burglary.
alarm :- earthquake.
calls(john) :- alarm, hears_alarm(john).
evidence(calls(john)).
query(burglary).
\end{lstlisting}

Finally, the ProbLog program in Listing \ref{lst:problogburglary} queries the value of \texttt{burglary}, hence we need to compute the probability of \texttt{burglary} given \texttt{calls(john)}, 

\begin{equation}\label{eq:Pproblogburglary}
    \displaystyle{p(\mtt{burglary} \mid \mtt{calls(john)}) = \frac{p(\mtt{burglary} \land \mtt{calls(john)})}{p(\mtt{calls(john)})}} 
\end{equation}

\begin{figure*}
    \centering
    \includegraphics[width=\textwidth]{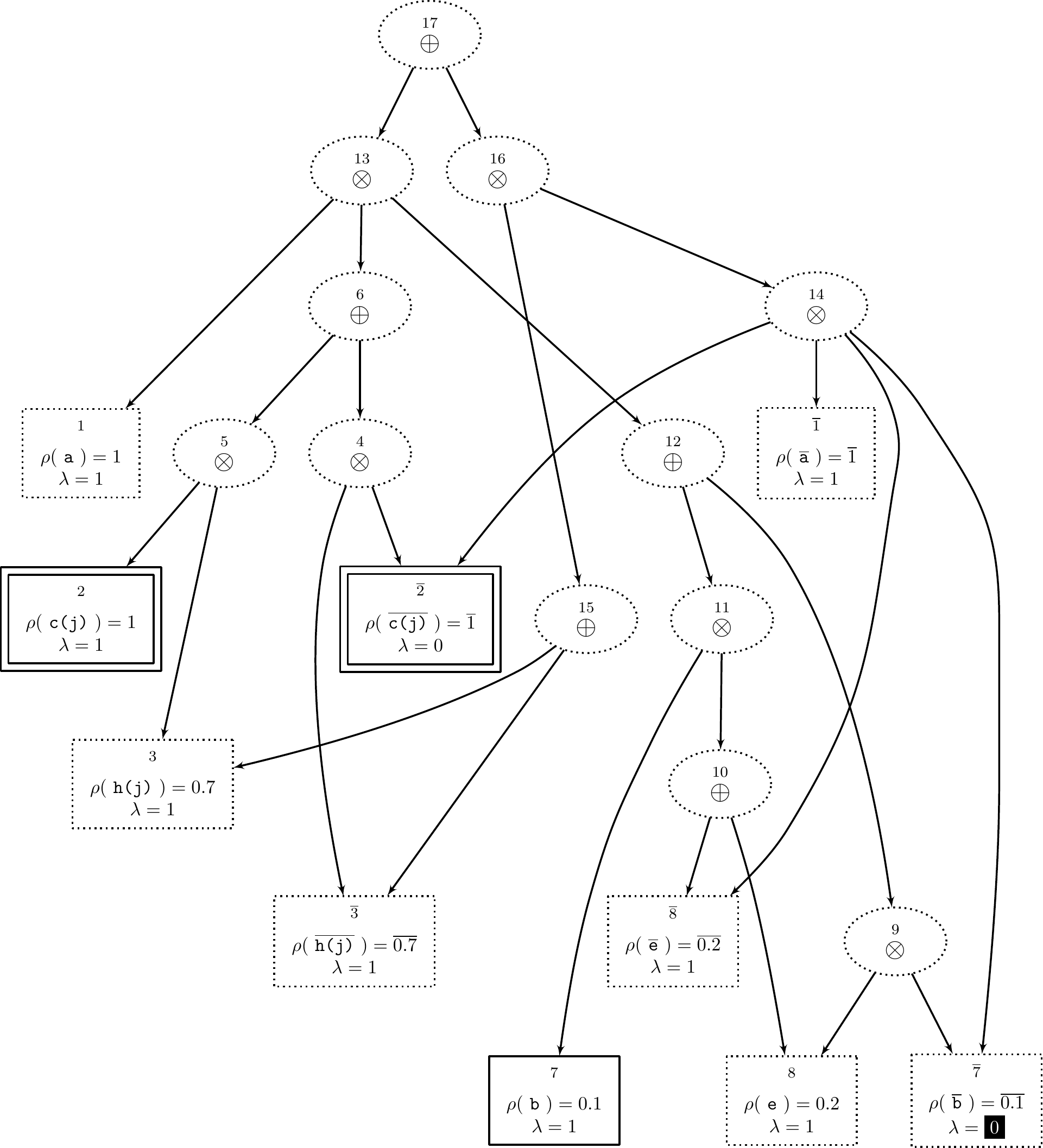}
    \caption{Circuit computing $p(\mtt{burglary} \land \mtt{calls(john)})$  for the Burglary example (Listing \ref{lst:problogburglary}). Solid box for query, double box for evidence.
    White over black for the numeric value that has changed from Figure \ref{fig:burglarycircuit}.
    In particular, in this case, $\lambda$ for the node labelled with $\overline{\mtt{burglary}}$ is set to 0.}
    \label{fig:burglarycircuitquery}
\end{figure*}

While the denominator of \eqref{eq:Pproblogburglary} is given by \textsc{Eval} of the circuit in Figure \ref{fig:burglarycircuit}, we need to modify it in order to obtain the numerator $p(\mtt{burglary} \land \mtt{calls(john)})$ as depicted in Figure \ref{fig:burglarycircuitquery}, where $\lambda = 0$ for the node labelled with $\overline{\mtt{burglary}}$. \textsf{Eval} on the circuit in Figure \ref{fig:burglarycircuitquery} will thus return the value of the denominator in \eqref{eq:Pproblogburglary}.

It is worth highlighting that computing $p(query \mid evidences)$ for an arbirtrary $query$ and arbirtrary set of $evidences$ requires \textsc{Eval} to be executed at least twice on slightly modified circuits.

In this paper, similarly to \cite{Kisa2014}, we are interested in learning the parameters of our circuit, i.e. the $\labelfun$ function for each of the leaves nodes, or $\bm{\labelfun}$ in the following, thus representing it as a vector. We will learn $\bm{\labelfun}$ from a set of \emph{examples}, where each \emph{example} is an instantiation of all propositional variables: for $n$ propositional variables, there are $2^n$ of such instantiations. In the case the circuit is derived from a logic program, an example is a complete interpretation of all the ground atoms.  A \emph{complete dataset} $\Data$ is then a sequence (allowing for repetitions) of examples, each of those is a vector of instantiations of independent Bernoulli distributions with true but unknown parameter $\bm{p_x}$. From this, the \emph{likelihood} is thus:

\begin{equation}
    \label{eq:vec-likelihood}
    p(\Data \mid \bm{p}) = \prod_{i=1}^{|\Data|} p(\bm{x}_i \mid \bm{p}_{x_i})
\end{equation}

\noindent 
where $\bm{x}_i$ represents the $i$-th example in the dataset \Data.
Differently, however, from \cite{Kisa2014}, we do not search for a maximum likelihood solution of this problem, rather we provide a Bayesian analysis of it in Section \ref{sec:betasl}. 

The following analysis provides the distribution of the probabilities (second-order probabilities)  for each propositional variables.  For complete datasets, these distributions factor meaning that the second-order probabilities for the propositional variables are statistically independent (see Appendix \ref{sec:independence-posterior}). Nevertheless, it is shown that second-order probabilities of a variable and its negation are correlated because the first-order probabilities (i.e. the expected values of the distributions) sum up to one. 

The inference process proposed in this paper does not assume independent second-order probabilities as it encompasses dependencies between the random variables associated to the proposition in the form of a covariance matrix.  For complete datasets, the covariances at the leaves are only non-zero between a variable and its negation.  More generally, when the training dataset $\mathcal{D}$ is not complete (i.e., variable values cannot always be observed for various instantiations), the second-order probabilities become correlated.  The derivations of these correlations during the learning process with partial observations is left for future work. Nevertheless, the proposed inference method can accommodate such correlations without any modifications.

This is one of our main contributions, that separates our approach from the literature. Indeed, ours is clearly not the only Bayesian approach to learning parameters in circuits, see for instance \cite{Jaini2016,Zhao2016,Trapp2019,Vergari2019,Rashwan2016,Zhao2016a}. In addition, similarly to \cite{Rashwan2016,Jaini2016} we also apply the idea of moment matching instead of using sampling.

\section{A Bayesian Account of Uncertain Probabilities}
\label{sec:betasl}

Let us now expand further \eqref{eq:vec-likelihood}: for simplicity, let us consider here only the case of a single propositional variable, i.e. a single binary random variable $x \in \set{0, 1}$, e.g. flipping coin, not necessary fair, whose probability is thus conditioned by a parameter $0 \leq p_x \leq 1$:

\begin{equation}
    \label{eq:2.1}
    p(x = 1 \mid p_x) = p_x
\end{equation}

\noindent
The probability distribution over $x$ is known as the \emph{Bernoulli} distribution:

\begin{equation}
    \label{eq:2.2}
    \dbern(x \mid p_x) = p_x^x (1-p_x)^{1-x}
\end{equation}

Given a data set $\Data$ of i.i.d. observations  $\vt{\vp{x_1, \ldots, x_N}}$ drawn from the Bernoulli with parameter $p_x$, which is assumed unknown, the \emph{likelihood} of data given $p_x$ is:

\begin{equation}
    \label{eq:2.5}
    p(\Data \mid p_x) = \prod_{n=1}^{N} p(x_n \mid p_x) = \prod_{n=1}^{N} p_x^{x_n} (1 - p_x)^{1 - x_n}
\end{equation}

To develop a Bayesian analysis of the phenomenon, we can choose as prior the beta distribution, with parameters $\bm{\alpha} = \tuple{\alpha_x, \alpha_{\overline{x}}}$, $\alpha_x \geq 1$ and $\alpha_{\overline{x}} \geq 1$, that is conjugate to the Bernoulli:

\begin{equation}
    \label{eq:2.13}
    \dbeta(p_x \mid \bm{\alpha}) = \frac{\Gamma(\alpha_x + \alpha_{\overline{x}})}{\Gamma(\alpha_x) \Gamma(\alpha_{\overline{x}})} p_x^{\alpha_x-1} (1 - p_x)^{\alpha_{\overline{x}} -1}
\end{equation}
where
\begin{equation}
    \label{eq:1.141}
    \Gamma(t) \equiv \int_0^{\infty} u^{t-1} e^{-u} \dd u
\end{equation}
is the gamma function.

Given a beta-distributed random variable $X$, 
\begin{equation}
\label{eq:strength}
s_X = \alpha_x + \alpha_{\overline{x}}
\end{equation}
 is its \emph{Dirichlet strength} and 
\begin{equation}
    \label{eq:mean}
    \ev[X] = \frac{\alpha_x}{s_X}
\end{equation}
is its expected value. From \eqref{eq:strength} and \eqref{eq:mean} the beta parameters can  equivalently be written as:
\begin{equation}
    \label{eq:parameterssx}
    \bm{\alpha_X} = \tuple{\ev[X] s_X, ~(1-\ev[X]) s_X} .
\end{equation}
The variance of a beta-distributed random variable $X$ is
\begin{equation}
\label{e:pred_var}
\vv[X] = \vv[1-X] = \frac{\ev[X] (1 - \ev[X])}{s_X+1}
\end{equation}
and 
because $X+(1-X) = 1$, it is easy to see that
\begin{equation}
\label{e:covvv}
    \cv[X,1-X] = -\vv[X].
\end{equation}

From \eqref{e:pred_var} we can rewrite $s_X$ \eqref{eq:strength} as
\begin{equation}
\label{eq:sxvar}
s_X = \frac{\ev[X] (1-\ev[X])}{\vv[X]} - 1 .
\end{equation}

Considering a beta distribution prior and the binomial likelihood function, and given $N$ observations of $x$ such that for $r$ observations $x=1$ and for $s=N-r$ observations $x=0$

\begin{equation}
    \label{eq:2.17}
    \displaystyle{p(p_x \mid \Data, \prior{\bm{\alpha}}) = \frac{p(\Data \mid p_x) p(p_x \mid \prior{\bm{\alpha}})}{p(\Data)} ~\propto~ p_x^{r + \prior{\alpha_x} -1} (1-p_x)^{s + \prior{\alpha_{\overline{x}}} -1}}
\end{equation}

Hence $p(p_x \mid r, s, \prior{\bm{\alpha}})$ is another beta distribution such that after normalization via p(D),

\begin{equation}
    \label{eq:2.18}
    p(p_x \mid r, s, \prior{\bm{\alpha}}) = \frac{\Gamma(r + \prior{\alpha_x} + s + \prior{\alpha_{\overline{x}}})}{\Gamma(r + \prior{\alpha_x}) \Gamma(s + \prior{\alpha_{\overline{x}}})} p_x^{r + \prior{\alpha_x} -1} (1 - p_x)^{s + \prior{\alpha_{\overline{x}}} -1}
\end{equation}

We can specify the parameters for the prior we are using for deriving our beta distributed random variable $X$ as $\prior{\bm{\alpha}} = \tuple{a_X W, (1-a_X) W}$ where $\slbase{X}$ is the prior assumption, i.e. $p(x = 1)$ in the absence of observations;
and $W > 0$ is a prior weight indicating the strength of the prior assumption. Unless specified otherwise, in the following we will assume $\forall X, \slbase{X} = 0.5$ and $W = 2$, so to have an uninformative, uniformly distributed, prior.

The complete dataset $\mathcal{D}$ is modelled as samples from independent binomials distributions for facts and rules. As such, the posterior factors as a product of beta distributions representing the posterior distribution for each fact or rule as in (\ref{eq:2.18}) for a single fact (see Appendix~\ref{sec:independence-posterior} for further details). This posterior distribution enable the computation of the means and covariances for the leaves of the circuit, and because it factors, the different variables are statistically independent leading to zero covariances. Only the leaves associated to a variable and its complement exhibit nonzero covariance via (\ref{e:covvv}). Now, the means and covarainces of the leaves can be propagated through the circuit to determine the distribution of the queried conditional probability as described in Section~\ref{sec:contribution}.

Given an inference, like the conditioned query of our running example \eqref{eq:Pproblogburglary}, we approximate its distribution by a beta distribution by finding the corresponding Dirichlet strength to match the compute variance.
Given a random variable $Z$ with known mean $\mean{Z}$ and variance $\var{Z}$, we can use the method of moments and \eqref{eq:sxvar} to estimate the $\bm{\alpha}$ parameters of a beta-distributed variable $Z'$ of mean $\mean{Z'} = \mean{Z}$ and
\begin{equation}
    \label{eq:minvarcheck}
    s_{Z'} = \max\left\{\frac{\ev[Z] (1-\ev[Z])}{\vv[Z]} -1, \frac{W \slbase{Z}}{\ev[Z]}, \frac{W (1 - \slbase{Z})}{(1-\ev[Z])} \right\}.
\end{equation}
\eqref{eq:minvarcheck} is needed to ensure that the resulting beta-distributed random variable $Z'$ does not lead to a $\bm{\alpha}_{Z'} < \tuple{1,1}$.

\subsection{Subjective Logic}
Subjective logic \cite{Josang2016-SL} provides (1) an alternative, more intuitive, way of representing the parameters of beta-distributed random variables, and (2) a set of operators for manipulating them. A subjective opinion about a proposition $X$ is a tuple $\slop{X} = \sloptuple{X}$, representing the belief,
disbelief and uncertainty that $X$ is true at a given instance, and, as above, $\slbase{X}$  is the prior probability that $X$ is true in the absence of observations. These values are non-negative and $\slbel{X} + \sldis{X} + \slunc{X} = 1$. The
projected probability $p(x) =\slbel{X} + \slunc{X} \cdot \slbase{X}$, provides an estimate of the ground truth probability $p_x$. 

The mapping from a beta-distributed random variable $X$ with parameters $\bm{\alpha}_X = \tuple{\alpha_{x}, \alpha_{\overline{x}}}$ to a subjective opinion is:
\begin{equation}
    \slop{X} = \left\langle\frac{\alpha_x - W \slbase{X}}{s_X}, \frac{\alpha_{\overline{x}} - W (1 - \slbase{X})}{s_X}, \frac{W}{s_X}, \slbase{X}\right\rangle
\end{equation}
With this transformation, the mean of $X$ is equivalent to the projected probability $p(x)$, and the Dirichlet strength is inversely proportional to the uncertainty of the opinion:
\begin{equation}
    \ev[X] = p(x) = \slbel{X} + \slunc{X} \slbase{X}, \quad s_X = \frac{W}{\slunc{X}}
\end{equation}

Conversely, a subjective opinion $\slop{X}$  translates directly into a beta-distributed random variable with:
\begin{equation}
    \bm{\alpha}_X = \left\langle\frac{W}{\slunc{X}} \slbel{X} + W \slbase{X}, \frac{W}{\slunc{X}} \sldis{X} + W (1 - \slbase{X})\right\rangle
\end{equation}

Subjective logic is a framework that includes various operators to indirectly determine opinions from various logical operations. In particular, we will make use of $\boxplus_{SL}$, $\boxtimes_{SL}$, and $\boxslash_{SL}$, resp. summing, multiplying, and dividing two subjective opinions as they are defined in \cite{Josang2016-SL} (Appendix \ref{sec:sl-operators}).
Those operators aim at faithfully matching the projected probabilities: for instance the multiplication of two subjective opinions $\slop{X} \boxtimes_{SL} \slop{Y}$ results in an opinion $\slop{Z}$ such that $p(z) = p(x) \cdot p(z)$. 

\section{AMC-conditioning parametrisation with strong independence assumptions}
\label{sec:parametrisation}

\begin{figure}
    \centering
    \includegraphics[width=\textwidth]{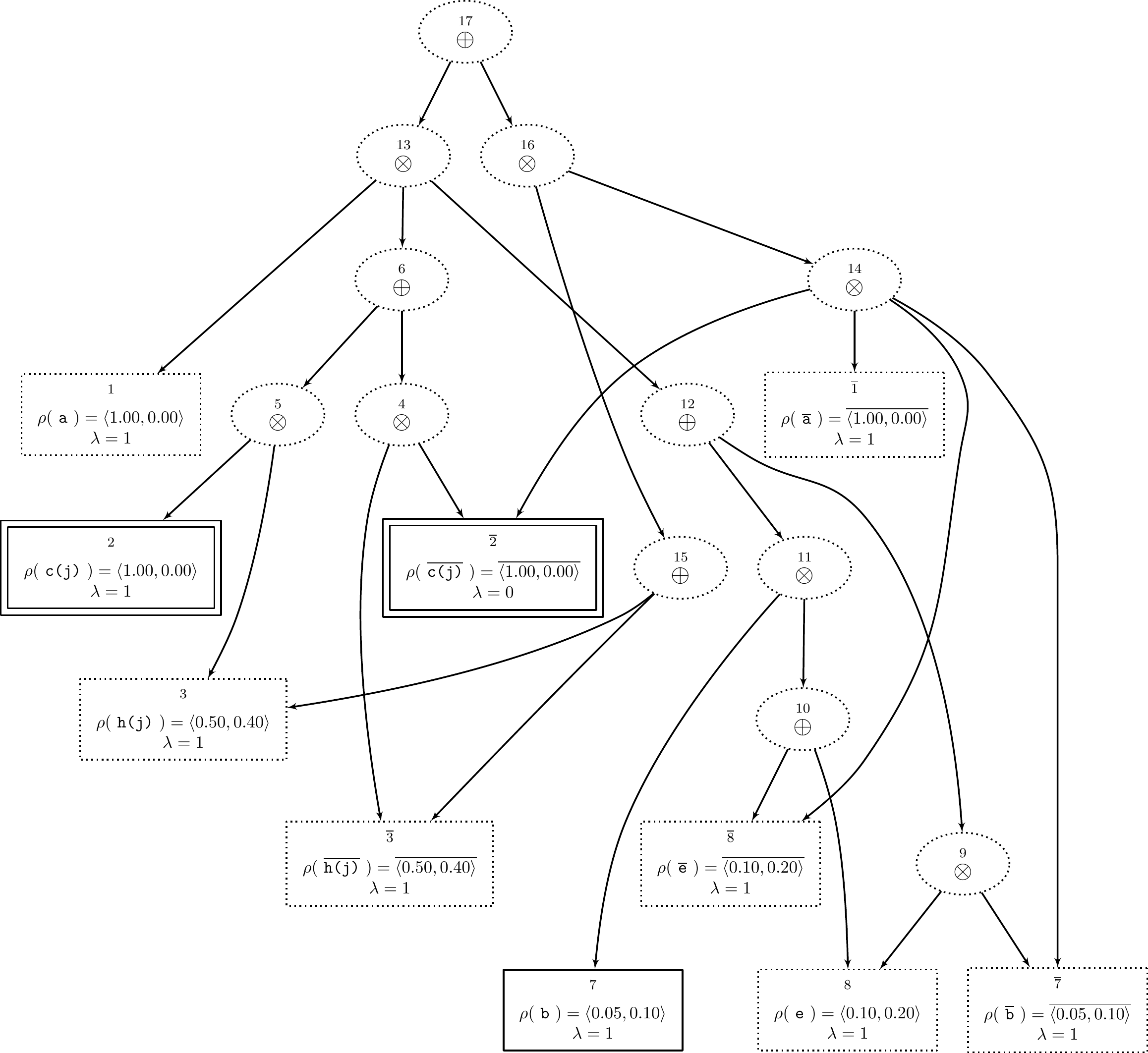}
    \caption{Variation on the circuit represented in Figure \ref{fig:burglarycircuit} with leaves labelled with imprecise probabilities represented as Subjective Logic opinions, listing only $\slbel{X}$ and $\slunc{X}$: $\sldis{X} = 1 - \slbel{X} - \slunc{X}$, and $\slbase{X} = 0.5$. Solid box for query, double box for evidence.}
    \label{fig:burglarySL}
\end{figure}

Building upon our previous work \cite{DBLP:conf/aaai/CeruttiKKS19}, we allow manipulation of imprecise probabilities as labels in our circuits.  Figure \ref{fig:burglarySL} shows an example of the circuits we will be manipulating, where probabilities from the circuit depicted in Fig. \ref{fig:burglarycircuit} has been replaced by uncertain probabilities represented as beta-distributed random variables and formalised as SL opinion, in a shorthand format listing only belief and uncertainty values. 

\subsection{SL AMC-conditioning parametrisation with strong independence assumptions}

The straightforward approach to derive an AMC-conditioning parametrisation under complete independence assumptions at each step of the evaluation of the probabilistic circuit using subjective logic, is to use the operators $\boxplus$, $\boxtimes$, and $\boxslash$. This gives rise to the 
SL AMC-conditioning parametrisation $\semiringsl$, defined as follows:
\begin{equation}
    \label{eq:semiringsloperators}
    \def\arraystretch{2}
    \displaystyle{\begin{array}{l}
          \mathcal{A_{\text{SL}}} = \mathbb{R}_{\geq 0}^4\\
          a ~\oplus_{\text{SL}}~ b = 
            \left\{
            {\def\arraystretch{1}
            \begin{array}{l@{\hskip 2em} l}
            a & \text{if } b = e^{\oplus_{\text{SL}}}\\
            b & \text{if } a = e^{\oplus_{\text{SL}}}\\
            a ~\boxplus_{\text{SL}}~ b & \text{otherwise}\\
            \end{array}}
            \right.\\
            \\[-1.5em]
          a ~\otimes_{\text{SL}}~ b = 
          \left\{
            {\def\arraystretch{1}
            \begin{array}{l@{\hskip 2em} l}
            a & \text{if } b = e^{\otimes{\text{SL}}}\\
            b & \text{if } a = e^{\otimes_{\text{SL}}}\\
            a ~\boxtimes_{\text{SL}} b & \text{otherwise}\\
            \end{array}}
            \right.\\
          e^{\oplus_{\text{SL}}} =  \tuple{0, 1, 0, 0}\\
          e^{\otimes_{\text{SL}}} = \tuple{1, 0, 0, 1}\\
          \rho_{\text{SL}}(f_i) = \sloptuple{f_i}\\
          \rho_{\text{SL}}(\lnot f_i) = \tuple{\sldis{f_i}, \slbel{f_i}, \slunc{f_i}, 1 - \slbase{f_i}}\\
          a ~\oslash_{\text{SL}}~ b 
           = \left\{
          {\def\arraystretch{1}
            \begin{array}{l@{\hskip 2em} l}
            a & \text{if } b = e^{\otimes_{\text{SL}}}\\
            a ~\boxslash_{\text{SL}}~ b  & \mbox{if defined}\\
            \tuple{0,0,1,0.5} & \mbox{otherwise}
          \end{array}}
          \right.
    \end{array}}
\end{equation}

Note that $\tuple{\mathcal{A_{\text{SL}}}, \oplus_{\text{SL}}, \otimes_{\text{SL}}, e^{\oplus_{\text{SL}}}, e^{\otimes_{\text{SL}}}}$ does not form a commutative semiring in general. If we consider only the projected probabilities---i.e. the means of the associated beta distributions---then $\boxplus$ and $\boxtimes$ are indeed commutative, associative, and $\boxtimes$ distributes over $\boxplus$. However, the uncertainty of the resulting opinion depends on the order of operands.

\subsection{Moment Matching AMC-conditioning parametrisation with strong independence assumptions}

In \cite{DBLP:conf/aaai/CeruttiKKS19} we derived another set of operators operating with moment matching: they aim at maintaining a stronger connection to beta distribution as the result of the manipulation.
Indeed, while SL operators try to faithfully characterise the projected probabilities, they employ an uncertainty maximisation principle to limit the belief commitments, hence they have a looser connection to the beta distribution. Instead, in \cite{DBLP:conf/aaai/CeruttiKKS19} we first represented beta distributions (and thus also SL opinions) not parametric in $\bm{\alpha}$, but rather parametric on mean and variance. Hence we then propose operators that manipulate means and variances, and then we transformed them back into beta distributions by moment matching.

In \cite{DBLP:conf/aaai/CeruttiKKS19} we first defined a sum operator between two independent beta-distributed random variables $X$ and $Y$ as the beta-distributed random variable $Z$ such that $\mean{Z} = \mean{X + Y}$ and $\variance{Z} = \variance{X + Y}$. The sum (and in the following the product as well) of two beta random variables is not necessarily a beta random variable. Consistently with \cite{KAPLAN2018132}, the resulting distribution is then approximated as a beta distribution via moment matching on mean and variance.

Given $X$ and $Y$ independent beta-distributed random variables represented by the subjective opinion $\slop{X}$ and $\slop{Y}$, the \emph{sum} of $X$ and $Y$ ($\slop{X} \boxplus^\beta \slop{Y}$) is defined as the beta-distributed random variable $Z$ such that:
    \begin{equation}
        \mean{Z} = \mean{X + Y} = \mean{X} + \mean{Y}
    \end{equation}
and
\begin{equation}
    \label{eq:sumbetavar}
    \variance{Z} = \variance{X + Y} = \variance{X} + \variance{Y}.
\end{equation}
$\slop{Z} = \slop{X} \boxplus^\beta \slop{Y}$ can then be obtained as discussed in Section \ref{sec:betasl}, taking \eqref{eq:minvarcheck} into consideration. The same applies for the following operators as well.

The product operator between two independent beta-distributed random variables $X$ and $Y$ is then defined as the beta-distributed random variable $Z$ such that $\mean{Z} = \mean{XY}$ and $\variance{Z} = \variance{XY}$. 
Given $X$ and $Y$ independent beta-distributed random variables represented by the subjective opinion $\slop{X}$ and $\slop{Y}$, the \emph{product} of $X$ and $Y$ ($\slop{X} \boxtimes^\beta \slop{Y}$) is defined as the beta-distributed random variable $Z$ such that:
\begin{equation}
    \mean{Z} = \mean{XY} = \mean{X} ~ \mean{Y}
\end{equation}
and 
\begin{equation}
\label{eq:prodbetavar}
        \variance{Z} = \variance{XY} =  \variance{X} (\mean{Y})^2 + \variance{Y} (\mean{X})^2 + \variance{X} \variance{Y}.
\end{equation}

Finally, the conditioning-division operator between two independent beta-distributed random variables $X$ and $Y$, represented by subjective opinions $\slop{X}$ and $\slop{Y}$, is the beta-distributed random variable $Z$ such that $\mean{Z} = \mean{\frac{X}{Y}}$ and $\variance{Z} = \variance{\frac{X}{Y}}$. Given $\slop{X} = \sloptuple{X}$ and $\slop{Y} = \sloptuple{Y}$ subjective opinions such that X and Y are beta-distributed random variables, $Y = \mathbf{A}(\mathcal{I}(\bm{E}=\bm{e})) = \mathbf{A}(\mathcal{I}(q ~\land~ \bm{E}=\bm{e})) ~\oplus~ \mathbf{A}(\mathcal{I}(\lnot q ~\land~ \bm{E}=\bm{e}))$, with $\mathbf{A}(\mathcal{I}(q ~\land~ \bm{E}=\bm{e})) = X$. The \emph{conditioning-division} of $X$ by $Y$ ($\slop{X} \boxslash^\beta \slop{Y}$) is defined as the beta-distributed random variable $Z$ such that:

\begin{equation}
    \begin{split}
        \mean{Z} = \ev\left[\frac{X}{Y}\right] = \mean{X} ~\ev\left[\frac{1}{Y}\right] \simeq \frac{\mean{X}}{\mean{Y}}
    \end{split}
\end{equation}
and\footnote{Please note that \eqref{eq:variance-division} corrects a typos that is present in its version in \cite{DBLP:conf/aaai/CeruttiKKS19}.}

\begin{equation}
    \label{eq:variance-division}
    \begin{split}
        \variance{Z} & \simeq  (\mean{Z})^2 (1 - \mean{Z})^2  \left( \frac{\variance{X}}{(\mean{X})^2}
        + \frac{\variance{Y} + \variance{X}}{(\mean{Y} - \mean{X})^2} +  \frac{2 \variance{X}}{\mean{X} (\mean{Y} - \mean{X})}
        \right)
    \end{split}
\end{equation}

Similarly to \eqref{eq:semiringsloperators}, 
the moment matching AMC-conditioning parametrisation $\semiringbeta$ is defined as follows:
\begin{equation}
    \label{eq:semiringsloperatorsbeta}
    {\def\arraystretch{2}
    \begin{array}{l}
          \mathcal{A^\beta} = \mathbb{R}_{\geq 0}^4\\
          a ~\oplus^\beta~ b = 
            a ~\boxplus^\beta~ b\\
          a ~\otimes^\beta~ b = 
            a ~\boxtimes^\beta b\\
          e^{\oplus^\beta} = \tuple{1, 0, 0, 0.5}\\
          e^{\otimes^\beta} = \tuple{0, 1, 0, 0.5}\\
          \rho^\beta(f_i) = \sloptuple{f_i} \in [0,1]^4\\
          \rho^\beta(\lnot f_i) = \tuple{\sldis{f_i}, \slbel{f_i}, \slunc{f_i}, 1 - \slbase{f_i}}\\
          a ~\oslash^\beta~ b = a ~\boxslash^\beta~ b
    \end{array}
    }
\end{equation}

As per \eqref{eq:semiringsloperators}, also $\tuple{\mathcal{A^\beta}, \oplus^{\beta}, \otimes^{\beta}, e^{\oplus^{\beta}}, e^{\otimes^{\beta}}}$ is not in general a commutative semiring. Means are correctly matched to projected probabilities, therefore for them $\semiringbeta$ actually operates as a semiring. However, for what concerns variance, by using \eqref{eq:prodbetavar} and \eqref{eq:sumbetavar}---thus under independence assumption---the product is not distributive over addition:  $\var{X (Y + Z)} = \var{X} (\mean{Y} + \mean{Z})^2 + (\var{Y} + \var{Z}) \mean{X}^2 + \var{X} (\var{Y} + \var{Z}) \neq \var{X} (\mean{Y}^2 + \mean{Z}^2) + (\var{Y} + \var{Z}) \mean{X}^2 + \var{X} (\var{Y} + \var{Z}) = \var{(XY) + (XZ)}$.

To illustrate the discrepancy, let's consider node 6 in Figure \ref{fig:burglarySL}: the disjunction operator there is summing up probabilities that are not statistically independent, despite the independence assumption used in developing the operator.  Due to the dependencies between nodes in the circuit, the error grows during propagation, and then the numerator and denominator in the conditioning operator exhibit strong correlation due to redundant operators.  Therefore, \eqref{eq:variance-division} introduces further error leading to an overall inadequate characterisation of variance.  The next section reformulates the operations to account for the existing correlations.

\section{CPB: Covariance-aware Probabilistic inference with beta-distributed random variables}
\label{sec:contribution}

We now propose an entirely novel approach to the AMC-conditioning problem that  considers the covariances between the various distributions we are manipulating. Indeed, our approach for computing Covariance-aware Probabilistic entailment with beta-distributed random variables \lbetaprobcov\ is designed to satisfy the total probability theorem, and in particular to enforce that for any $X$ and $Y$ beta-disributed random variables,

\begin{equation}
   \vv[Y \otimes X] \oplus \vv[Y \otimes \overline{X}] = \vv[Y] 
\end{equation}

Algorithm \ref{alg:overview} provides an overview of \lbetaprobcov, that comprises three stages: (1) pre-processing; (2) circuit shadowing; and (3) evaluation.
The overall approach is to view the second-order probability of each node in the circuit as a beta distribution. The determination of the distributions is through moment matching via the first and second moments through \eqref{e:pred_var} and \eqref{eq:sxvar}.  Effectively, the collection of nodes are treated as multivariate Gaussian characterised by a mean vector and covariance matrix that it computed via the propagation process described below. When analysing the distribution for particular node (via marginalisation of the Gaussian), it is approximated via the best-fitting beta distribution through moment-matching. 

\subsection{Pre-processing} 
\label{sec:preprocessing}

We assume that the circuit we are receiving has the leaves labelled with unique identifiers of beta-distributed random variables. We also allow for the specification of the covariance matrix between the beta-distributed random variables, bearing in mind that $\cv[X, 1-X] = -\vv[X]$, cf. \eqref{e:pred_var} and \eqref{e:covvv}.
We do not provide a specific algorithm for this, as it would depend on the way the circuit is computed. In our running example, we assume the ProbLog code from Listing \ref{lst:problogburglary} has been transformed into the aProbLog\footnote{aProbLog \cite{DBLP:conf/aaai/KimmigBR11} is the algebraic version of ProbLog that allows  for arbitrary labels to be used.} code in Listing \ref{lst:problogburglaryuniqueids}.
    
We also expect there is a table associating the identifier with the actual value of the beta-distributed random variable. In the following, we assume that $\omega_1$ is a reserved indicator for the $\dbeta(\infty, 1.00)$ (in Subjective Logic term $\tuple{1.0, 0.0, 0.0, 0.5}$). For instance, Table \ref{tab:assoctable} provides the associations for code in Listing \ref{lst:problogburglaryuniqueids}, and Table \ref{tab:covariance} the covariance matrix for those beta-distributed random variables that we assume being learnt from complete observations of independent random variables, and hence the posterior beta-distributed random variables are also independent (cf. Appendix~\ref{sec:independence-posterior}).

\newcommand{\algoverviewname}{\textsc{CovProbBeta}}
\newcommand{\alguniqueidsname}{\textsc{IDsRVs}}
\newcommand{\algexpandname}{\textsc{ShadowCircuit}}
\newcommand{\algevalname}{\textsc{EvalCovProbBeta}}
\newcommand{\fqchildren}{\textsc{shchildren}}

\begin{algorithm}[t]
  \caption[\textsc{Label}]{Solving the \textbf{PROB} problem on a circuit $N_A$ labelled with identifier of beta-distributed random variables and the associative table $A$, and covariance matrix $C_A$.}
\label{alg:overview}
\begin{algorithmic}[1]
\Procedure{\algoverviewname}{$N_A$, $C_A$}
\State $\sh{N_A}$ := \algexpandname($N_A$)
\State \textbf{return} \algevalname($\sh{N_A}$, $C_A$)
\EndProcedure
\end{algorithmic}
\end{algorithm}

\begin{lstlisting}[columns=fullflexible,caption={Problog code for the Burglary example with unique identifier for the random variables associated to the database, originally Example 6 in \protect\cite{Fierens2015}},label={lst:problogburglaryuniqueids},captionpos=b,float,numbers=left,
    stepnumber=1,
   mathescape=true]
$\omega_2$::burglary.
$\omega_3$::earthquake.
$\omega_4$::hears_alarm(john).
alarm :- burglary.
alarm :- earthquake.
calls(john) :- alarm, hears_alarm(john).
evidence(calls(john)).
query(burglary).
\end{lstlisting}

\begin{table}
    \centering
    \begin{tabu}{l l l}
    \textbf{Identifier} & \textbf{Beta parameters} & \textbf{Subjective Logic opinion}\\
    \toprule
$\omega_{1}$ & $\dbeta(\infty, \num{1})$ & $\langle 1.00, 0.00, 0.00, 0.50 \rangle$\\ 
 \dashedline
$\overline{\omega_{1}}$ & $\dbeta(\num{1}, \infty)$ & $\langle 0.00, 1.00, 0.00, 0.50 \rangle$\\ 
 \dashedline
$\omega_{2}$ & $\dbeta(\num{2}, \num{18})$ & $\langle 0.05, 0.85, 0.10, 0.50 \rangle$\\ 
 \dashedline
$\overline{\omega_{2}}$ & $\dbeta(\num{18}, \num{2})$ & $\langle 0.85, 0.05, 0.10, 0.50 \rangle$\\ 
 \dashedline
$\omega_{3}$ & $\dbeta(\num{2}, \num{8})$ & $\langle 0.10, 0.70, 0.20, 0.50 \rangle$\\ 
 \dashedline
$\overline{\omega_{3}}$ & $\dbeta(\num{8}, \num{2})$ & $\langle 0.70, 0.10, 0.20, 0.50 \rangle$\\ 
 \dashedline
$\omega_{4}$ & $\dbeta(\num{3.5}, \num{1.5})$ & $\langle 0.50, 0.10, 0.40, 0.50 \rangle$\\ 
 \dashedline
$\overline{\omega_{4}}$ & $\dbeta(\num{1.5}, \num{3.5})$ & $\langle 0.10, 0.50, 0.40, 0.50 \rangle$\\ 
\bottomrule
    \end{tabu}
    \caption{Associative table for the aProbLog code in Listing \ref{lst:problogburglaryuniqueids}}
    \label{tab:assoctable}
\end{table}

\newcommand{\shortv}[1]{\ensuremath{\sigma^2_{#1}}}
\begin{table}[!ht]
    \centering
    \begin{tabu}{X[1,c] | X[1,c] X[1,c] X[1,c] X[1,c] X[1,c] X[1,c] X[1,c] X[1,c] }
         & $\omega_{1}$ & $\overline{\omega_{1}}$ 
         & $\omega_{2}$ & $\overline{\omega_{2}}$ 
         & $\omega_{3}$ & $\overline{\omega_{3}}$ 
         & $\omega_{4}$ & $\overline{\omega_{4}}$ \\
         \hline
         $\omega_{1}$ & $\shortv{1}$ & $-\shortv{1}$\\
         $\overline{\omega_{1}}$ & $-\shortv{1}$ & $\shortv{1}$\\
         $\omega_{2}$ & & & $\shortv{2}$ & $-\shortv{2}$\\
         $\overline{\omega_{2}}$ & & & $-\shortv{2}$ & $\shortv{2}$\\
         $\omega_{3}$ & & & & & $\shortv{3}$ & $-\shortv{3}$\\
         $\overline{\omega_{3}}$ & & & & & $-\shortv{3}$ & $\shortv{3}$\\
         $\omega_{4}$ & & & & & &  & $\shortv{4}$ & $-\shortv{4}$\\
         $\overline{\omega_{4}}$ & & &  & & & & $-\shortv{4}$ & $\shortv{4}$
    \end{tabu}
    \caption{Covariance matrix for the associative table (Tab. \ref{tab:assoctable}) under the assumption that all the beta-distributed random variables are independent each other. We use a short-hand notation for clarity: $\shortv{i} = \cv[\omega_{i}]$. Zeros are omitted.}
    \label{tab:covariance}
\end{table}

\subsection{Circuit shadowing} 
\label{sec:shadowing}

We then augment the circuit adding \emph{shadow} nodes to superinpose a second circuit to enable the possibility to assess, in a single forward pass, both $p(query \land evidence)$ and $p(evidence)$. This can provide a benefit time-wise at the expense of memory, but more importantly it simplifies the bookkeeping of indexes in the covariance matrix as we will see below.

\begin{algorithm}[t]
  \caption[\textsc{Label}]{Shadowing the circuit $N_A$.}
\label{alg:expand}
\begin{algorithmic}[1]
\Procedure{\algexpandname}{$N_A$}
\State $\sh{N_A}$ := $N_A$
\State \emph{links} := \textsc{stack()}
\For{$p \in $ \textsc{parents}($N_A$, $\overline{\textsc{qnode}(N_A)}$)}
    \State \textsc{push}(\emph{links}, $\tuple{\overline{\textsc{qnode}(N_A)}, p}$)
\EndFor 
\While{$\lnot$\textsf{empty}(\emph{links)}}
    \State $\tuple{c, p}$ := \textsc{pop}(\emph{links})
    \State $\sh{N_A}$ := $\sh{N_A} \cup \set{\sh{c}}$
    \If{$\sh{p} \notin \sh{N_A}$}
        \State $\sh{N_A}$ := $\sh{N_A} \cup \set{\sh{p}}$
        \State \textsc{children}($\sh{N_A}$, $\sh{p}$) := \textsc{children}($\sh{N_A}$, $p$)
    \EndIf
    \State \textsc{children}($\sh{N_A}$, $\sh{p}$) := (\textsc{children}($\sh{N_A}$, $\sh{p}$)$\setminus \set{c})\cup\set{\sh{c}}$
    \For{$p' \in$ \textsc{parents}($N_A$, $p$)}
        \State \textsc{push}(\emph{links}, $\tuple{p, p'}$)
    \EndFor
\EndWhile
\State \textbf{return} $\sh{N_A}$
\EndProcedure
\end{algorithmic}
\end{algorithm}

\begin{figure}
    \centering
    \includegraphics[width=\textwidth]{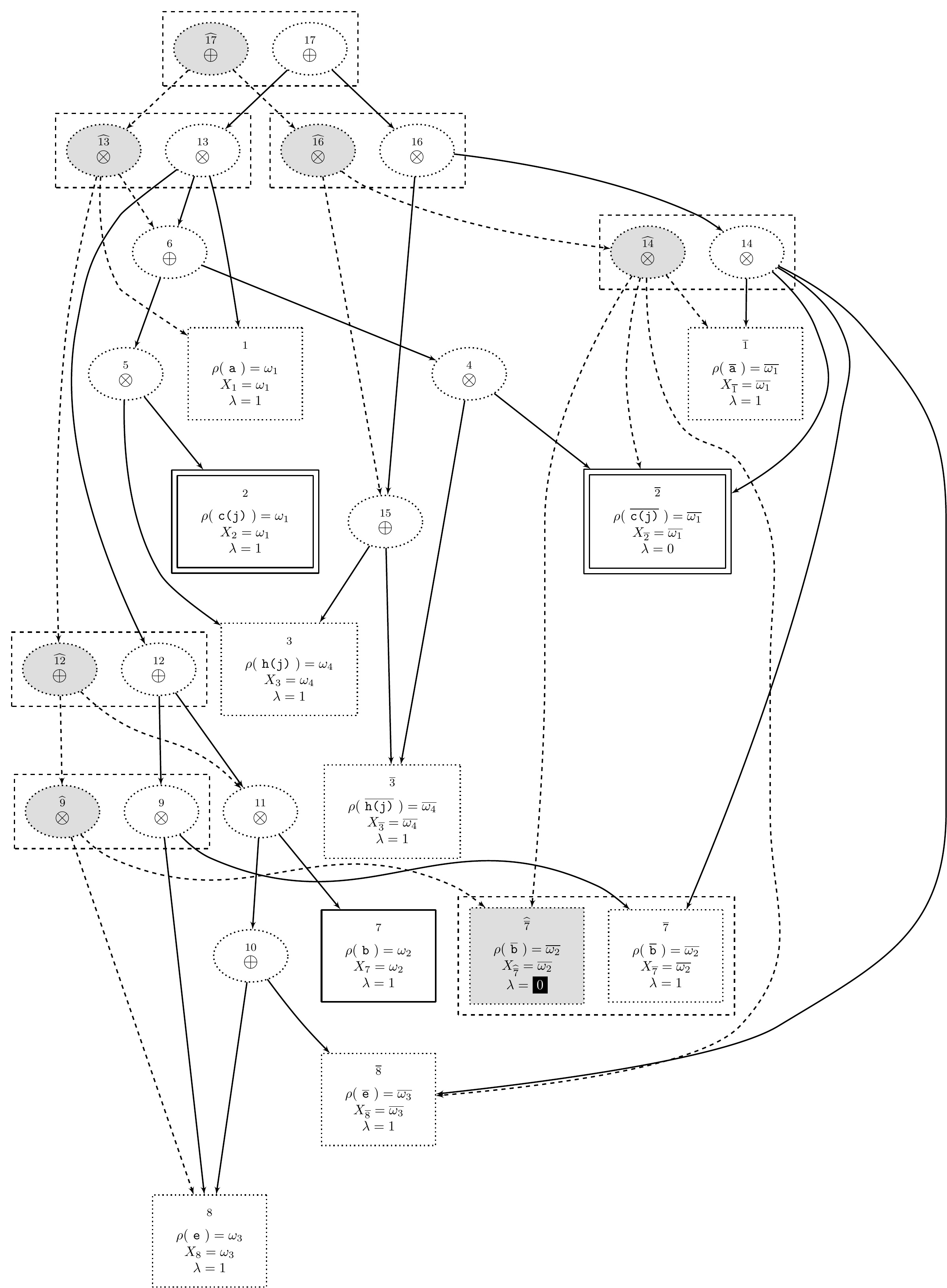}
    \caption{Shadowing of the circuit represented in Figure \ref{fig:burglarycircuit} according to Algorithm \ref{alg:expand}. Solid box for query, double box for evidence, in grey the \emph{shadow} nodes added to the circuit. If a node has a shadow, they are grouped together with a dashed box. Dashed arrows connect shadow nodes to their children.}
    \label{fig:burglarycircuits}
\end{figure}

Algorithm \ref{alg:expand} focuses on the node that identifies the negation of the query we want to evaluate with this circuit, identified as $\overline{\textsc{qnode}(N_A)}$):\footnote{In this paper we focus on a query composed by a single literal.} indeed, to evaluate $p(query \land evidence)$, the $\lambda$ parameter for such a node must be set to 0.
In lines 4--18 Algorithm \ref{alg:expand} superimpose a new circuit by creating \emph{shadow nodes}, e.g. $\sh{c}$ at line 9, that will represent random variables affected by the change in the $\lambda$ parameter for the $\overline{\textsc{qnode}(N_A)}$). The result of Algorithm \ref{alg:expand} on the circuit for our running example is depicted in Figure \ref{fig:burglarycircuits}.

In Algorithm \ref{alg:expand} we make use of a stack data structure with associated pop and push functions (cf. lines 3, 5, 8, 16): that is for ease of presentation as the algorithm does not require an ordered list.

\subsection{Evaluating the shadowed circuit} 
\label{sec:mainalgorithm}

Each of the nodes in the shadowed circuit (e.g. Figure \ref{fig:burglarycircuits}) has associated a (beta-distributed) random variable. In the following, and in Algorithm \ref{alg:evalshadow}, given a node $n$, its associated random variable is identified as \noderv{n}. For the nodes for which exists a $\labelfun$ label, its associated random variable is the beta-distributed random variable labelled via the $\labelfun$ function, cf. Figure \ref{fig:burglarycircuits}.

\newcommand{\vmeans}{\mbox{\emph{means}}}
\newcommand{\vcov}{\mbox{\emph{cov}}}
\newcommand{\ff}[1]{\mbox{\textsc{#1}}}
\newcommand{\tvar}{\mbox{\emph{tvar}}}
\newcommand{\nqueue}{\mbox{\emph{nqueue}}}
\newcommand{\nvisited}{\mbox{\emph{nvisited}}}

\begin{algorithm}[!h]
  \caption[\textsc{Label}]{Evaluating the shadowed circuit $\sh{N_A}$ taking into consideration the given $C_A$ covariance matrix.}
\label{alg:evalshadow}
\begin{algorithmic}[1]
\Procedure{\algevalname}{$\sh{N_A}, C_A$}
\State \vmeans\ := $\ff{zeros}(|\sh{N_A}|, 1)$
\State \vcov\ := $\ff{zeros}(|\sh{N_A}|, |\sh{N_A}|)$
\State \nvisited\ := $\set{\sh{\overline{\ff{qnode}(N_A)}}}$
\For{$n \in \ff{leaves}(\sh{N_A})\setminus \sh{\overline{\ff{qnode}(N_A)}}$}
    \State \nvisited\ := $\nvisited \cup \set{n}$
    \State \tvar\ := 0
    \If{$\lambda = 1$}
        \State{$\vmeans[n] := \ev[\noderv{n}]$}
        \State \tvar\ := \var{\noderv{n}}
    \State \textbf{else} $\vmeans[n] := 0$
    \EndIf
    \For{$n' \in \ff{leaves}(\sh{N_A})\setminus \sh{\overline{\ff{qnode}(N_A)}}$}
        \State $\vcov[n,n'] := C_A[X_n, X_{n'}]$
    \EndFor
\EndFor
\State \nqueue\ := $\sh{N_A}\setminus\nvisited$
\While{$\nqueue \neq \emptyset$}
    \State $n$ := $n \in \nqueue \text{ s.t. } \ff{children}(\sh{N_A}, n) \subseteq \nvisited$
    \State $\nqueue := \nqueue \setminus\set{n}$
    \State $\nvisited := \nvisited\cup\set{n}$
    \If{$n$ is a (shadowed) disjunction over $C := \ff{children}(\sh{N_A}, n)$}
        \State $\vmeans[n] := \sum_{c \in C} \vmeans[\noderv{c}]$
        \State $\vcov[n,n] := \sum_{c \in C} \sum_{c' \in C} \vcov[c,c']$
        \State $\vcov[z, n] := \vcov[n, z] := \sum_{c \in C} \vcov[c, z] ~\forall z \in \sh{N_A}\setminus\set{n}$
    \ElsIf{$n$ is a (shadowed) conjunction over $C := \ff{children}(\sh{N_A}, n)$}
        \State $\vmeans[n] := \prod_{c\in C} \vmeans[\noderv{c}]$
        \State $\vcov[n,n] := \sum_{c \in C} \sum_{c' \in C} \frac{\vmeans[\noderv{n}]^2}{\vmeans[\noderv{c}] \vmeans[\noderv{c'}]} \vcov[c, c']$
        \State $\vcov[z, n] := \vcov[n, z] := \sum_{c \in C} \frac{\vmeans[\noderv{n}]}{\vmeans[\noderv{c}]} \vcov[c, z]~ \forall z \in \sh{N_A}\setminus\set{n}$
    \EndIf
\EndWhile
\State $r := \ff{root}(\sh{N_A})$
\State \textbf{return} $\left\langle
\frac{\vmeans[r]}{\vmeans[\sh{r}]}, ~~
\frac{1}{\vmeans[\sh{r}]^2} \vcov[r,r]+
\frac{\vmeans[r]^2}{\vmeans[\sh{r}]^4} \vcov[\sh{r},\sh{r}] -
2 \frac{\vmeans[r]}{\vmeans[r]^3} \vcov[r, \sh{r}]
\right\rangle$
\EndProcedure
\end{algorithmic}
\end{algorithm}

Algorithm \ref{alg:evalshadow} begins with building a vector of means (\vmeans), and a matrix of covariances (\vcov) of the random variables associated to the leaves of the circuit (lines 2--16) derived from the $C_A$ covariance matrix provided as input. The algorithm can be made more robust by handling the case where $C_A$ is empty or a matrix of zeroes: in this case, assuming independence among the variables, it is straightforward to obtain a matrix such as Table \ref{tab:covariance}.

Then, Algorithm \ref{alg:evalshadow} proceeds to compute the means and covariances for all the remaining nodes in the circuit (lines 17--31). Here two cases arises.

Let $n$ be a $\oplus$-gate over $C$ nodes, its children: hence (lines 22--35)
\begin{eqnarray}
\ev[\noderv{n}] &=& \sum_{c\in C} \ev[\noderv{c}], \label{e:sp-meansum}\\
\cv[\noderv{n}] &=& \sum_{c\in C} \sum_{c' \in C} \cv[\noderv{c}, \noderv{c'}], \label{e:sumvar} \label{e:sp-varsum}\\
\cv[\noderv{n}, \noderv{z}] &=& \sum_{c \in C} \cv[\noderv{c}, \noderv{z}] \hspace{.1in} \mbox{for $z \in \sh{N_A}\setminus\set{n}$}
\label{e:sp-covsum}
\end{eqnarray}

\noindent
with
\begin{equation}
    \cv[X,Y] = \ev[X Y] - \ev[X]\ev[Y]
\end{equation}

\noindent
and $\cv[X] \equiv \cv[X,X] = \vv[X]$.

Let $n$ be a $\otimes$-gate over $C$ nodes, its children (lines 26--30).
Due to the nature of the variable $\noderv{n}$, 
following  \cite[\S 4.3.2]{Benaroya2005}
we perform a Taylor approximation. Let's assume $\displaystyle{\noderv{n} = \Pi(\bm{\noderv{C}}) = \prod_{c\in C} \noderv{c}}$, with $\bm{\noderv{C}} = \vt{\vp{\noderv{c_1}, \ldots, \noderv{c_k}}}$ and $k = |C|$. 
Expanding the first two terms yields:
\begin{equation}
 \label{e:multapprox}
    \def\arraystretch{3}
    \begin{array}{r c l}
         \noderv{n} & \simeq & \Pi(\ev[\bm{\noderv{C}}]) + \vt{(\bm{\noderv{C}} - \ev[\bm{\noderv{C}}])} \nabla \Pi(\bm{\noderv{C}}) \Bigr|_{\bm{\noderv{C}} = \ev[\bm{\noderv{C}}]}\\
         & = & \displaystyle{\ev[\noderv{n}] +
         (\noderv{\mathclap{~c_1}} - \ev[\noderv{c_1}]) \prod_{\mathclap{c \in C\setminus\set{c_1}}} \ev[\noderv{c}] + 
         \mbox{...} +
         (\noderv{\mathclap{~c_k}} - \ev[\noderv{c_k}]) \prod_{\mathclap{c \in C\setminus\set{c_k}}} \ev[\noderv{c}]
         }\\
         & = & \displaystyle{
         \ev[\noderv{n}] + \sum_{c \in C} \frac{\prod_{c'\in C} \ev[\noderv{c'}]}{\ev[\noderv{c}]} (\noderv{c} - \ev[\noderv{c}])
         }\\
         & = & \displaystyle{
         \ev[\noderv{n}] + \sum_{c \in C} \frac{ \ev[\noderv{n}]}{\ev[\noderv{c}]} (\noderv{c} - \ev[\noderv{c}])
         }\\
    \end{array}
\end{equation}
where
the first term can be seen as an approximation for $\ev[\noderv{n}]$.

Using this approximation, then (lines 34--42 of Algorithm \ref{alg:evalshadow})
\begin{eqnarray}
\cv[\noderv{n}] &\simeq& \sum_{c \in C} \sum_{c' \in C} \frac{\ev[\noderv{n}]^2}{\ev[\noderv{c}]\ev[\noderv{c'}]}
\cv[\noderv{c}, \noderv{c'}], \label{e:multvar}\label{e:sp-varprod}\\
\cv[\noderv{n}, \noderv{z}] &\simeq& \sum_{c \in C} \frac{\ev[\noderv{n}]}{\ev[\noderv{c}]} \cv[\noderv{c}, \noderv{z}] \hspace{.1in} \mbox{for $z \in \sh{N_A} \setminus \set{n}$}. \label{e:sp-covprod}
\end{eqnarray}

Finally, Algorithm \ref{alg:evalshadow} computes a conditioning between $\noderv{r}$ and $\noderv{\sh{r}}$, with $r$ being the root of the circuit ($r := \ff{root}(\sh{N_A})$ at line 46). This shows how critical is to keep track of the non-zero covariances where they exist. The Taylor series approximation of 
$\noderv{r}$ and $\displaystyle{\frac{1}{\noderv{\sh{r}}}}$ about $\ev[\noderv{r}]$ and $\displaystyle{\frac{1}{\ev[\noderv{\sh{r}}]}}$ leads to
\newcommand{\rrv}{\noderv{r}}
\newcommand{\srrv}{\noderv{\sh{r}}}
\newcommand{\errv}{\ev[\rrv]}
\newcommand{\esrrv}{\ev[\srrv]}
\begin{equation}
\frac{\rrv}{\srrv} \simeq \frac{\ev[\rrv]}{\ev[\srrv]} + \frac{1}{\esrrv} (\rrv-\errv) - \frac{\errv}{\esrrv^2} (\srrv-\esrrv),
\end{equation}
which implies
\begin{eqnarray}
\ev\left[\frac{\rrv}{\srrv}\right] &\simeq& \frac{\errv}{\esrrv}, \label{e:sp-meancond}\\
\cv\left[\frac{\rrv}{\srrv}\right] &\simeq& \frac{1}{\esrrv^2} \cv[\rrv] + \frac{\errv^2}{\esrrv^4} \cv[\srrv] - 2\frac{\errv}{\esrrv^3} \cv[\rrv, \srrv]. \label{e:sp-varcond}
\label{e:divvar}
\end{eqnarray}

Tables \ref{tab:means} and \ref{tab:covariances} depicts respectively the non-zero values of the \vmeans\ vector and \vcov\ matrix for our running example. Overall, the mean and variance for $p(\mtt{burglary} | \mtt{calls(john)})$ are $0.3571$ and $0.0528$, respectively. 
Figure \ref{fig:burglaryresult} depicts the resulting %obtained 
beta-distributed random variable (solid line) against a Monte Carlo simulation.

\begin{table}[]
    \centering
    \def\arraystretch{1.7}
    \setlength\dashlinedash{0.2pt}
\setlength\dashlinegap{1.5pt}
\setlength\arrayrulewidth{0.3pt}
    {
    \setlength{\tabcolsep}{5pt}
\begin{tabular}{c c c c c c c c c c c c c c }
& \noderv{5} & \noderv{6} & \noderv{\widehat{\overline{7}}} & \noderv{9} & \noderv{10} & \noderv{11} & \noderv{12} & \noderv{\widehat{12}} & \noderv{13} & \noderv{\widehat{13}} & \noderv{15} & \noderv{17} & \noderv{\widehat{17}} \\
\toprule
$\mu$ & \num{0.7} & \num{0.7} & \cellcolor{shadow} &\num{0.18} & \num{1} & \num{0.1} & \num{0.28} & \cellcolor{shadow}\num{0.1} & \num{0.196} & \cellcolor{shadow}\num{0.07} & \num{1} & \num{0.196} & \cellcolor{shadow}\num{0.07} \\
\bottomrule
\end{tabular}}
    \caption{Means as computed by Algorithm \ref{alg:evalshadow} on our running example. In grey the shadow nodes. Values very close or equal to zero are omitted. Also, values for nodes labelled with negated variables are omitted.  $\sh{\overline{7}}$, i.e. the shadow of \textsc{qnode}($N_A$), is included for illustration purpose.}
    \label{tab:means}
\end{table}

\begin{table}[]
    \centering
    {
    \newcolumntype{a}{>{\columncolor{shadow}}r}
    \def\arraystretch{2}
    \setlength{\tabcolsep}{4.5pt}
    \sisetup{
table-number-alignment = right,
table-figures-integer = 1,
table-figures-decimal = 2}
    \begin{tabular}{l r r r r a r r r r a r a r a }
 & \noderv{3} & \noderv{5} & \noderv{6} & \noderv{7} & \noderv{\widehat{\overline{7}}} & \noderv{8} & \noderv{9} & \noderv{11} & \noderv{12} & \noderv{\widehat{12}} & \noderv{13} & \noderv{\widehat{13}} & \noderv{17} & \noderv{\widehat{17}} \\
\toprule
\noderv{3} & 3.5 &3.5 &3.5 & &  &  &  &  &  &  & 1.0 &0.4 &1.0 &0.4\\
\hdashline
\noderv{5} & 3.5 &3.5 &3.5 & &  &  &  &  &  &  & 1.0 &0.4 &1.0 &0.4\\
\hdashline
\noderv{6} & 3.5 &3.5 &3.5 & &  &  &  &  &  &  & 1.0 &0.4 &1.0 &0.4\\
\hdashline
\noderv{7} &  &  &  & 0.4 & &  & -0.1 &0.4 &0.3 &0.4 &0.2 &0.3 &0.2 &0.3\\
\hdashline
\rowcolor{shadow}
\noderv{\widehat{\overline{7}}} &  &  &  &  &  &  &  &  &  &  &  &  &  &  \\
\hdashline
\noderv{8} &  &  &  &  &  & 1.5 &1.3 & & 1.3 & & 0.9 & & 0.9 & \\
\hdashline
\noderv{9} &  &  &  & -0.1 & & 1.3 &1.2 &-0.1 &1.1 &-0.1 &0.8 &-0.1 &0.8 &-0.1\\
\hdashline
\noderv{11} &  &  &  & 0.4 & &  & -0.1 &0.4 &0.3 &0.4 &0.2 &0.3 &0.2 &0.3\\
\hdashline
\noderv{12} &  &  &  & 0.3 & & 1.3 &1.1 &0.3 &1.5 &0.3 &1.0 &0.2 &1.0 &0.2\\
\hdashline
\rowcolor{shadow}
\noderv{\widehat{12}} &  &  &  & 0.4 & &  & -0.1 &0.4 &0.3 &0.4 &0.2 &0.3 &0.2 &0.3\\
\hdashline
\noderv{13} & 1.0 &1.0 &1.0 &0.2 & & 0.9 &0.8 &0.2 &1.0 &0.2 &1.0 &0.3 &1.0 &0.3\\
\hdashline
\rowcolor{shadow}
\noderv{\widehat{13}} & 0.4 &0.4 &0.4 &0.3 & &  & -0.1 &0.3 &0.2 &0.3 &0.3 &0.2 &0.3 &0.2\\
\hdashline
\noderv{17} & 1.0 &1.0 &1.0 &0.2 & & 0.9 &0.8 &0.2 &1.0 &0.2 &1.0 &0.3 &1.0 &0.3\\
\hdashline
\rowcolor{shadow}
\noderv{\widehat{17}} & 0.4 &0.4 &0.4 &0.3 & &  & -0.1 &0.3 &0.2 &0.3 &0.3 &0.2 &0.3 &0.2\\
\bottomrule
    \end{tabular}
    }
    \caption{Covariances ($\times 10^{-2}$) as computed by Algorithm \ref{alg:evalshadow} on our running example. In grey the shadow nodes. In grey the shadow nodes. Values very close or equal to zero are omitted. Also, values for nodes labelled with negated variables are omitted. $\sh{\overline{7}}$, i.e. the shadow of \textsc{qnode}($N_A$), is included for illustration purpose.}
    \label{tab:covariances}
\end{table}

\begin{figure}
    \centering
    \includegraphics[width=\textwidth]{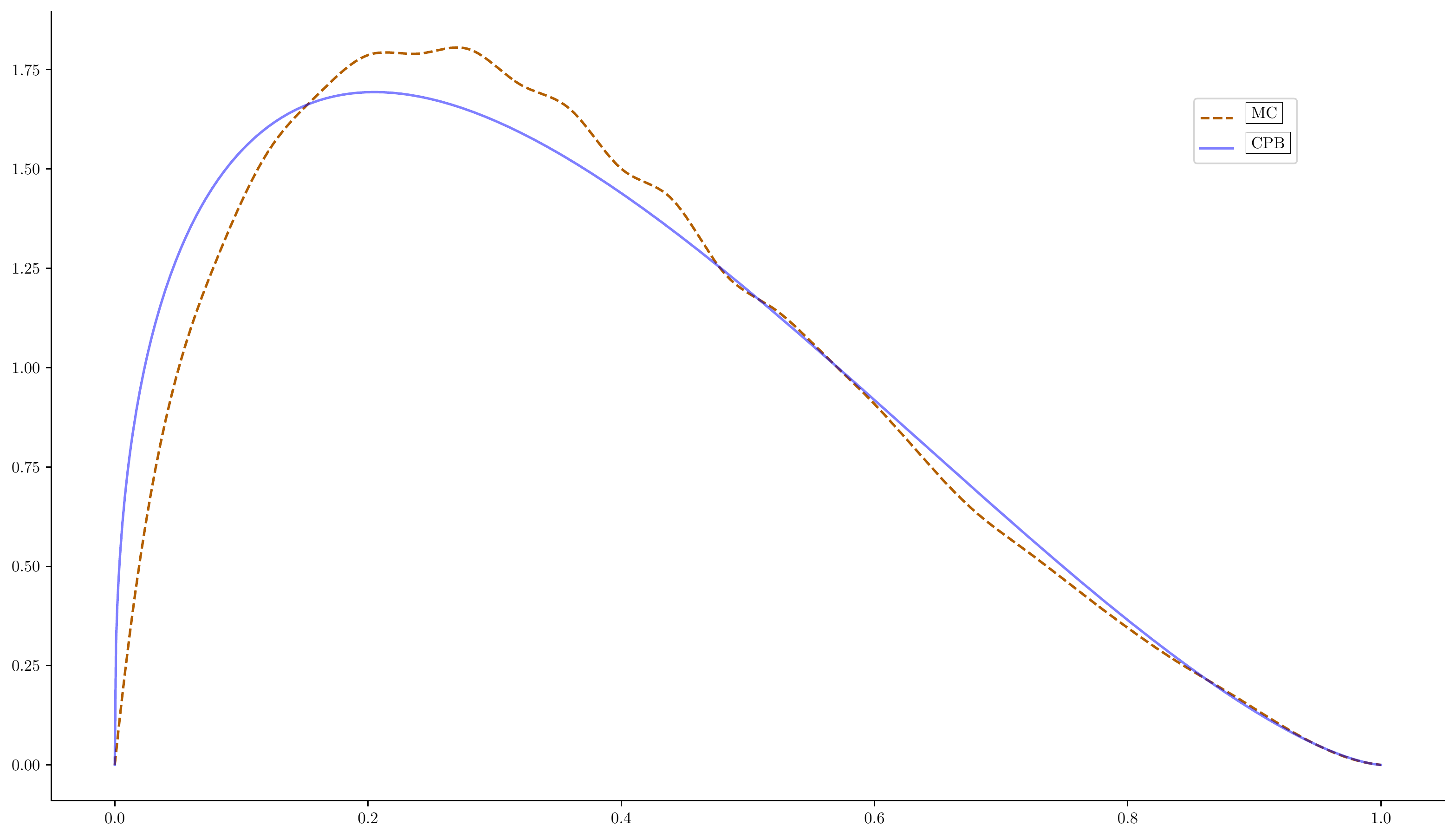}
    \caption{Resulting distribution of probabilities for our running example using Algorithm \ref{alg:overview} (solid line), and a Monte Carlo simulation with 100,000 samples grouped in 25 bins and then interpolated with a cubic polynomial (dashed line).}
    \label{fig:burglaryresult}
\end{figure}

\subsection{Memory performance}
\label{sec:memory-performance}

Algorithms~\ref{alg:overview} returns the mean and variance of the probability for the query conditioned on the evidence.  Algorithm~\ref{alg:expand} adds shadow nodes to the initial circuit formed by the evidence to avoid redundant computations in the second pass.  For the sake of clarity, Algorithm~\ref{alg:evalshadow} is presented in its most simple form. As formulated, it requires a $|\sh{N_A}| \times |\sh{N_A}|$ array to store the covariance values between the nodes. For large circuits, this memory requirement can significantly slow down the processing (e.g., disk swaps) or simply become prohibitive.  The covariances of a particular node are only required after it is computed via lines~24-25 or 34-35 in Algorithm~\ref {alg:evalshadow}.  Furthermore, these covariances are no longer needed once all the parent node values have been computed.  Thus, it is straightforward to dynamically allocate/de-allocate portions of the covariance array as needed. In fact, the selection of node $n$ to compute in line~19, which is currently arbitrary, can be designed to minimise processing time in light of the resident memory requirements for the covariance array.  Such an optimisation depends on the computing architecture and complicates the presentation.  Thus, further details are beyond the scope of this paper. 

\section{Experimental results}
\label{sec:experiment}

\subsection{The benefits of considering covariances}

To illustrate the benefits of Algorithm \ref{alg:overview} (Section \ref{sec:contribution}), we run an experimental analysis involving several circuits with unspecified labelling function. For each circuit, first labels are derived for the case of \amccond\ parametrisation $\semiringp$ \eqref{eq:semiringprobability} by selecting the ground truth probabilities from a uniform random distribution. Then, for each label, we derive a subjective opinion by observing $N_{ins}$ instantiations of a  random variables derived from the chosen probability, so to simulate data sparsity \cite{KAPLAN2018132}.

We then proceed analysing the inference on specific query nodes $\bm{q}$ in the presence of a set of evidence $\bm{E}=\bm{e}$ using:
\begin{itemize}
    \item \lBetaProblog\ as articulated in Section \ref{sec:contribution};
    \item \lHonolulu, cf. \eqref{eq:semiringsloperatorsbeta};
    \item \lJosang, cf. \eqref{eq:semiringsloperators};
    \item \lMonteCarlo, a Monte Carlo analysis with 100 samples from the derived random variables to obtain probabilities, and then computing the probability of queries in presence of evidence using the \amccond\ parametrisation $\semiringp$.
\end{itemize}

We then compare the RMSE to the actual ground truth. This process of inference to determine the marginal beta distributions is repeated 1000 times by considering 100 random choices for each label of the circuit, i.e. the ground truth,  and for each ground truth  10 repetitions of sampling the interpretations 
used to derive the subjective opinion labels observing $N_{ins}$ instantiations of all the variables. 

We judge the quality of the beta distributions of the queries on how well its expression of uncertainty captures the spread between its projected probability and the actual ground truth probability, as also \cite{KAPLAN2018132} did. In simulations where the ground truths are known, such as ours, confidence bounds  can be formed around the projected probabilities at a significance level of $\gamma$ and determine the fraction of cases when the ground truth falls within the bounds. If the uncertainty is well determined by the beta distributions, then this fraction should correspond to the strength $\gamma$ of the confidence interval \cite[Appendix C]{KAPLAN2018132}. 

\begin{lstlisting}[columns=fullflexible,caption={Smoker and Friends aProbLog code },label={lst:sf},captionpos=b,float,numbers=left,
    stepnumber=1,
   mathescape=true]
$\omega_1$::stress(X) :- person(X).
$\omega_1$::influences(X,Y) :- person(X), person(Y).
smokes(X) :- stress(X).
smokes(X) :- friend(X,Y), influences(Y,X), smokes(Y).
$\omega_3$::asthma(X) :- smokes(X).
person(1).
person(2).
person(3).
person(4).
friend(1,2).
friend(2,1).
friend(2,4).
friend(3,2).
friend(4,2).
evidence(smokes(2),true).
evidence(influences(4,2),false).
query(smokes(1)).
query(smokes(3)).
query(smokes(4)).
query(asthma(1)).
query(asthma(2)).
query(asthma(3)).
query(asthma(4)).
\end{lstlisting}

\begin{figure*}[t]
\centering
    \begin{subfigure}[b]{0.3\textwidth}
        \includegraphics[width=\textwidth]{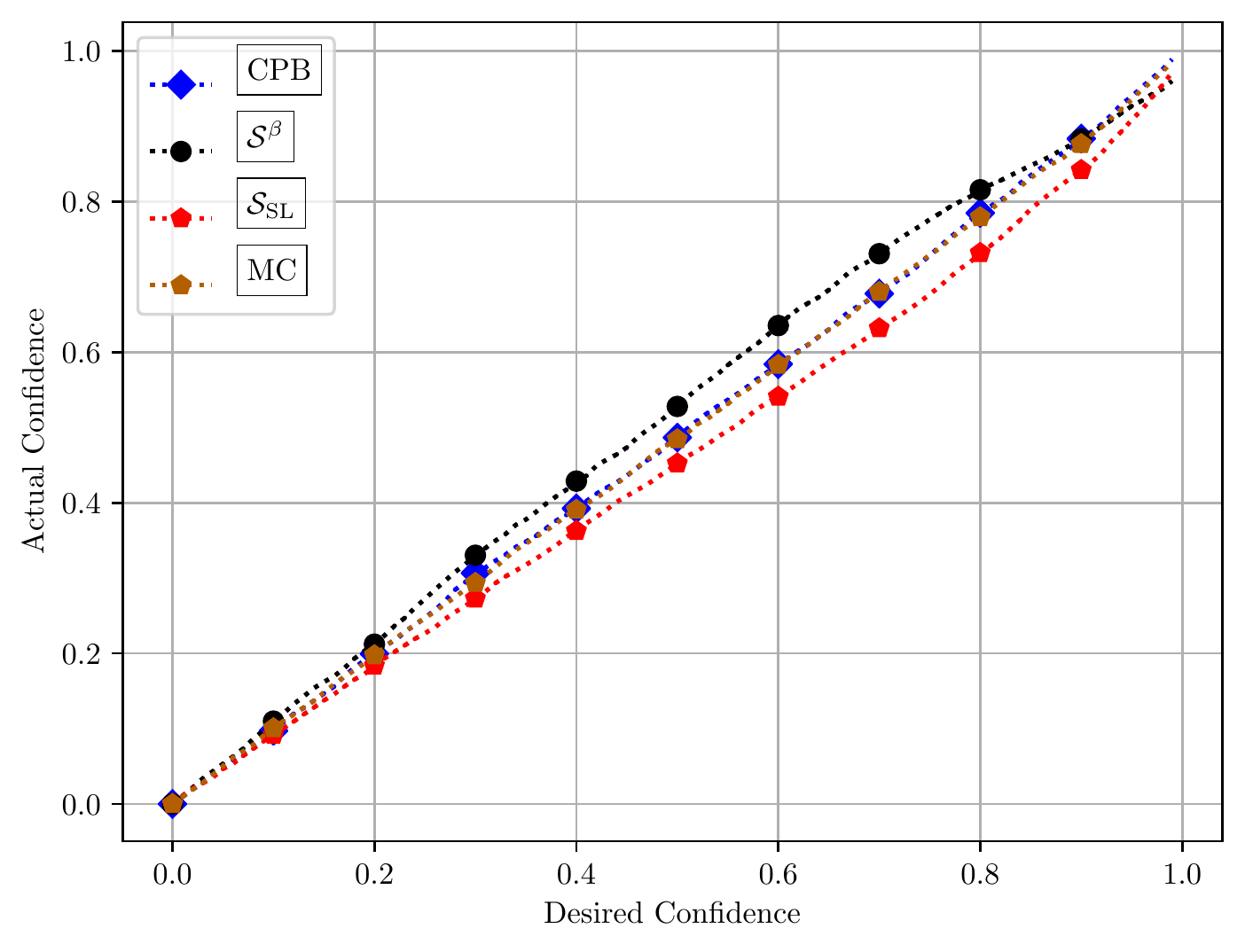}
        \caption{}
        \label{fig:sfa}
    \end{subfigure}
    ~
    \begin{subfigure}[b]{0.3\textwidth}
        \includegraphics[width=\textwidth]{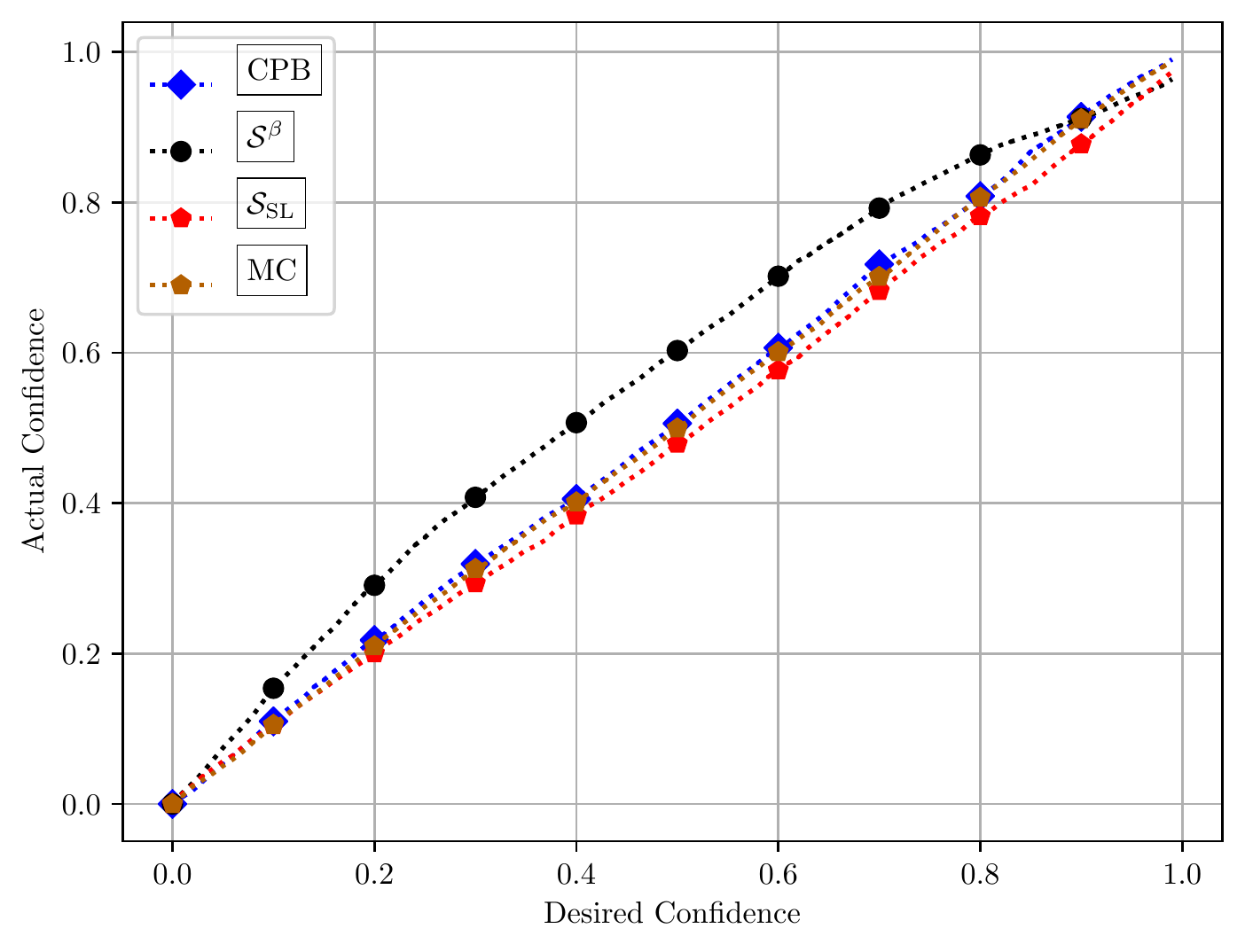}
        \caption{}
        \label{fig:sfb}
    \end{subfigure}
    ~
    \begin{subfigure}[b]{0.3\textwidth}
        \includegraphics[width=\textwidth]{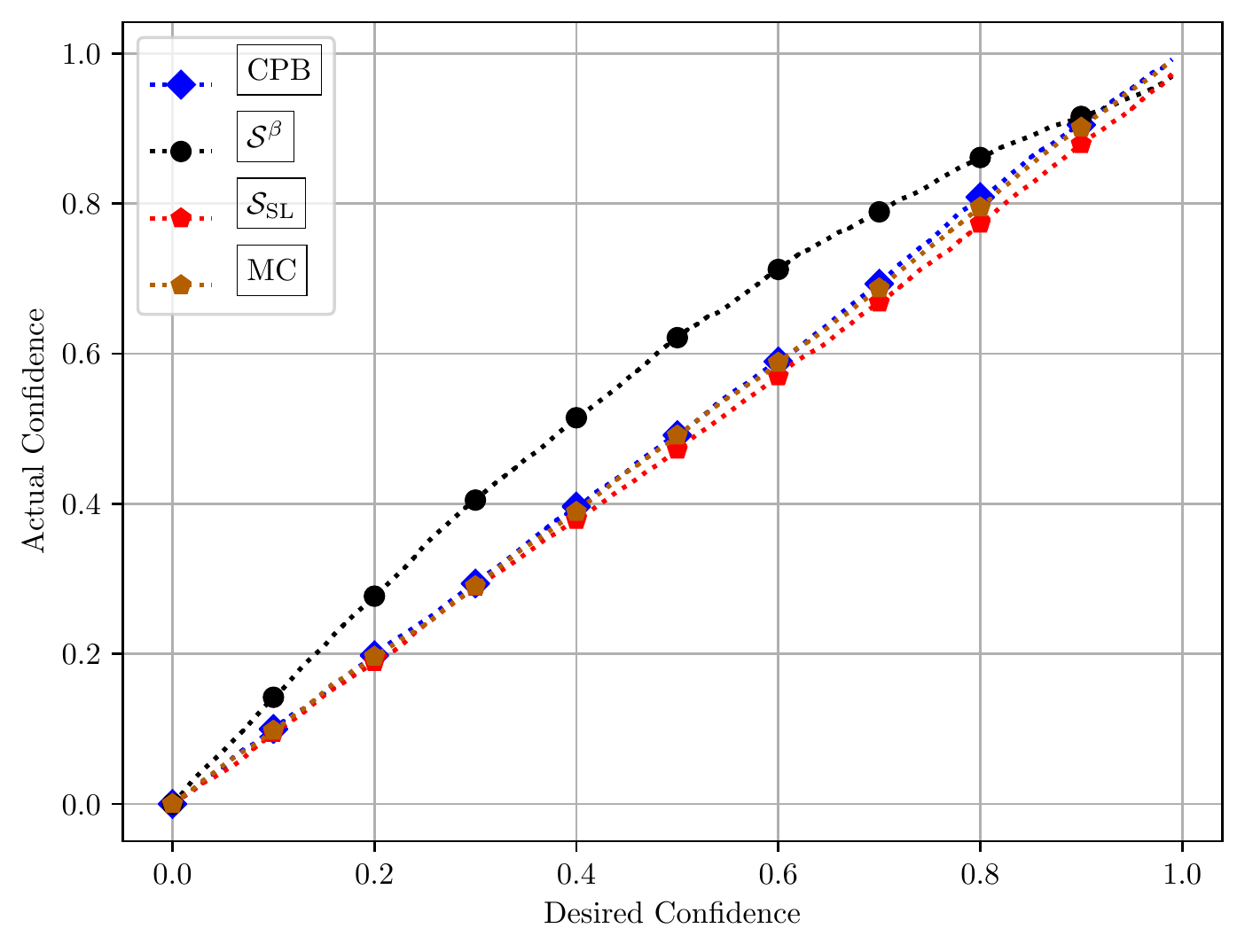}
        \caption{}
        \label{fig:sfc}
    \end{subfigure}
    
    \caption{Actual versus desired significance of bounds derived from the uncertainty for Smokers \& Friends with: (\subref{fig:sfa}) $N_{ins} = 10$; (\subref{fig:sfb} $N_{ins} = 50$; and (\subref{fig:sfc}) $N_{ins} = 100$. Best closest to the diagonal.  Monte Carlo approach has been run over 100 samples.}
    \label{fig:sf}
\end{figure*}

\begin{figure*}[t]
\centering
    \begin{subfigure}[b]{0.3\textwidth}
        \includegraphics[width=\textwidth]{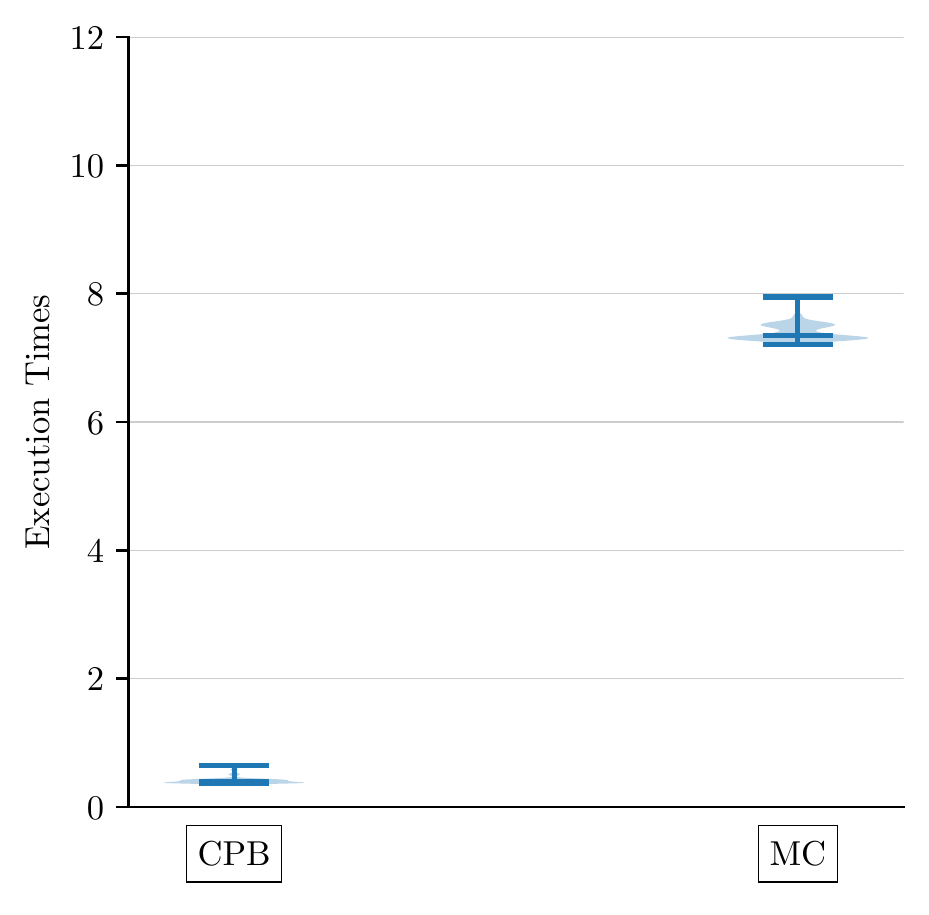}
        \caption{}
        \label{fig:sfatimes}
    \end{subfigure}
    ~
    \begin{subfigure}[b]{0.3\textwidth}
        \includegraphics[width=\textwidth]{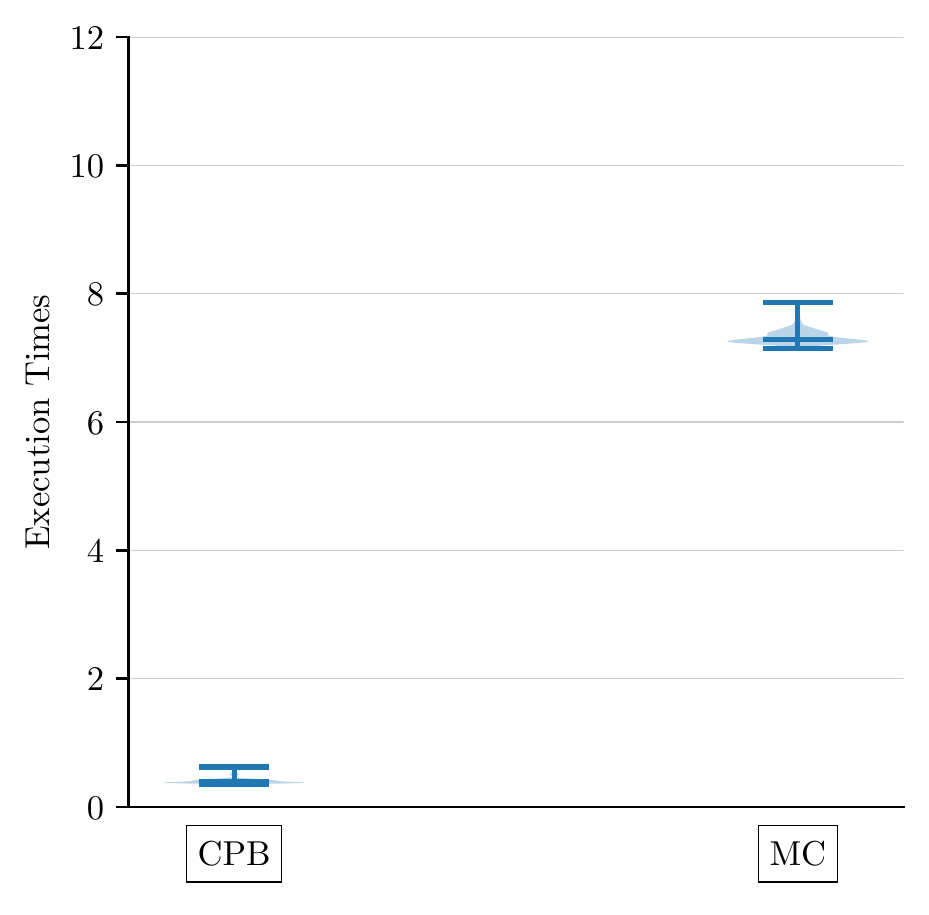}
        \caption{}
        \label{fig:sfbtimes}
    \end{subfigure}
    ~
    \begin{subfigure}[b]{0.3\textwidth}
        \includegraphics[width=\textwidth]{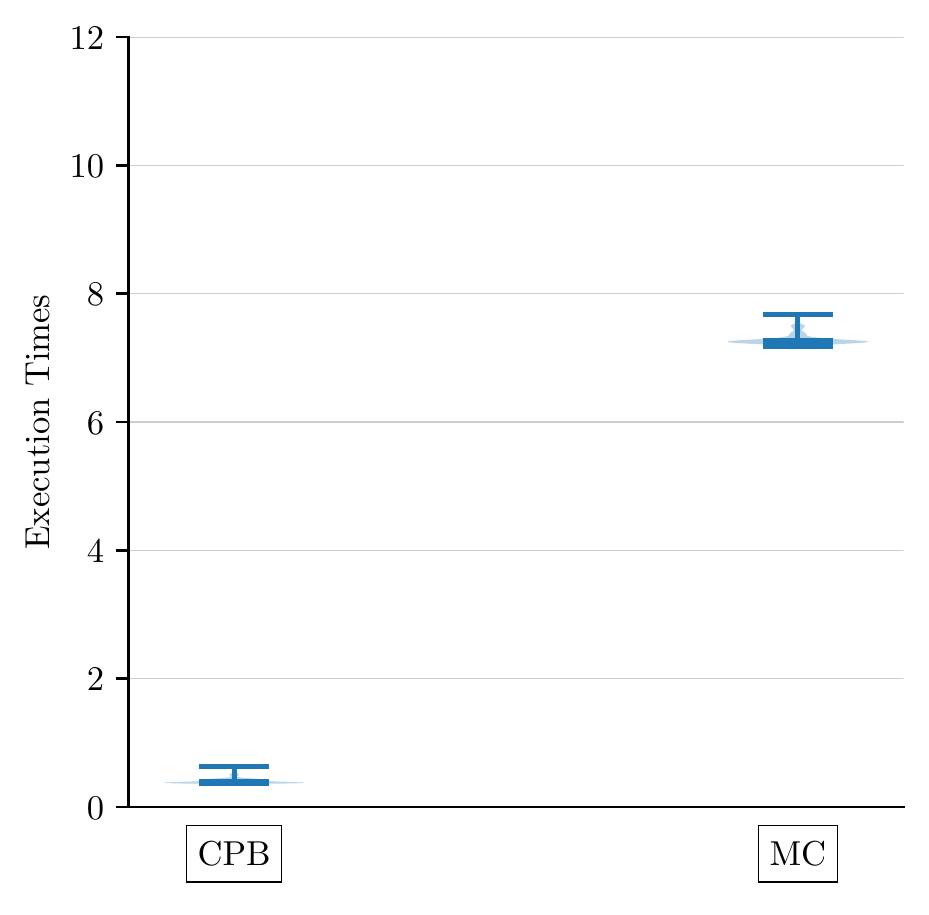}
        \caption{}
        \label{fig:sfctimes}
    \end{subfigure}
    
    \caption{Distribution of execution time for running the different algorithms for Smokers \& Friends with: (\subref{fig:sfatimes}) $N_{ins} = 10$; (\subref{fig:sfbtimes}) $N_{ins} = 50$; and (\subref{fig:sfctimes}) $N_{ins} = 100$. Best lowest.  Monte Carlo approach has been run over 100 samples.}
    \label{fig:sftimes}
\end{figure*}

\begin{figure*}[t]
\centering
    \begin{subfigure}[b]{0.3\textwidth}
        \includegraphics[width=\textwidth]{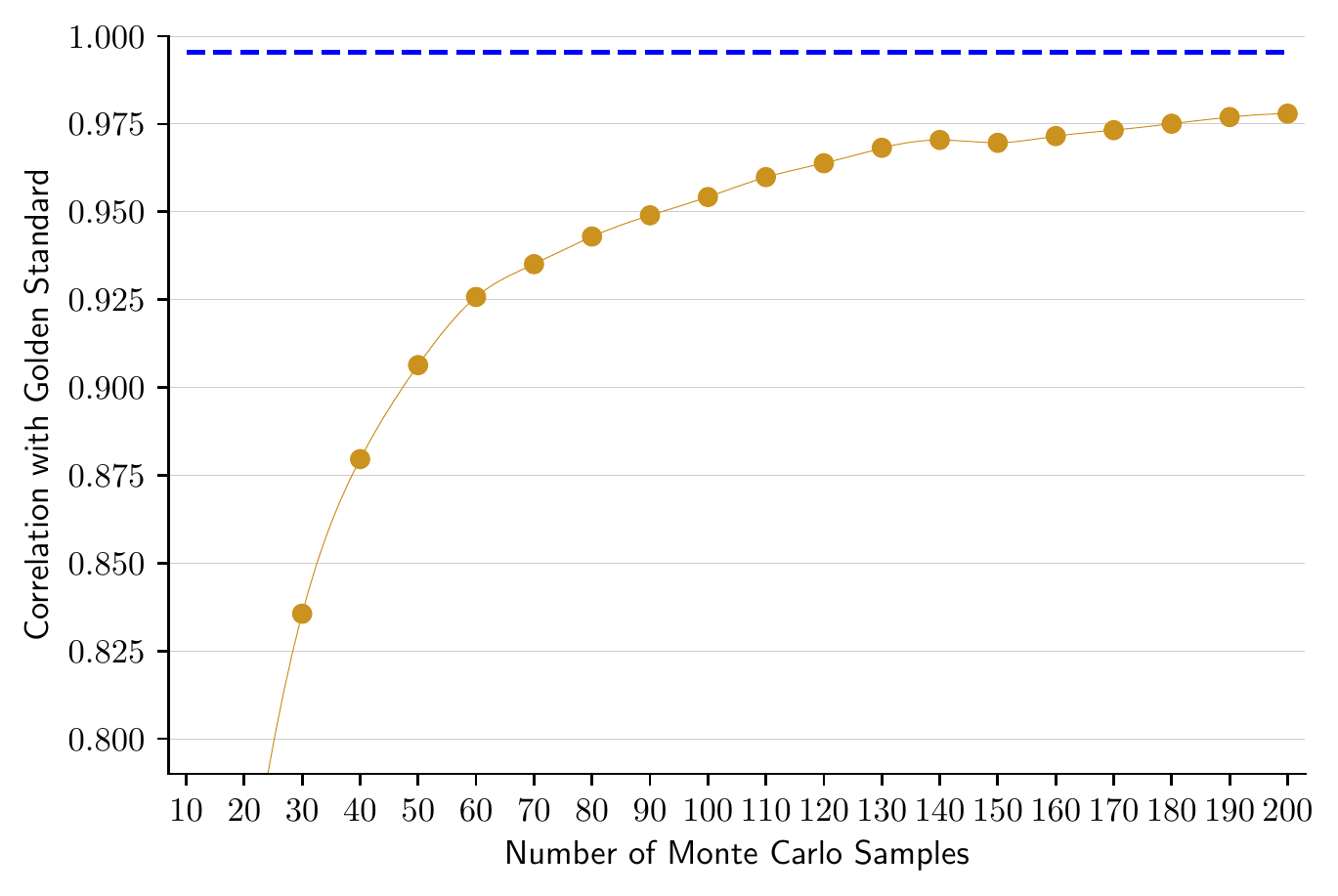}
        \caption{}
        \label{fig:sfacorr}
    \end{subfigure}
    ~
    \begin{subfigure}[b]{0.3\textwidth}
        \includegraphics[width=\textwidth]{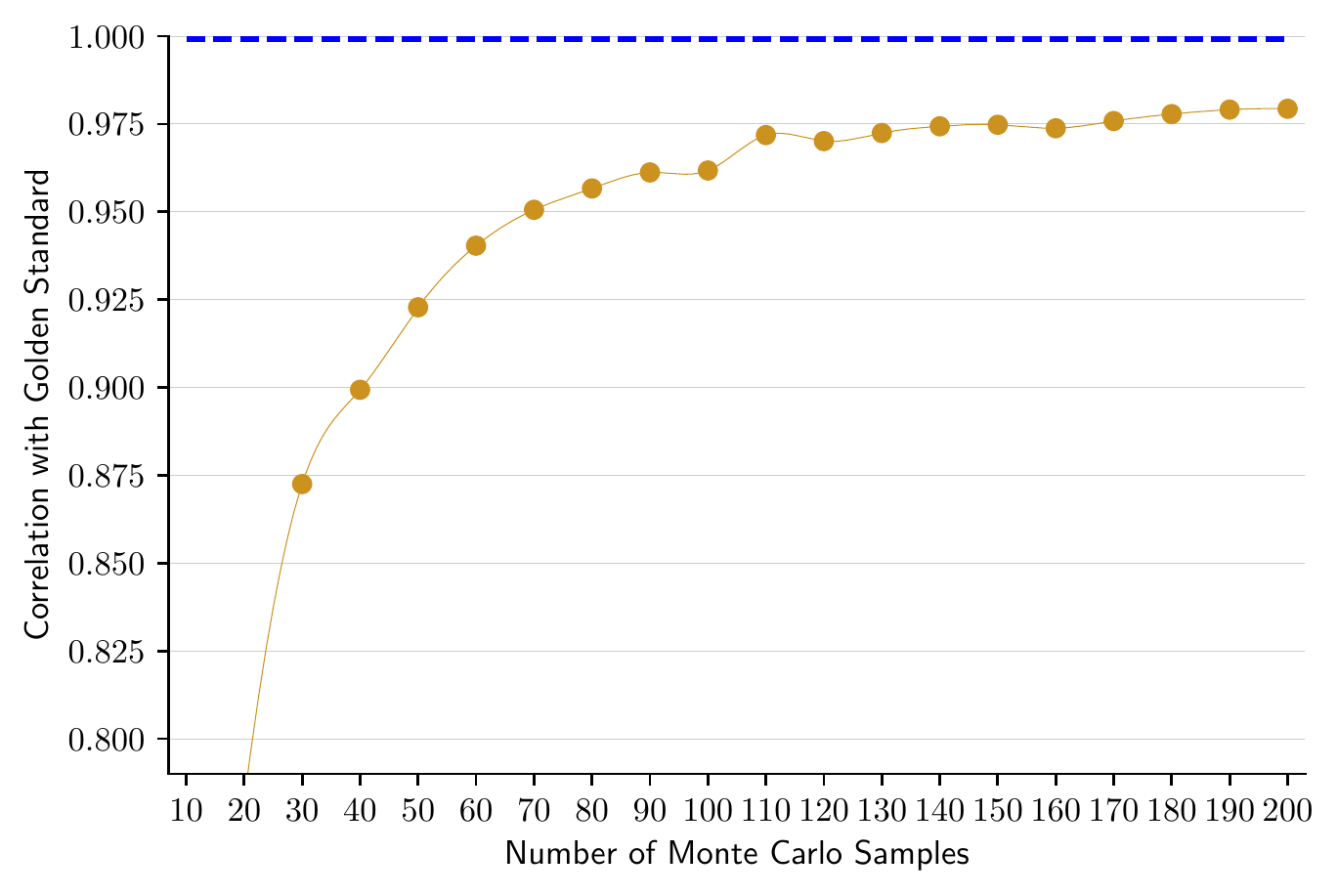}
        \caption{}
        \label{fig:sfbcorr}
    \end{subfigure}
    ~
    \begin{subfigure}[b]{0.3\textwidth}
        \includegraphics[width=\textwidth]{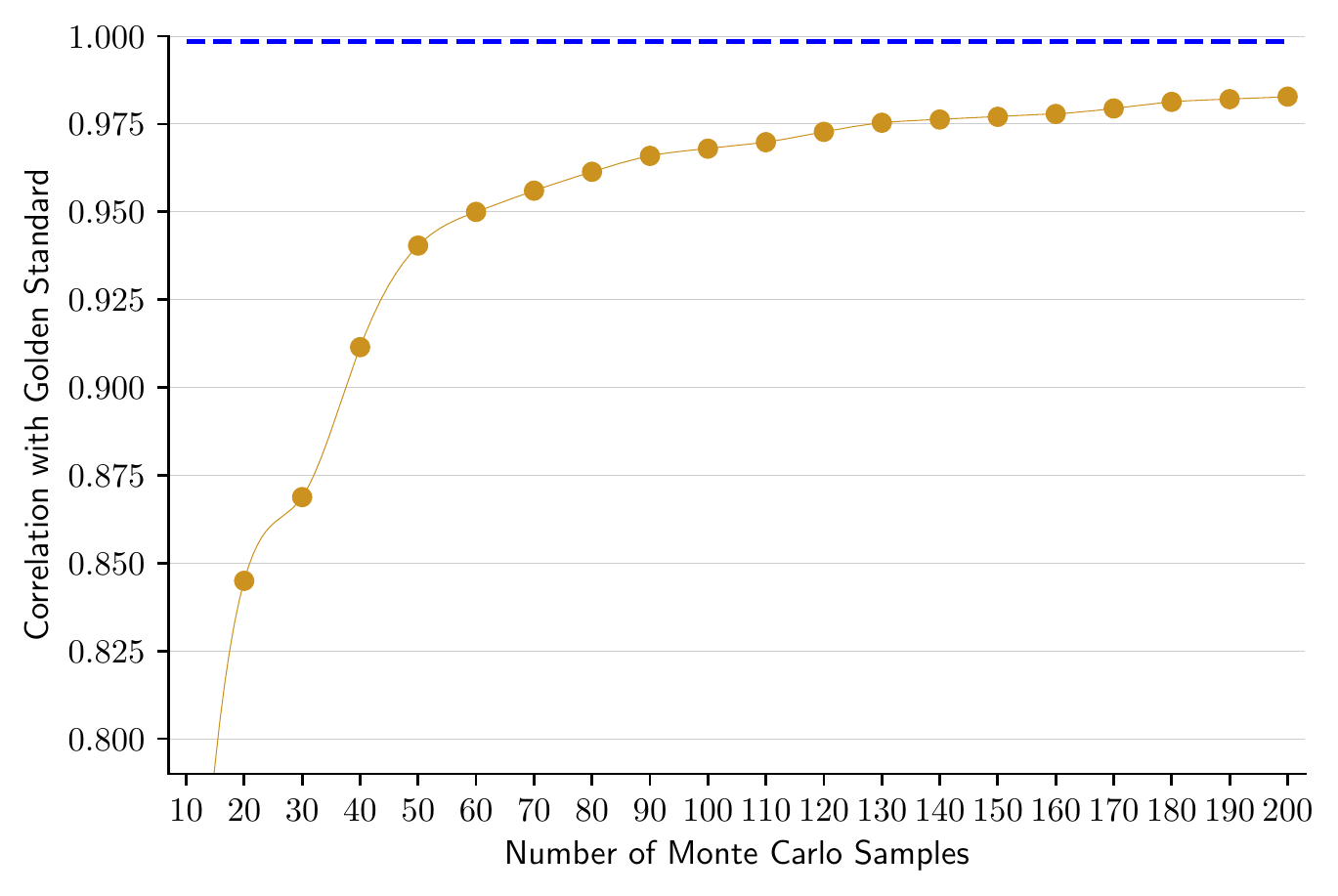}
        \caption{}
        \label{fig:sfccorr}
    \end{subfigure}
    
    \caption{Correlation of Dirichlet strengths between runs of Monte Carlo approach varying the number of samples and golden standard (i.e. a Monte Carlo run with 10,000 samples) as well as between the proposed approach and golden standard with cubic interpolation---that is independent of the number of samples used in Monte Carlo ---for Smokers \& Friends with: (\subref{fig:sfacorr}) $N_{ins} = 10$; (\subref{fig:sfbcorr}) $N_{ins} = 50$; and (\subref{fig:sfccorr}) $N_{ins} = 100$.}
    \label{fig:sfcorr}
\end{figure*}

Following \cite{DBLP:conf/aaai/CeruttiKKS19}, we consider the famous Friends \& Smokers problem, cf. Listing \ref{lst:sf},\footnote{\url{https://dtai.cs.kuleuven.be/problog/tutorial/basic/05_smokers.html} (on 29th April 2020).} with fixed queries and set of evidence. 
Table \ref{tab:smokers} provides the root mean square error (RMSE) between the projected probabilities and the ground truth probabilities for all the inferred query variables for $N_{ins}$ = 10, 50, 100. The table also includes the predicted RMSE by taking the square root of the average---over the number of runs---variances from the inferred marginal beta distributions, cf. \eqref{e:pred_var}. Figure \ref{fig:sf} plots the desired and actual significance levels for the confidence intervals (best closest to the diagonal), i.e. the fractions of times the ground truth falls within confidence bounds set to capture x\% of the data. Figure \ref{fig:sftimes}  depicts the distribution of execution time for running the various algorithm over this dataset: for each circuit, all algorithms have been computed on a Intel(R) Xeon(R) Skylake Quad-core \@ 2.30GHz and 128 GB of RAM.
Finally, Figure \ref{fig:sfcorr} depicts the correlation of Dirichlet strengths between \lMonteCarlo\ runs with variable number of samples and the golden standard (i.e. a Monte Carlo run with 10,000 samples), as well as between \lbetaprobcov\ and the golden standard, which is clearly independent of the number of samples used for \lMonteCarlo. Given $\bm{X}_{\bm{q}}^g$ (resp. $\bm{X}_{\bm{q}}$) the random variable associated to the queries $\bm{q}$ computed using the golden standard (resp. computed using either \lMonteCarlo\ or \lbetaprobcov), the Pearson's correlation coefficient displayed in Figure \ref{fig:sfcorr} is given by:
\begin{equation}
    r = \frac{
        \cv[\bm{s}_{\bm{X}_{\bm{q}}^g}, \bm{s}_{\bm{X}_{\bm{q}}}]
        }
        {\cv[\bm{s}_{\bm{X}_{\bm{q}}^g}] \cv[\bm{s}_{\bm{X}_{\bm{q}}}]}
\end{equation}

\noindent
This is a measure of the quality of the epistemic uncertainty associated with the evaluation of the circuit using \lMonteCarlo\ with varying number of samples, and \lbetaprobcov: the closer the Dirichlet strengths are to those of the golden standard, the better the computed epistemic uncertainty represents the actual uncertainty,\footnote{The Dirichlet strengths are inversely proportional to the epistemic uncertainty.}  
hence the closer the correlations are to 1 in Figure \ref{fig:sfcorr} the better.

From Table \ref{tab:smokers}, \lbetaprobcov\ exhibits the lowest RMSE and the best prediction of its own RMSE. As already noticed in \cite{DBLP:conf/aaai/CeruttiKKS19}, \lHonolulu\ is a little conservative in estimating its own RMSE, while \lJosang\ is overconfident. This is reflected in Figure \ref{fig:sf}, with the results of \lHonolulu\ being over the diagonal, and those of \lJosang\ being below it, while \lbetaprobcov\ sits exactly on the diagonal, like also \lMonteCarlo. However, \lMonteCarlo\ with 100 samples does not exhibit the lowest RMSE according to Table \ref{tab:smokers}, although the difference with the best one is much lower compared with \lJosang. 

Considering the execution time, Figure \ref{fig:sftimes}, we can see that there is a substantial difference between \lbetaprobcov\ and \lMonteCarlo\ with 100 samples.

Finally, Figure \ref{fig:sfcorr} depicts the correlation of the Dirichlet strength between the golden standard, i.e. a Monte Carlo simulation with 10,000 samples, and both \lbetaprobcov\ and \lMonteCarlo, this last one varying the number of samples used. It is straightforward to see that \lMonteCarlo\ improves the accuracy of the computed epistemic uncertainty when increasing the number of samples considered, approaching the same level of \lbetaprobcov\ when considering more than 200 samples.

\begin{table}
    \small
    \centering
    \begin{tabu}{X[2,l] l l >{\leavevmode\color{cBetaProblog}}X[4,r] >{\leavevmode\color{cHonolulu}}X[4,r] >{\leavevmode\color{cJosang}}X[4,r]
>{\leavevmode\color{cMonteCarlo}}X[4,r]}
    \toprule
 & $N_{ins}$ &   & $\lBetaProblog$ & $\lHonolulu$ & $\lJosang$    & \lMonteCarlo\\
    \midrule
    \multirow[t]{ 6}{=}{\setlength\parskip{\baselineskip}Friends \newline{} \& Smokers}
    & 10 & A & \best{0.1065} & 0.1065 & 0.1198 & 0.1072 \\
    &  & P & 0.1024 & 0.1412 & 0.1060 & 0.1027 \\
    & 50 & A & \best{0.0489} & 0.0489 & 0.0617 & 0.0490 \\
    &  & P & 0.0491 & 0.0898 & 0.0587 & 0.0489 \\
    & 100 & A & \best{0.0354} & 0.0354 & 0.0521 & 0.0355 \\
    &  & P & 0.0357 & 0.0709 & 0.0487 & 0.0356 \\
    \bottomrule
    \end{tabu}
    \caption{RMSE for the queried variables in the Friends \& Smokers program: A stands for Actual, P for Predicted. Best results---also considering hidden decimals---for the actual RMSE boxed. Monte Carlo approach has been run over 100 samples.}
    \label{tab:smokers}
\end{table}

\subsection{Comparison with other approaches for dealing with uncertain probabilities}
\label{sec:bnexp}

\begin{lstlisting}[columns=fullflexible,caption={An example of aProblog code that can be seen also as a Bayesian network, cf. Fig. \ref{fig:net1} in Appendix \ref{sec:bn}. $e_i$ are randomly assigned as either \texttt{True} or \texttt{False}.},label={lst:net1},captionpos=b,float,numbers=left,
    stepnumber=1,
   mathescape=true]
$\omega_{2}$::n1.
$\omega_{3}$::n2 :- \+n1.
$\omega_{4}$::n2 :- n1.
$\omega_{5}$::n3 :- \+n2.
$\omega_{6}$::n3 :- n2.
$\omega_{7}$::n4 :- \+n2.
$\omega_{8}$::n4 :- n2.
$\omega_{9}$::n5 :- \+n3.
$\omega_{10}$::n5 :- n3.
$\omega_{11}$::n6 :- \+n3.
$\omega_{12}$::n6 :- n3.
$\omega_{13}$::n7 :- \+n6.
$\omega_{14}$::n7 :- n6.
$\omega_{15}$::n8 :- \+n5.
$\omega_{16}$::n8 :- n5.
$\omega_{17}$::n9 :- \+n5.
$\omega_{18}$::n9 :- n5.
evidence(n1, $e_1$).
evidence(n4, $e_2$).
evidence(n7, $e_3$).
evidence(n8, $e_4$).
evidence(n9, $e_5$).
query(n2).
query(n3).
query(n5).
query(n6).
\end{lstlisting}

To compare our approach against the state-of-the-art approaches for reasoning with uncertain probabilities, following \cite{DBLP:conf/aaai/CeruttiKKS19} we restrict ourselves to the case of circuits representing inferences over a Bayesian network. For instance, Listing \ref{lst:net1} shows an aProblog code that can also be interpreted as a Bayesian network. We considered three circuits and their Bayesian network representation: Net1 (Listing \ref{lst:net1}); Net2; and Net3. Figure \ref{fig:nets} in Appendix \ref{sec:bn} depicts the Bayesian networks that can be derived from such circuits. In the following, we will refer to NetX as both the circuit and the Bayesian network without distinction. 
We then compared \lbetaprobcov\ against three approaches specifically designed for dealing with uncertain probabilities in Bayesian networks: Subjective Bayesian Networks; Belief Networks; and Credal Networks.

\begin{description}

\item[\emph{Subjective Bayesian Network} \lSBN] \cite{ivanovska.15,kaplan.16.fusion,KAPLAN2018132},
was first proposed in \cite{ivanovska.15}, and it is an uncertain Bayesian network where the conditionals are subjective opinions instead of dogmatic probabilities. In other words, the conditional probabilities are known within a beta distribution. \lSBN\ uses subjective belief propagation (SBP), which was introduced for trees in \cite{kaplan.16.fusion} and extended for singly-connected networks in \cite{KAPLAN2018132}, that extends the Belief Propagation (BP) inference method of  \cite{pearl.86}.  In BP, $\pi$- and $\lambda$-messages are passed from parents and children, respectively, to a node, i.e., variable.  The node uses these messages to formulate the inferred marginal probability of the corresponding variable.  The node also uses these messages to determine the $\pi$- and $\lambda$-messages to send to its children and parents, respectively.  In SBP, the $\pi$- and $\lambda$-messages are subjective opinions characterised by a projected probability and Dirichlet strength. The SBP formulation approximates output messages as beta-distributed random variables using the methods of moments and a first-order Taylor series approximation to determine the mean and variance of the output messages in light of the beta-distributed input messages. The details of the derivations are provided in \cite{kaplan.16.fusion,KAPLAN2018132}.  

\item[\emph{Belief Networks} \lGBT] \cite{Smets2005} introduced a computationally efficient method to reason over networks via Dempster-Shafer theory \cite{DEMPSTER68}.  It is an approximation of a valuation-based system.  Namely, a (conditional) subjective opinion $\omega_X=[b_x, b_{\bar{x}}, u_X]$  from our circuit obtained from data is converted to the following belief mass assignment:
$m(x) = b_{x}$, $m(\bar{x})  = b_{\bar{x}}$ and $m(x \cup \bar{x}) = u_{X}$. Note that in the binary case, the belief function overlaps with the belief mass assignment. The method exploits the disjunctive rule of combination to compose beliefs conditioned on the Cartesian product space of the binary power sets. This enables both forward propagation and backward propagation after inverting the belief conditionals via the generalized Bayes'  theorem (GBT). By operating in the Cartesian product space of the binary power sets, the computational complexity grows exponentially with respect to the number of parents.

\item[\emph{Credal Networks} \lCredal] \cite{credal98}.
A credal network over binary random variables extends a Bayesian network by replacing single probability values with closed intervals representing the possible range of probability values. The extension of Pearl's message-passing algorithm  by the 2U algorithm for credal networks is described in \cite{credal98}. This algorithm  works by  determining the maximum and minimum value (an interval) for each of the target probabilities based on the given input intervals.  It turns out that these extreme values lie at the vertices of the polytope dictated by the extreme values of the input intervals.  As a result, the computational complexity grows exponentially with respect to the number of parents nodes. For the sake of comparison, we assume that the random variables we label our circuts with and elicited from the given data corresponds to a credal network in the following way: if $\omega_x=[b_x, b_{\bar{x}}, u_X]$ is a subjective opinion on the probability $p_x$, then we have $[b_x, b_x+u_X]$ as an interval corresponding to this probability in the credal network. It should be noted that this mapping from the beta-distributed random variables to an interval is consistent with past studies of credal networks~\cite{karlsson.08}.
\end{description}

\begin{table}
\small
\begin{tabu}{l l l >{\leavevmode\color{cBetaProblog}}X[3,r] >{\leavevmode\color{cHonolulu}}X[3,r] >{\leavevmode\color{cJosang}}X[3,r]
>{\leavevmode\color{cMonteCarlo}}X[3,r]
>{\leavevmode\color{cSBN}}X[3,r] >{\leavevmode\color{cGBT}}X[3,r] >{\leavevmode\color{cCredal}}X[3,r]}
\toprule
 & $N_{ins}$ &   & \lbetaprobcov & \lHonolulu & \lJosang    & \lMonteCarlo & \lSBN & \lGBT    & \lCredal\\
\midrule
\netone{} & 10 & A & 0.1511 & \best{0.1511} & 0.2078 & 0.1517 & 0.1511 & 0.1542 & 0.1633 \\
 &  & P & 0.1473 & 0.1864 & 0.1559 & 0.1465 & 0.1472 & 0.0873 & 0.2009 \\
\netone{} & 50 & A & 0.0816 & \best{0.0816} & 0.1237 & 0.0818 & 0.0816 & 0.0848 & 0.0827 \\
 &  & P & 0.0802 & 0.1227 & 0.0825 & 0.0789 & 0.0794 & 0.0372 & 0.1069 \\
\netone{} & 100 & A & \best{0.0544} & 0.0544 & 0.0837 & 0.0550 & 0.0544 & 0.0601 & 0.0557 \\
 &  & P & 0.0572 & 0.0971 & 0.0592 & 0.0564 & 0.0566 & 0.0262 & 0.0766 \\
\dashedline
\nettwo{} & 10 & A & 0.1389 & \best{0.1389} & 0.1916 & 0.1392 & 0.1389 & 0.1418 & 0.1473 \\
 &  & P & 0.1391 & 0.1808 & 0.1457 & 0.1381 & 0.1399 & 0.1058 & 0.1856 \\
\nettwo{} & 50 & A & \best{0.0701} & \best{0.0701} & 0.1092 & 0.0702 & \best{0.0701} & 0.0730 & 0.0702 \\
 &  & P & 0.0722 & 0.1148 & 0.0755 & 0.0714 & 0.0720 & 0.0486 & 0.0952 \\
\nettwo{} & 100 & A & \best{0.0534} & 0.0534 & 0.0901 & 0.0536 & 0.0534 & 0.0553 & 0.0537 \\
 &  & P & 0.0533 & 0.0937 & 0.0601 & 0.0526 & 0.0531 & 0.0340 & 0.0696 \\
\dashedline
\netthree{} & 10 & A & \best{0.1481} & 0.1481 & 0.2160 & 0.1488 & 0.1481 & 0.1511 & 0.1634 \\
 &  & P & 0.1453 & 0.1708 & 0.1578 & 0.1438 & 0.1454 & 0.0821 & 0.1947 \\
\netthree{} & 50 & A & 0.0737 & 0.0737 & 0.1167 & 0.0741 & \best{0.0737} & 0.0760 & 0.0756 \\
 &  & P & 0.0777 & 0.1115 & 0.0780 & 0.0763 & 0.0772 & 0.0348 & 0.1003 \\
\netthree{} & 100 & A & \best{0.0574} & 0.0574 & 0.0909 & 0.0578 & 0.0574 & 0.0608 & 0.0582 \\
 &  & P & 0.0564 & 0.0882 & 0.0584 & 0.0553 & 0.0560 & 0.0239 & 0.0728 \\
\bottomrule
\end{tabu}
\caption{RMSE for the queried variables in the various networks: A stands for Actual, P for Predicted. Best results---also considering hidden decimals---for the Actual RMSE boxed. Monte Carlo approach has been run over 100 samples.}
    \label{tab:netsres}
\end{table}

\begin{figure*}[t]
\centering
    \begin{subfigure}[b]{0.3\textwidth}
        \includegraphics[width=\textwidth]{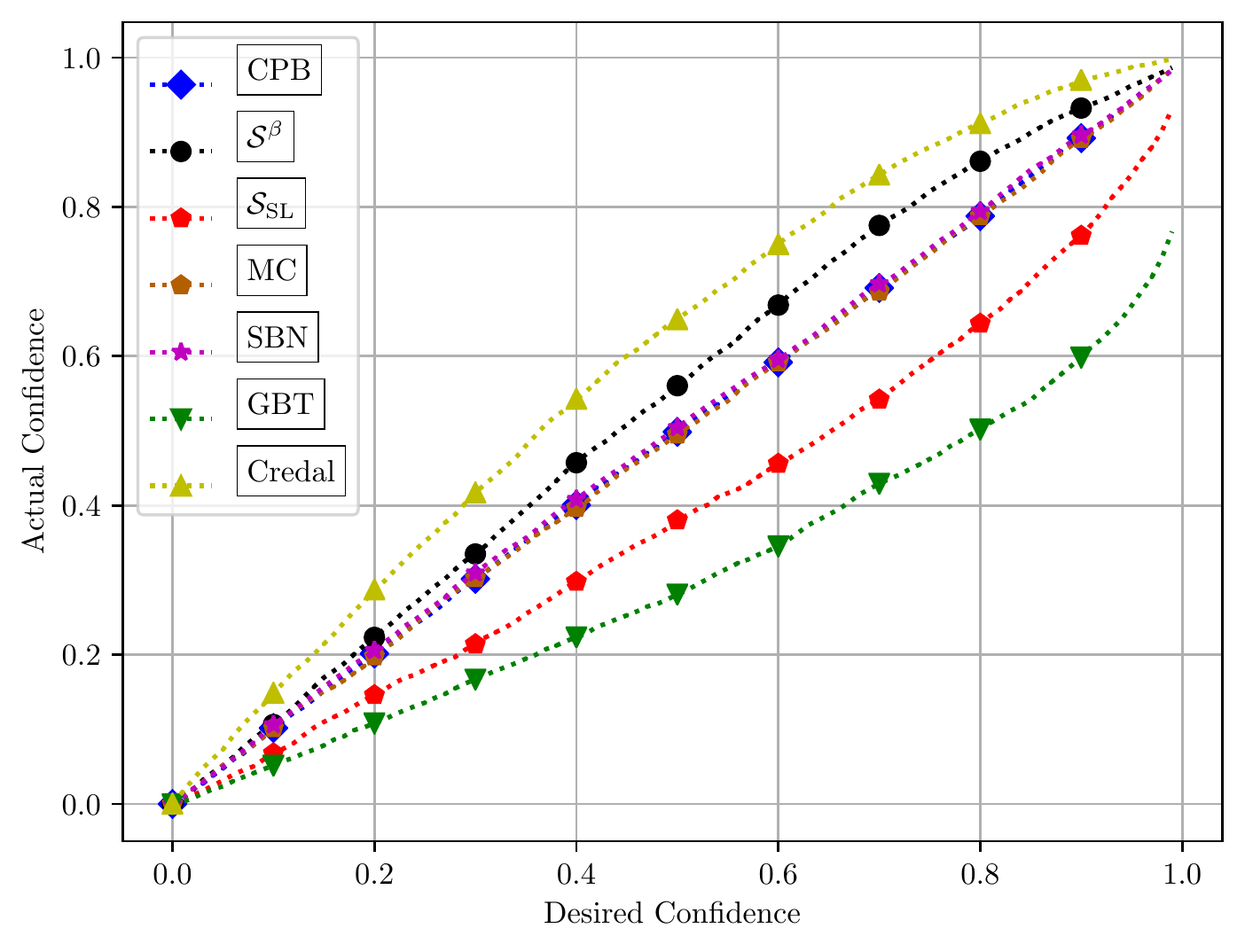}
        \caption{}
        \label{fig:net1a}
    \end{subfigure}
    ~
    \begin{subfigure}[b]{0.3\textwidth}
        \includegraphics[width=\textwidth]{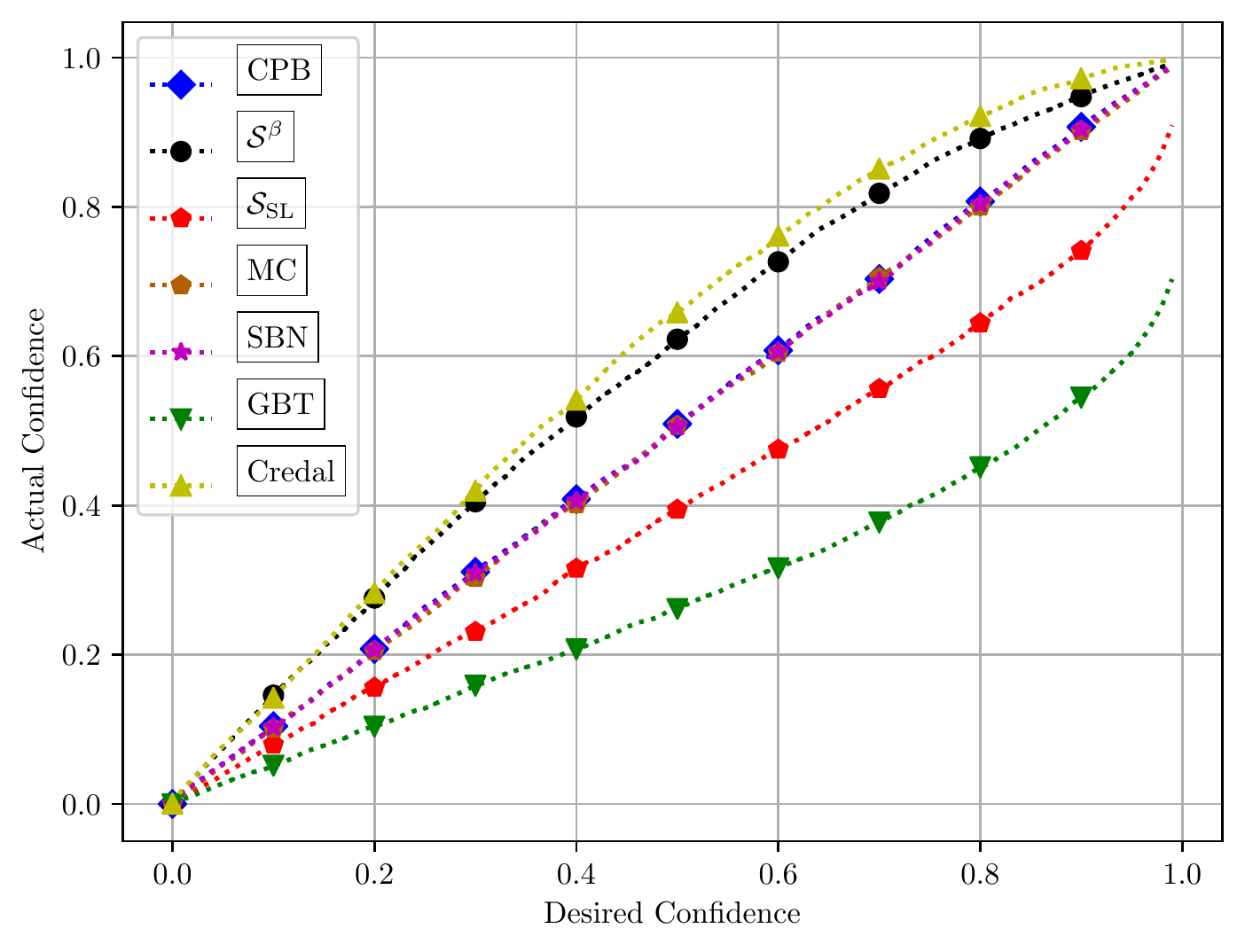}
        \caption{}
        \label{fig:net1b}
    \end{subfigure}
    ~
    \begin{subfigure}[b]{0.3\textwidth}
        \includegraphics[width=\textwidth]{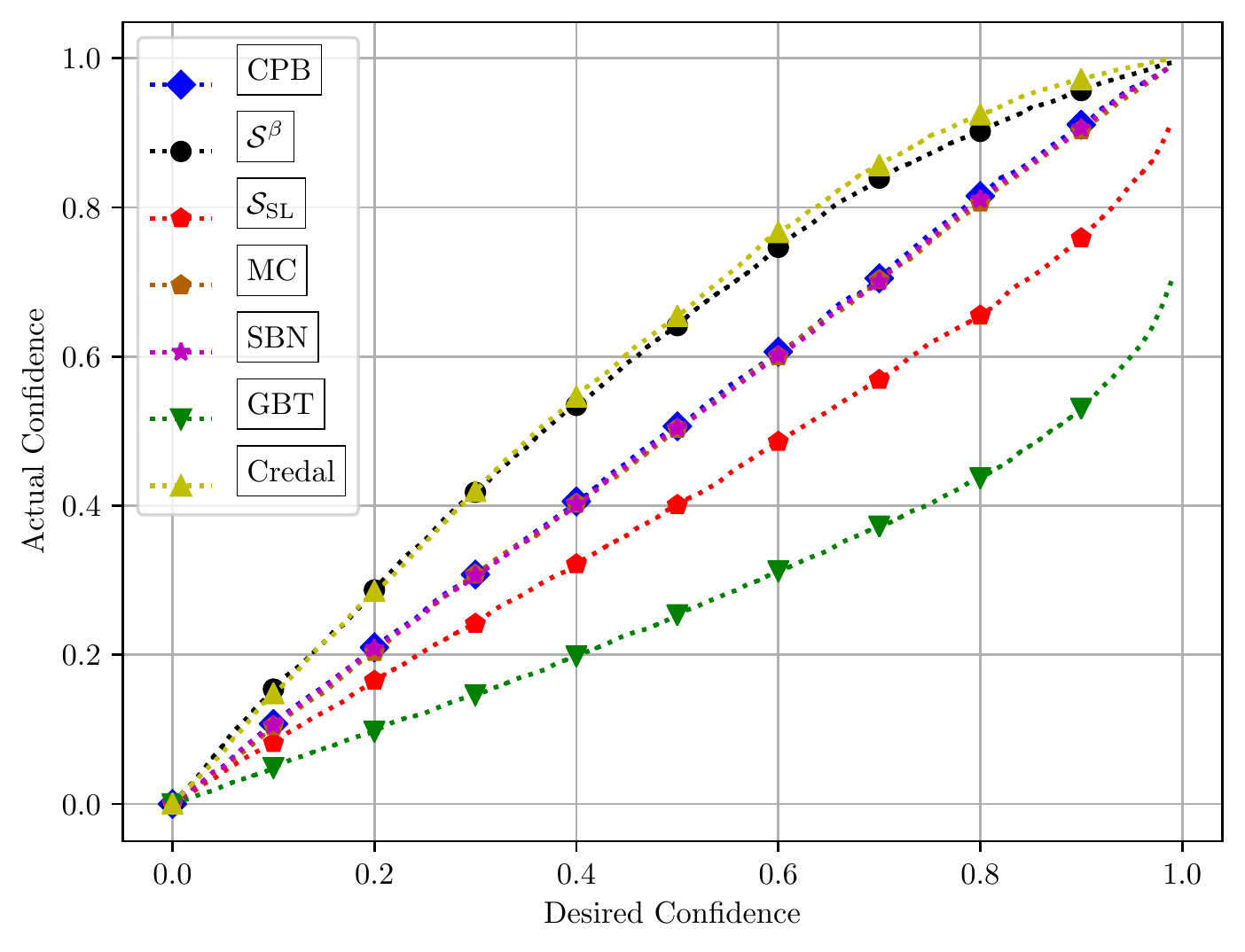}
        \caption{}
        \label{fig:net1c}
    \end{subfigure}
    
    \begin{subfigure}[b]{0.3\textwidth}
        \includegraphics[width=\textwidth]{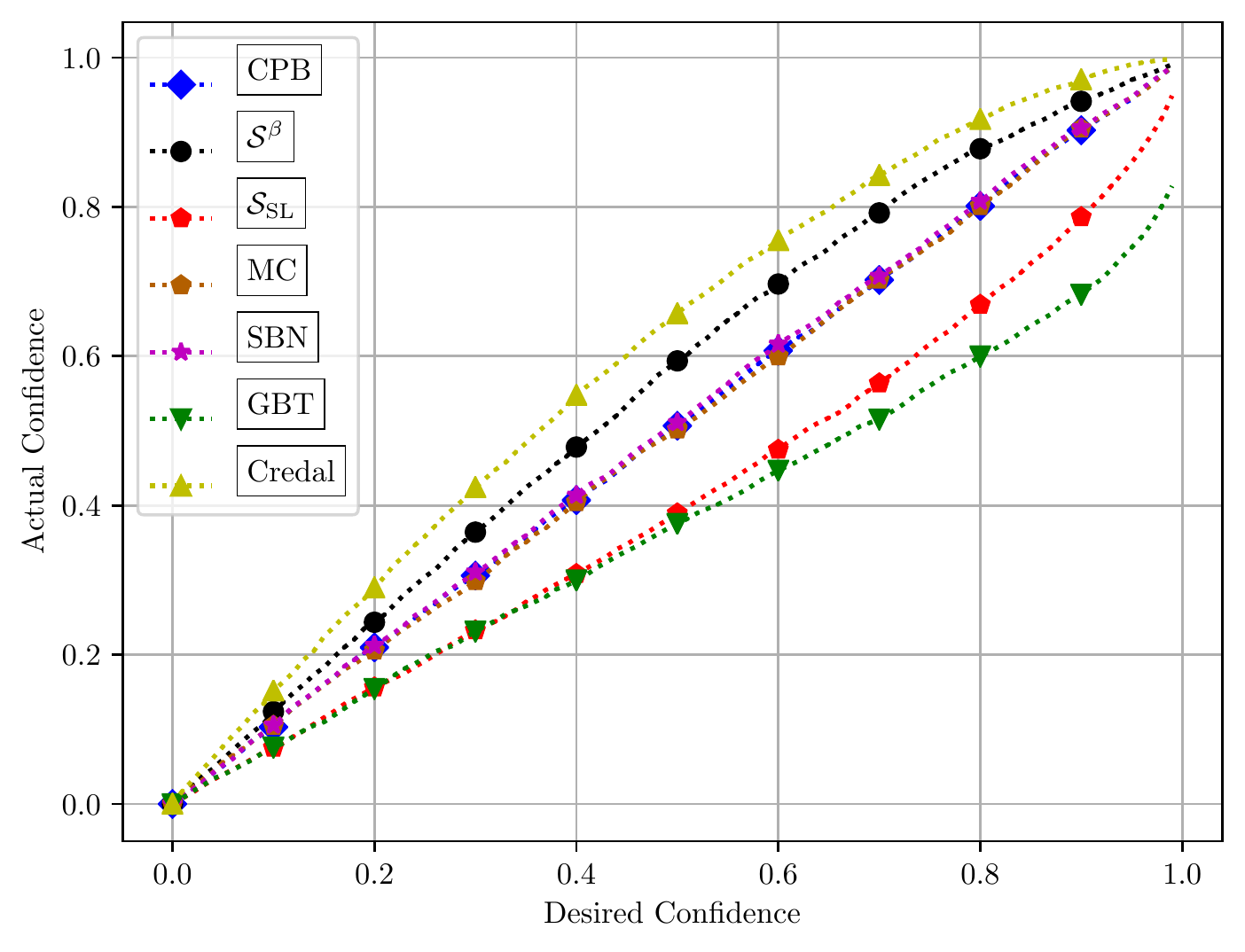}
        \caption{}
        \label{fig:net2a}
    \end{subfigure}
    ~
    \begin{subfigure}[b]{0.3\textwidth}
        \includegraphics[width=\textwidth]{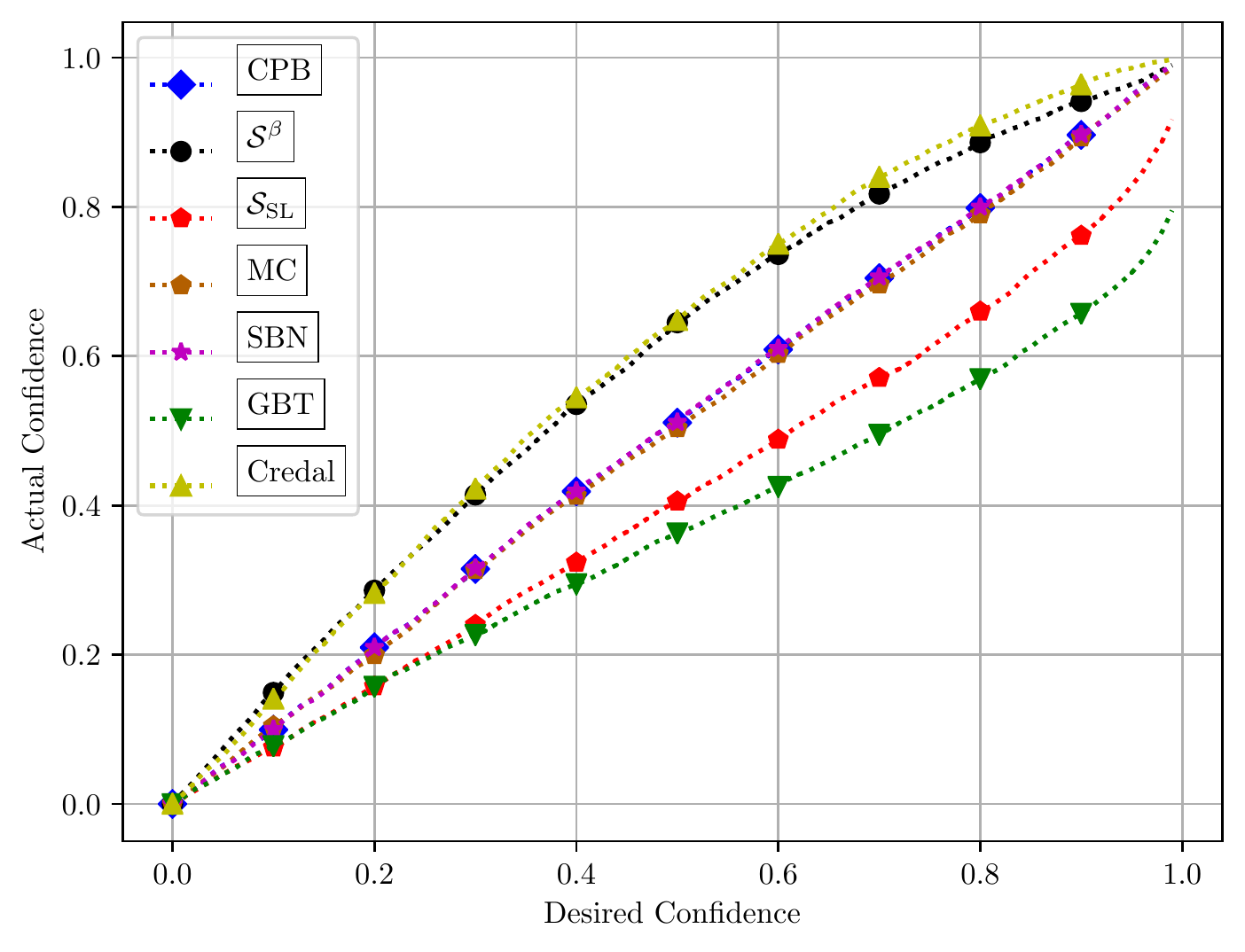}
        \caption{}
        \label{fig:net2b}
    \end{subfigure}
    ~
    \begin{subfigure}[b]{0.3\textwidth}
        \includegraphics[width=\textwidth]{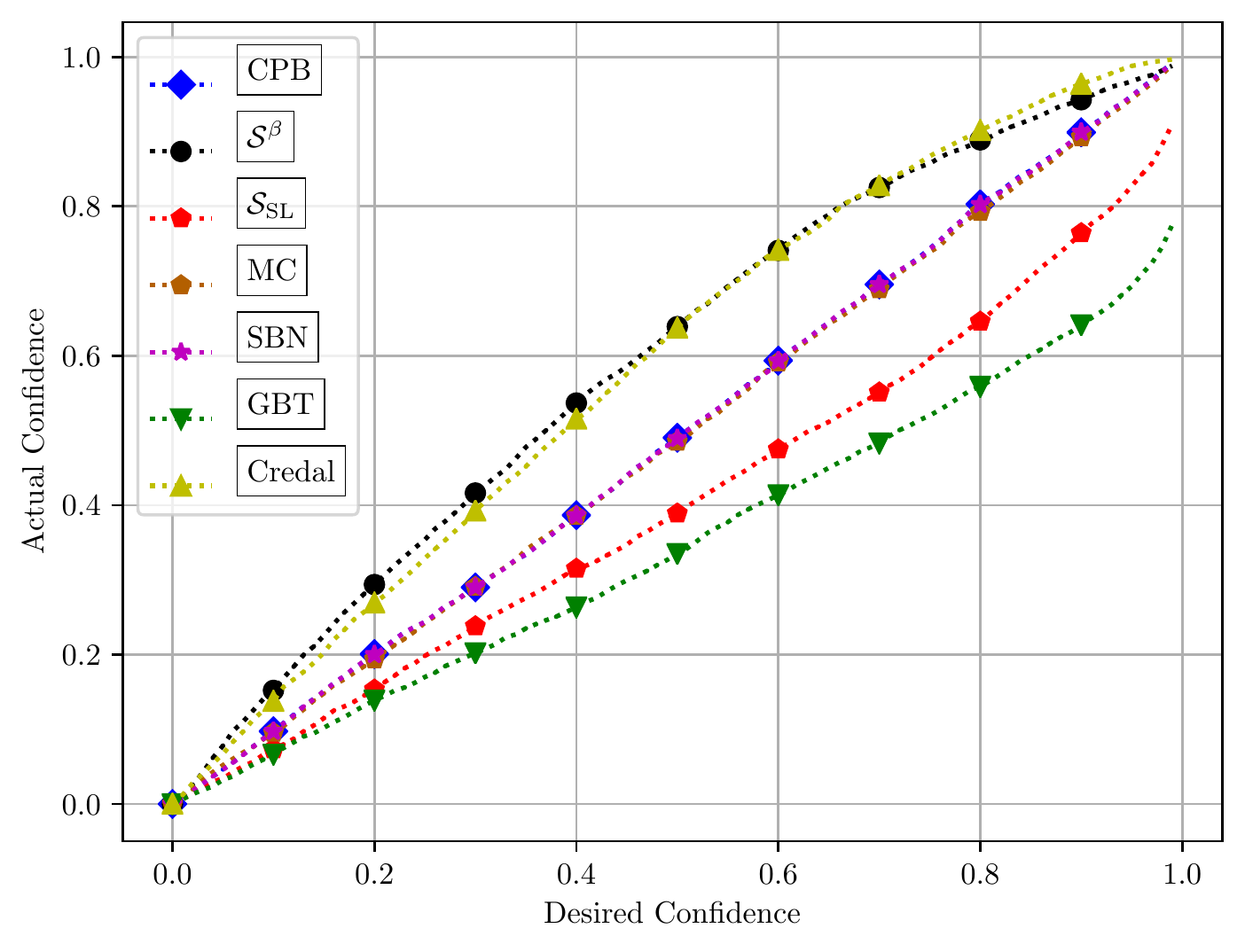}
        \caption{}
        \label{fig:net2c}
    \end{subfigure}
    
    \begin{subfigure}[b]{0.3\textwidth}
        \includegraphics[width=\textwidth]{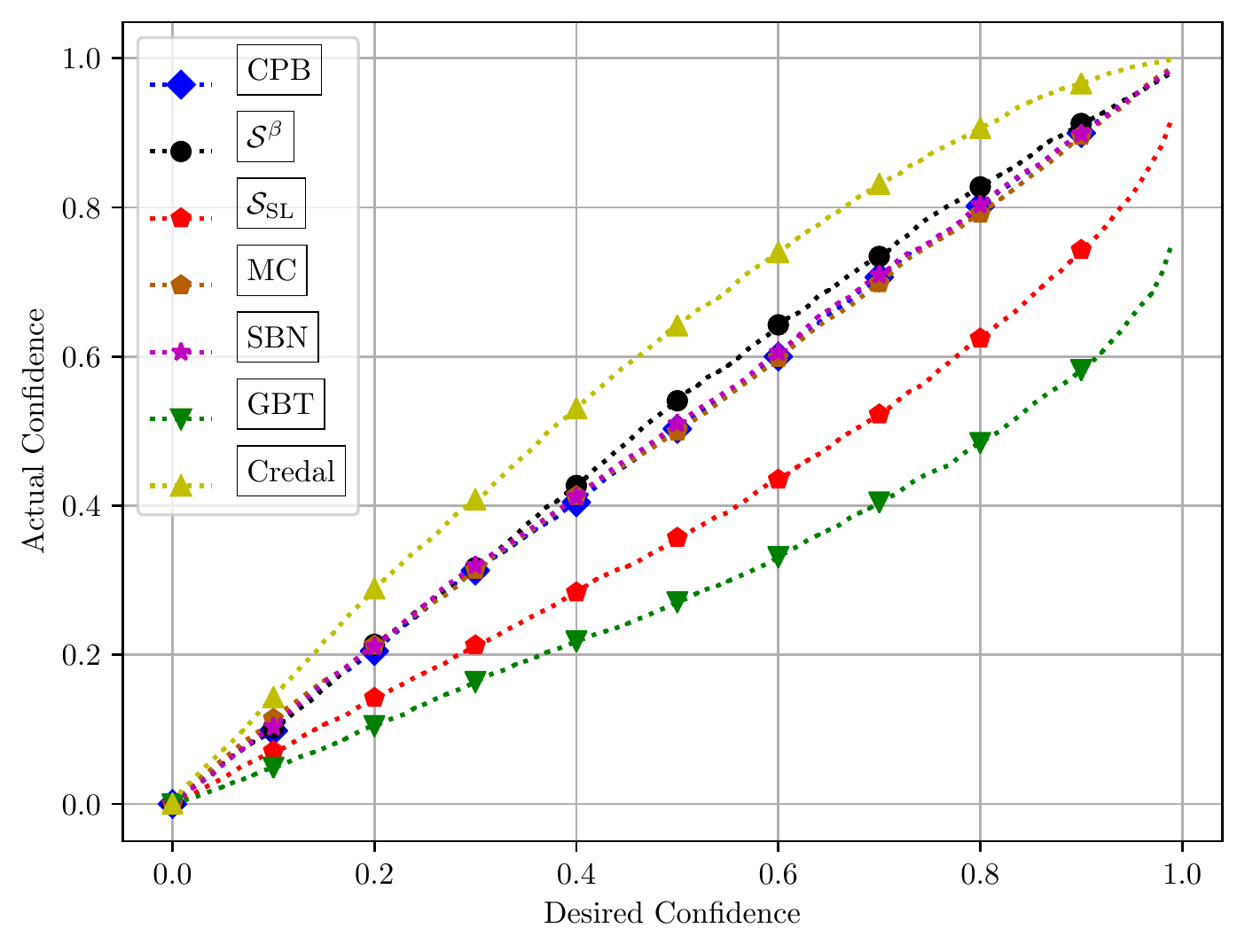}
        \caption{}
        \label{fig:net3a}
    \end{subfigure}
    ~
    \begin{subfigure}[b]{0.3\textwidth}
        \includegraphics[width=\textwidth]{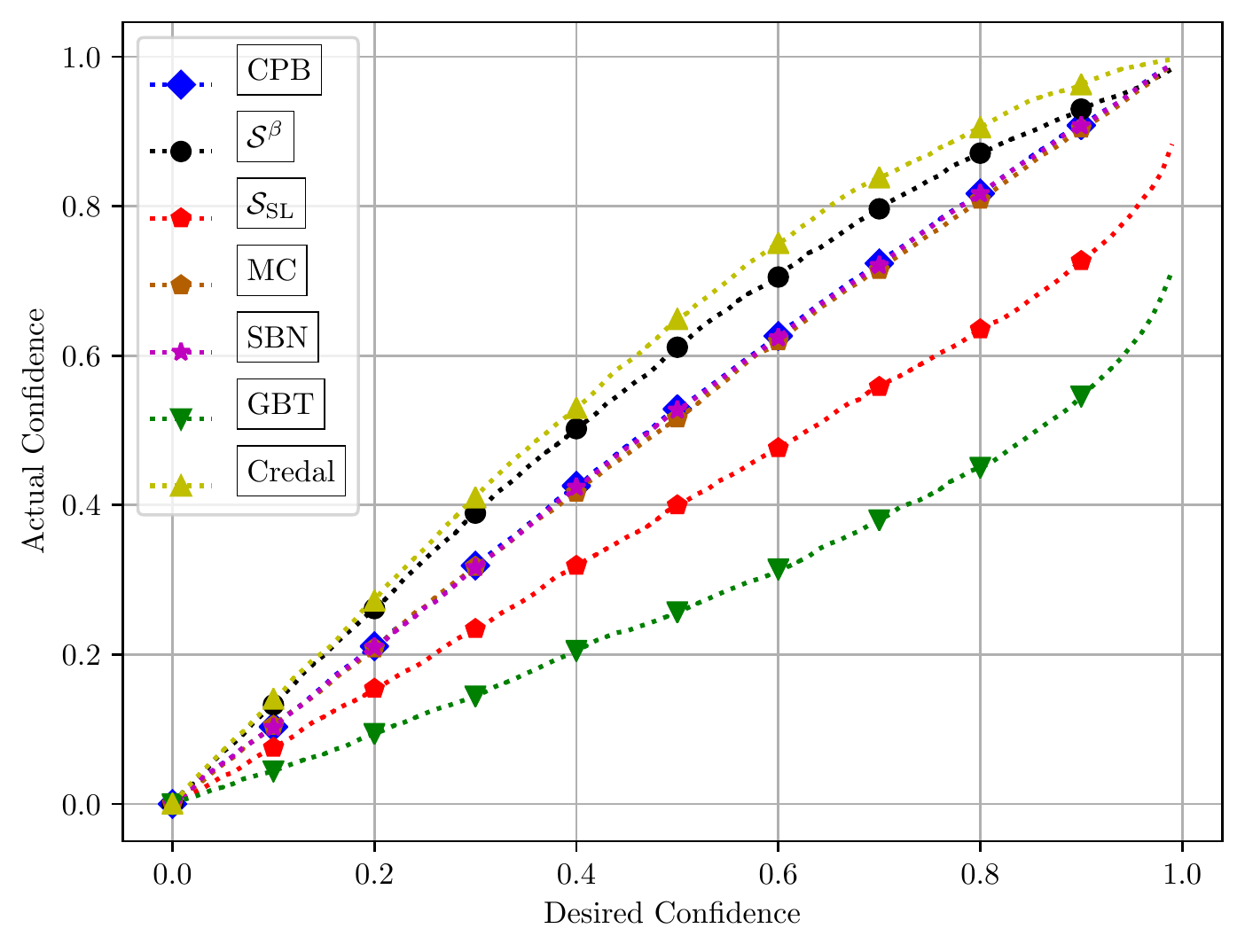}
        \caption{}
        \label{fig:net3b}
    \end{subfigure}
    ~
    \begin{subfigure}[b]{0.3\textwidth}
        \includegraphics[width=\textwidth]{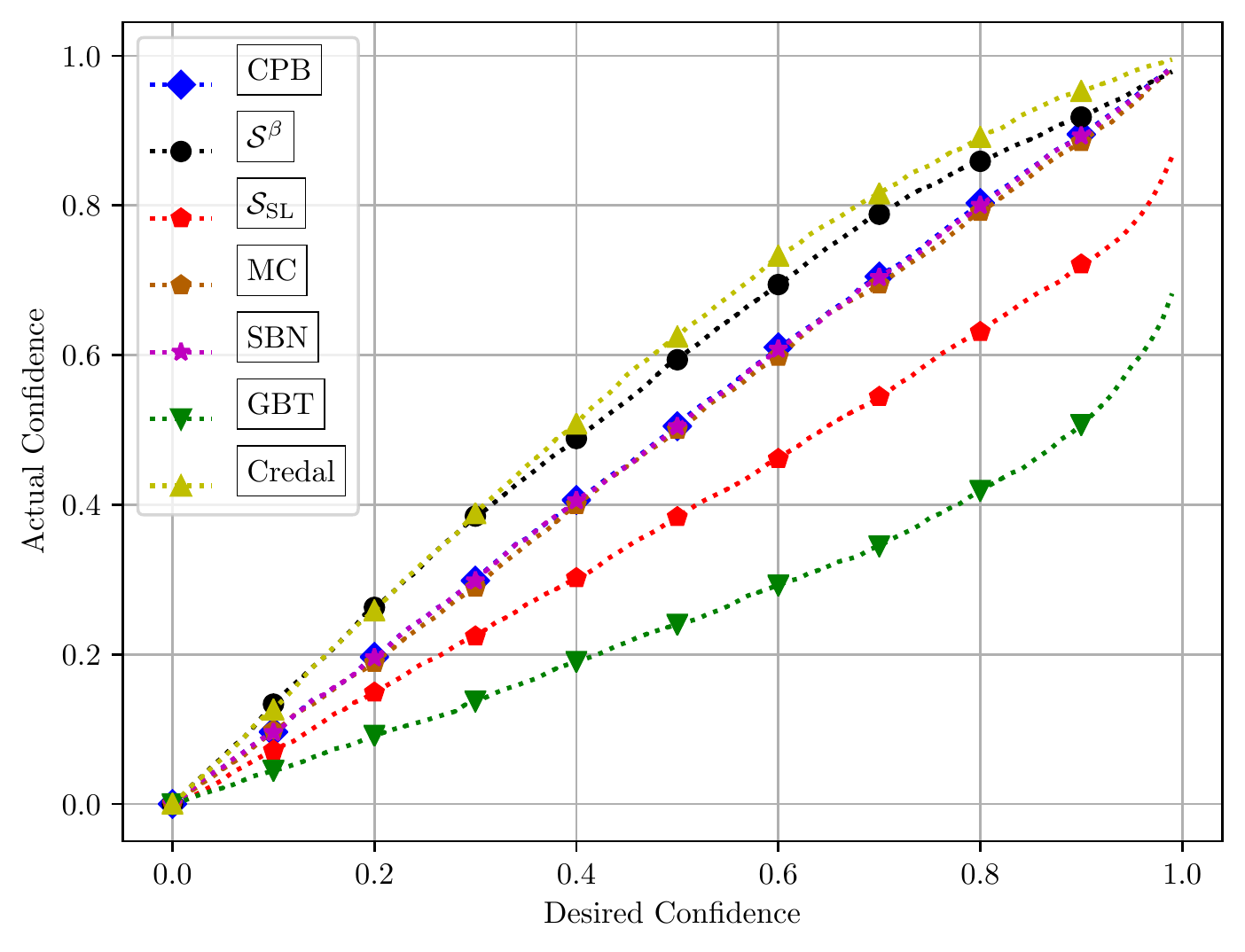}
        \caption{}
        \label{fig:net3c}
    \end{subfigure}
    
    \caption{Actual versus desired significance of bounds derived from the uncertainty for: (\subref{fig:net1a}) Net1 with $N_{ins} = 10$; (\subref{fig:net1b}) Net1 with $N_{ins} = 50$; (\subref{fig:net1c}) Net1 with $N_{ins} = 100$; 
    (\subref{fig:net2a}) Net2 with $N_{ins} = 10$; (\subref{fig:net2b}) Net2 with $N_{ins} = 50$; (\subref{fig:net2c}) Net2 with $N_{ins} = 100$;
    (\subref{fig:net3a}) Net3 with $N_{ins} = 10$; (\subref{fig:net3b}) Net3 with $N_{ins} = 50$; (\subref{fig:net3c}) Net3 with $N_{ins} = 100$. Best closest to the diagonal.  Monte Carlo approach has been run over 100 samples.}
    \label{fig:netsres}
\end{figure*}

\begin{figure*}[t]
\centering
    \begin{subfigure}[b]{0.3\textwidth}
        \includegraphics[width=\textwidth]{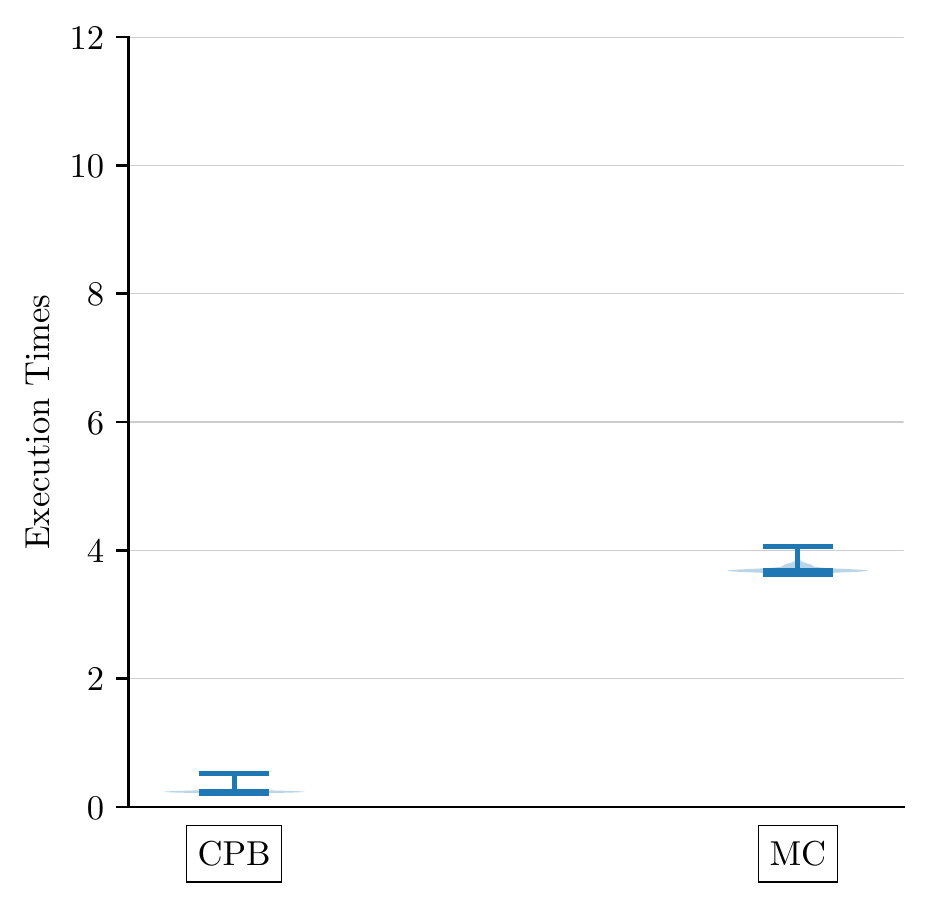}
        \caption{}
        \label{fig:net1atimes}
    \end{subfigure}
    ~
    \begin{subfigure}[b]{0.3\textwidth}
        \includegraphics[width=\textwidth]{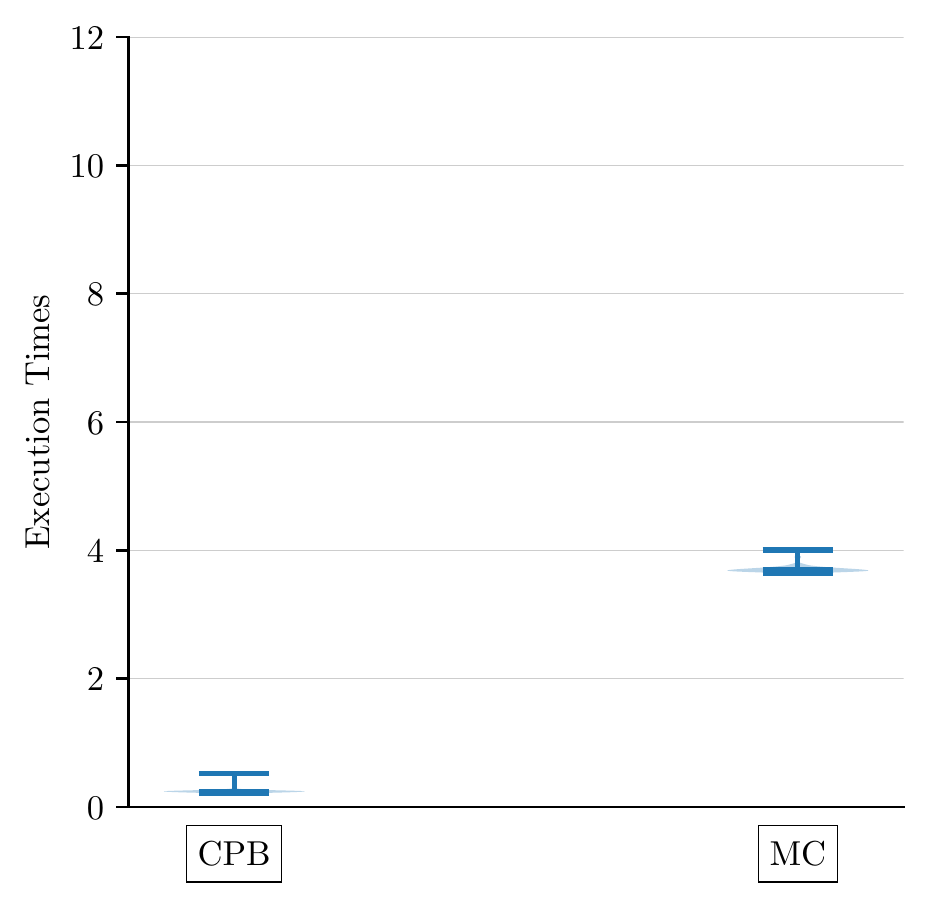}
        \caption{}
        \label{fig:net1btimes}
    \end{subfigure}
    ~
    \begin{subfigure}[b]{0.3\textwidth}
        \includegraphics[width=\textwidth]{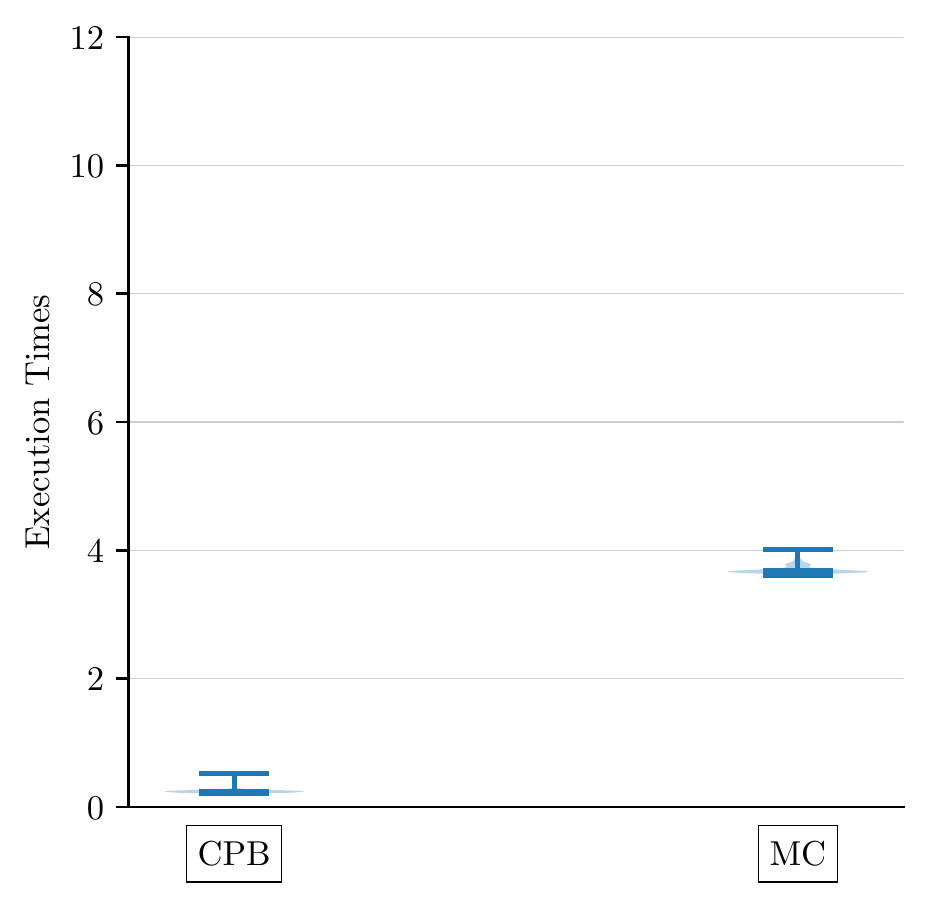}
        \caption{}
        \label{fig:net1ctimes}
    \end{subfigure}
    
    \begin{subfigure}[b]{0.3\textwidth}
        \includegraphics[width=\textwidth]{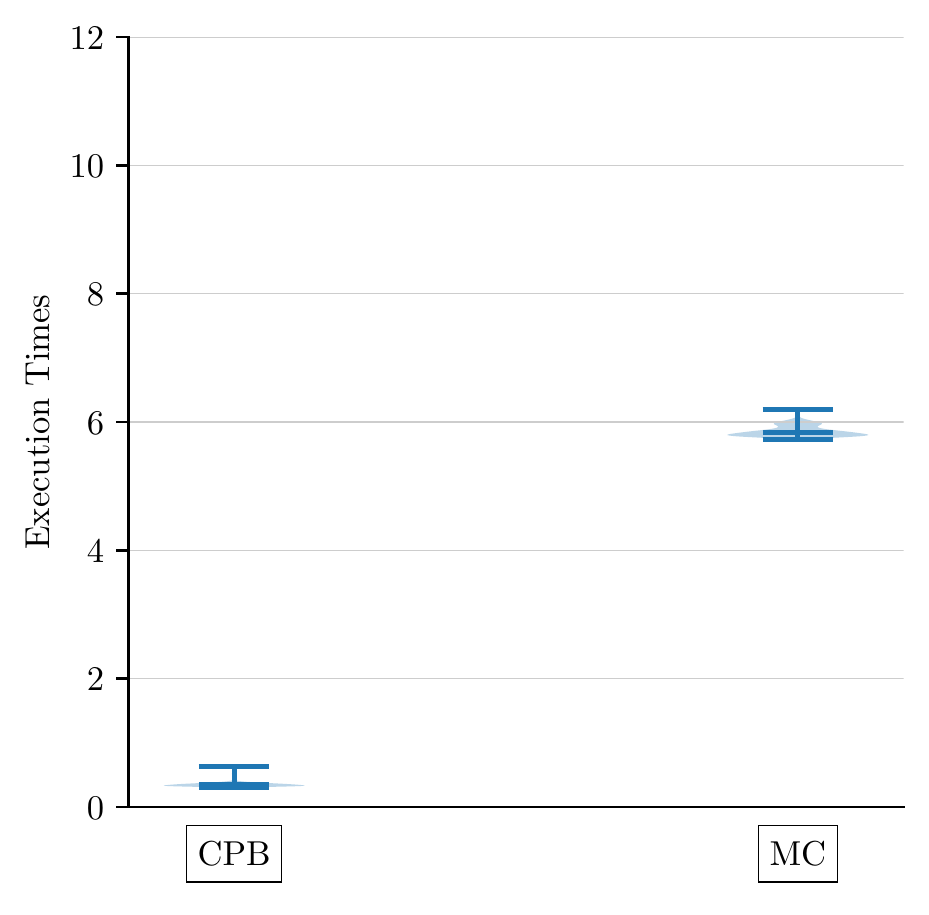}
        \caption{}
        \label{fig:net2atimes}
    \end{subfigure}
    ~
    \begin{subfigure}[b]{0.3\textwidth}
        \includegraphics[width=\textwidth]{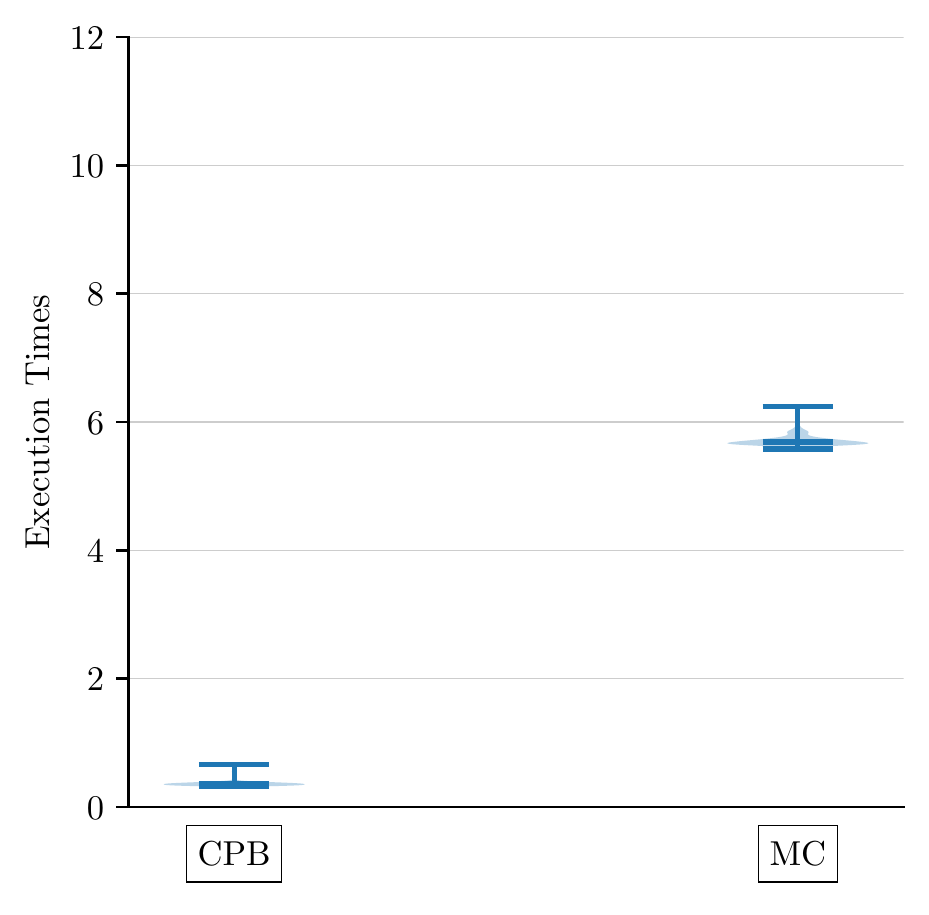}
        \caption{}
        \label{fig:net2btimes}
    \end{subfigure}
    ~
    \begin{subfigure}[b]{0.3\textwidth}
        \includegraphics[width=\textwidth]{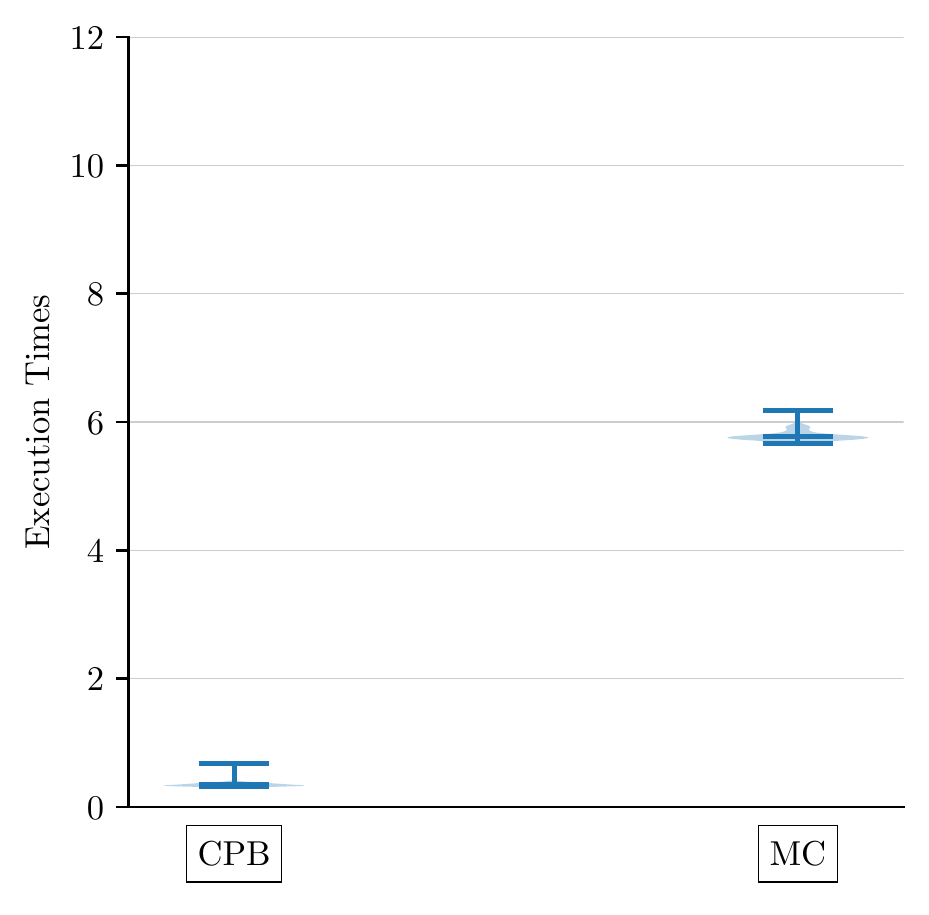}
        \caption{}
        \label{fig:net2ctimes}
    \end{subfigure}
    
    \begin{subfigure}[b]{0.3\textwidth}
        \includegraphics[width=\textwidth]{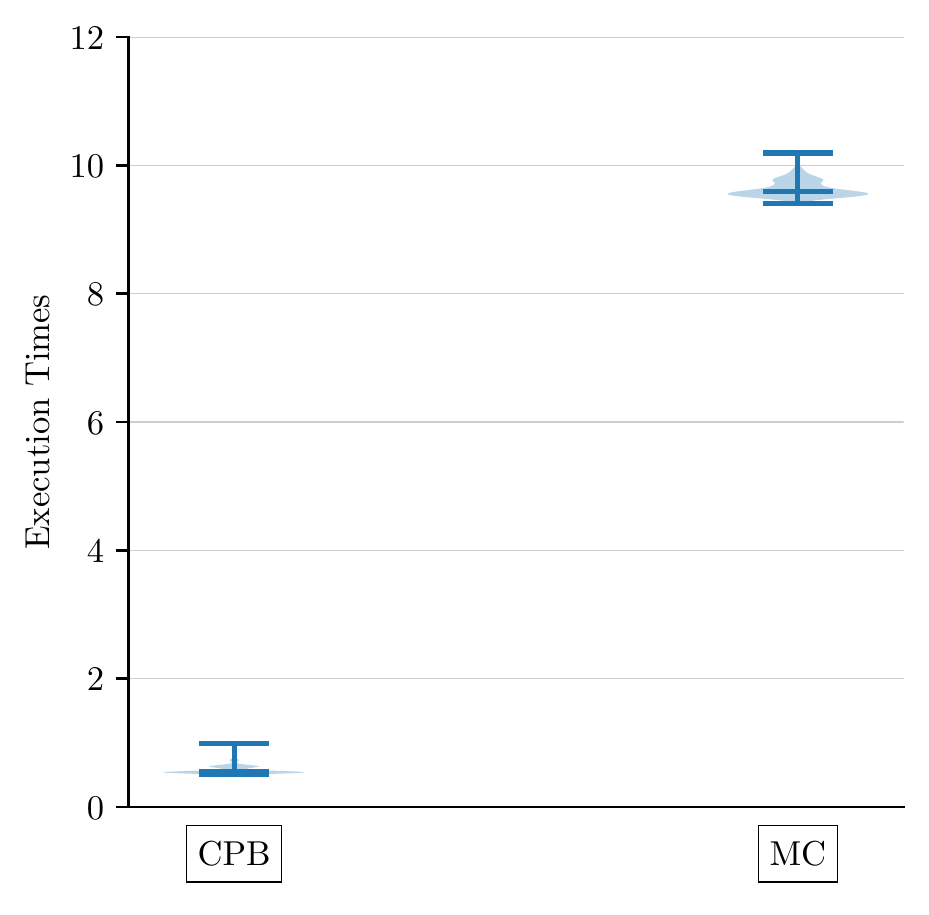}
        \caption{}
        \label{fig:net3atimes}
    \end{subfigure}
    ~
    \begin{subfigure}[b]{0.3\textwidth}
        \includegraphics[width=\textwidth]{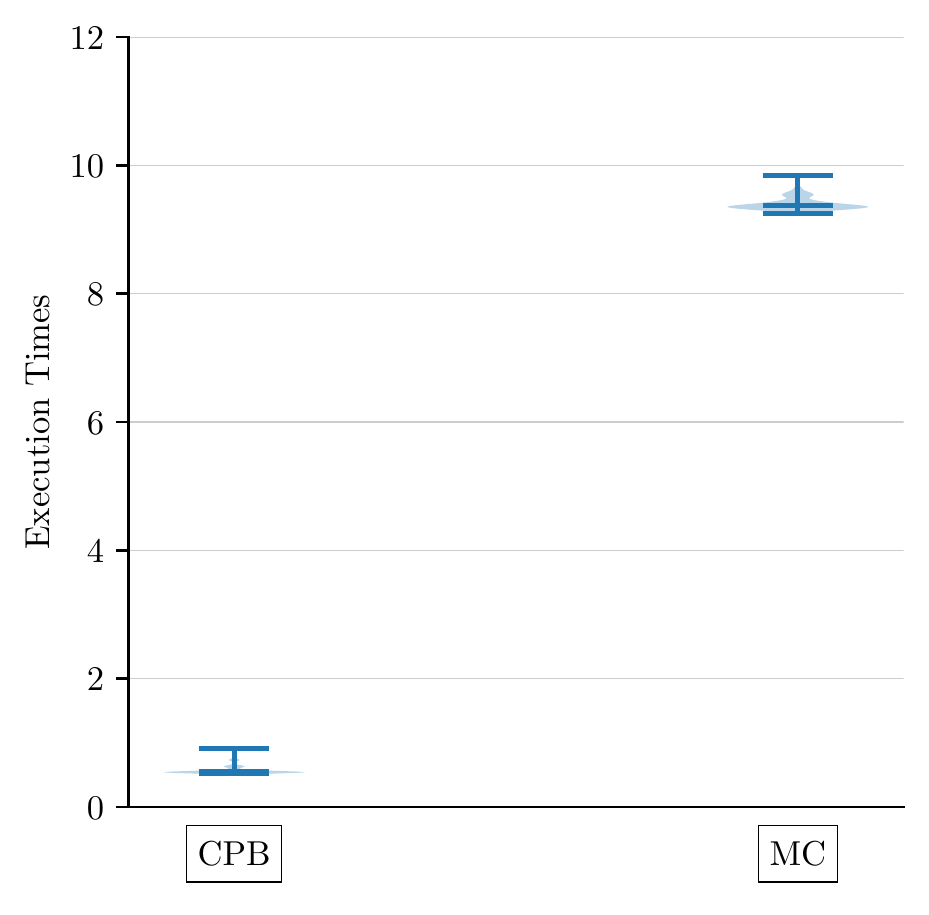}
        \caption{}
        \label{fig:net3btimes}
    \end{subfigure}
    ~
    \begin{subfigure}[b]{0.3\textwidth}
        \includegraphics[width=\textwidth]{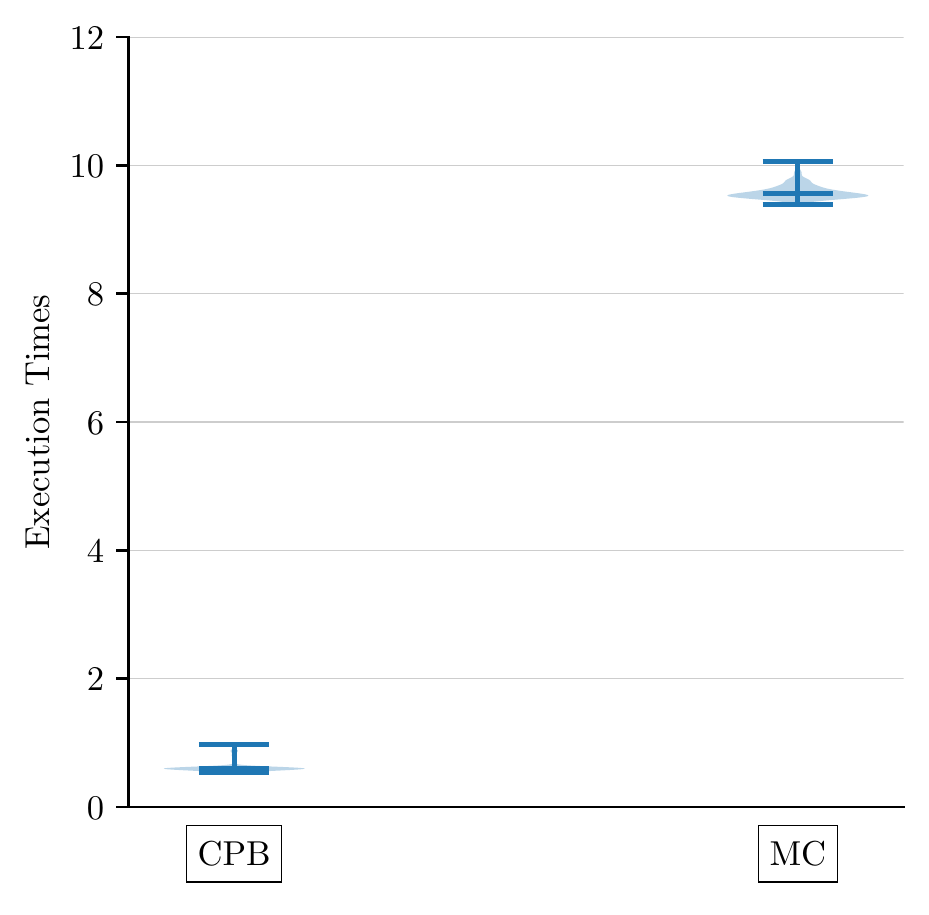}
        \caption{}
        \label{fig:net3ctimes}
    \end{subfigure}
    
    \caption{Distribution of computational time for running the different algorithms for: (\subref{fig:net1atimes}) Net1 with $N_{ins} = 10$; (\subref{fig:net1btimes}) Net1 with $N_{ins} = 50$; (\subref{fig:net1ctimes}) Net1 with $N_{ins} = 100$; 
    (\subref{fig:net2atimes}) Net2 with $N_{ins} = 10$; (\subref{fig:net2btimes}) Net2 with $N_{ins} = 50$; (\subref{fig:net2ctimes}) Net2 with $N_{ins} = 100$;
    (\subref{fig:net3atimes}) Net3 with $N_{ins} = 10$; (\subref{fig:net3btimes}) Net3 with $N_{ins} = 50$; (\subref{fig:net3ctimes}) Net3 with $N_{ins} = 100$.
     Monte Carlo approach has been run over 100 samples.}
    \label{fig:nettimes}
\end{figure*}

\begin{figure*}[t]
\centering
    \begin{subfigure}[b]{0.3\textwidth}
        \includegraphics[width=\textwidth]{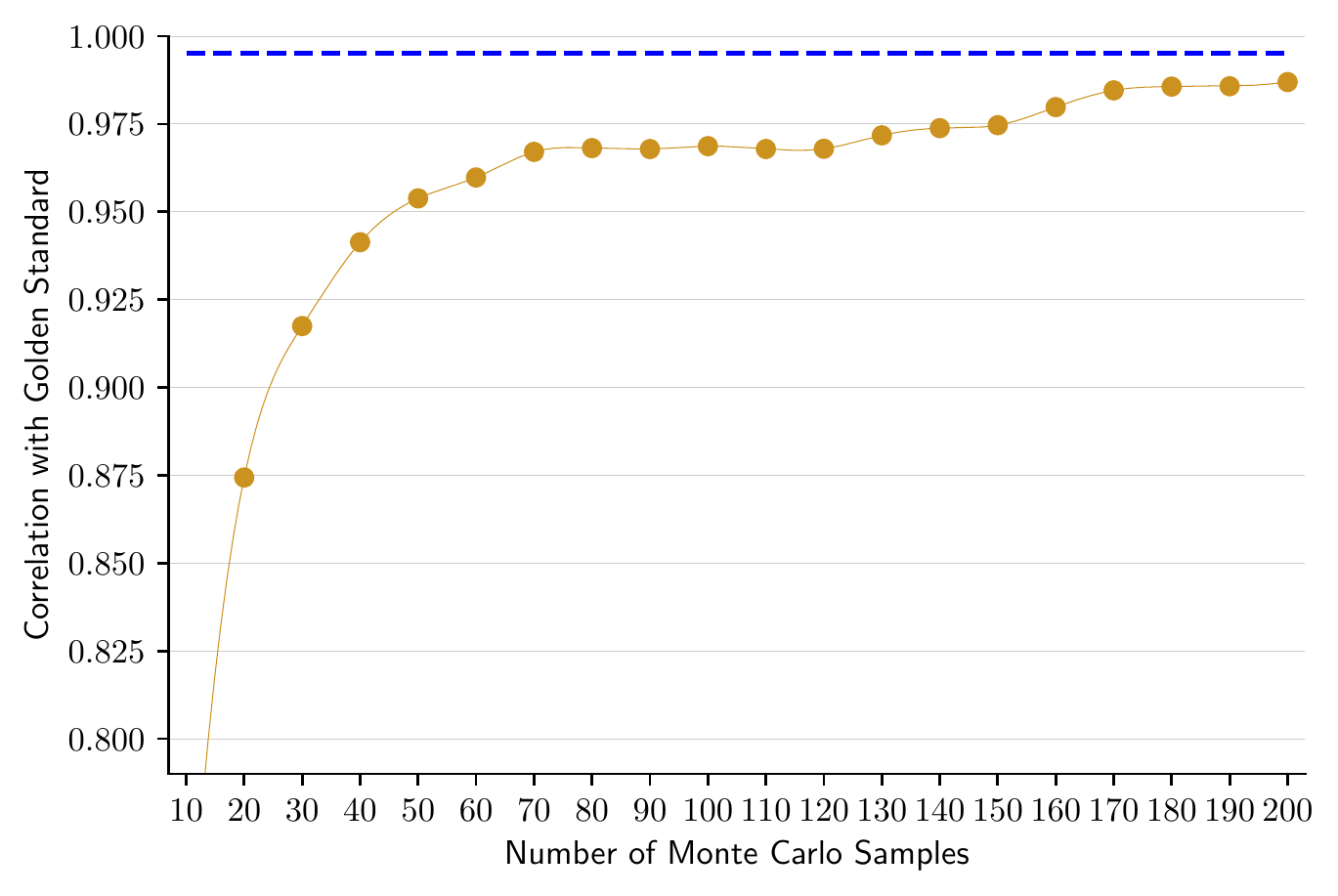}
        \caption{}
        \label{fig:net1acorr}
    \end{subfigure}
    ~
    \begin{subfigure}[b]{0.3\textwidth}
        \includegraphics[width=\textwidth]{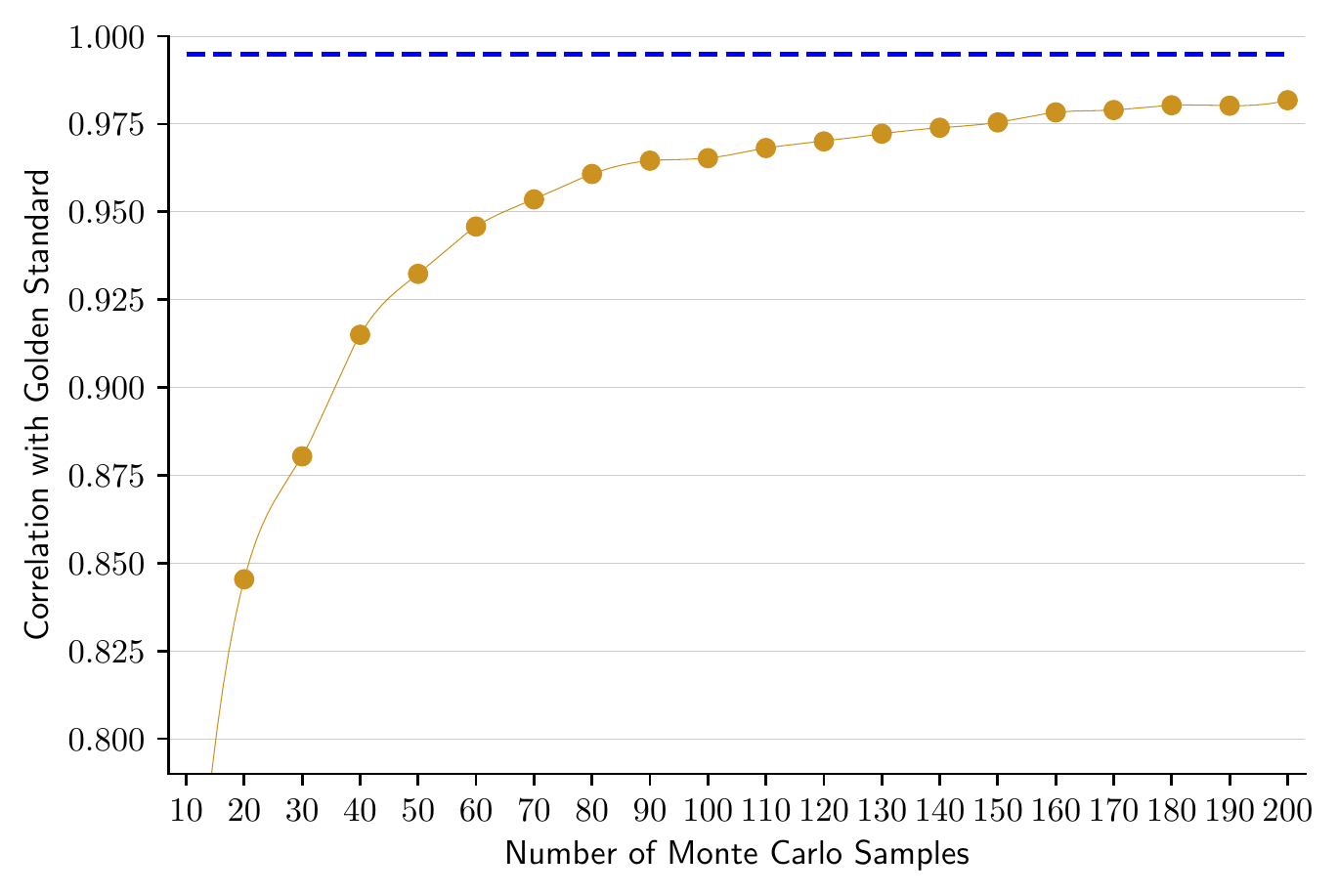}
        \caption{}
        \label{fig:net1bcorr}
    \end{subfigure}
    ~
    \begin{subfigure}[b]{0.3\textwidth}
        \includegraphics[width=\textwidth]{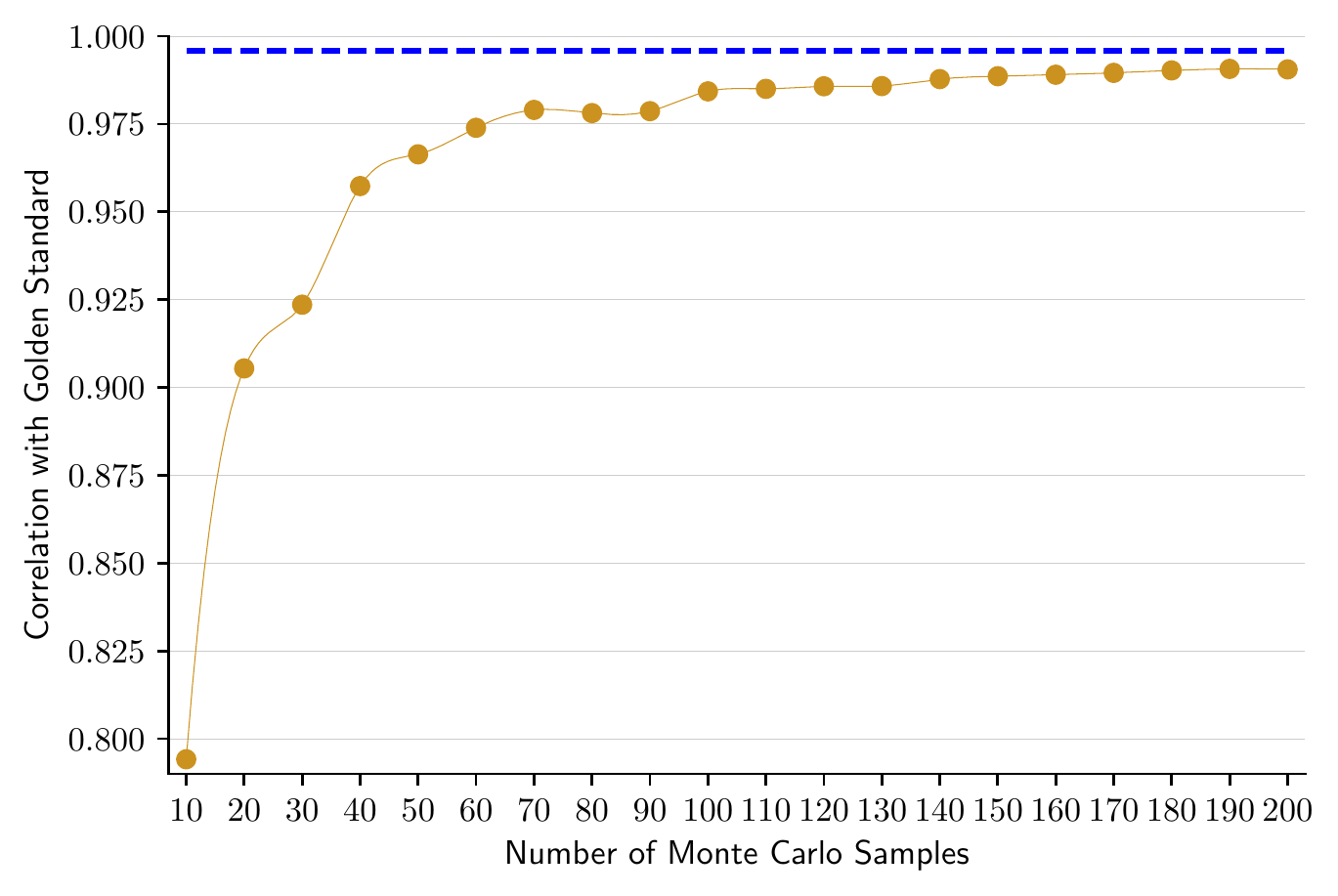}
        \caption{}
        \label{fig:net1ccorr}
    \end{subfigure}
    
    \begin{subfigure}[b]{0.3\textwidth}
        \includegraphics[width=\textwidth]{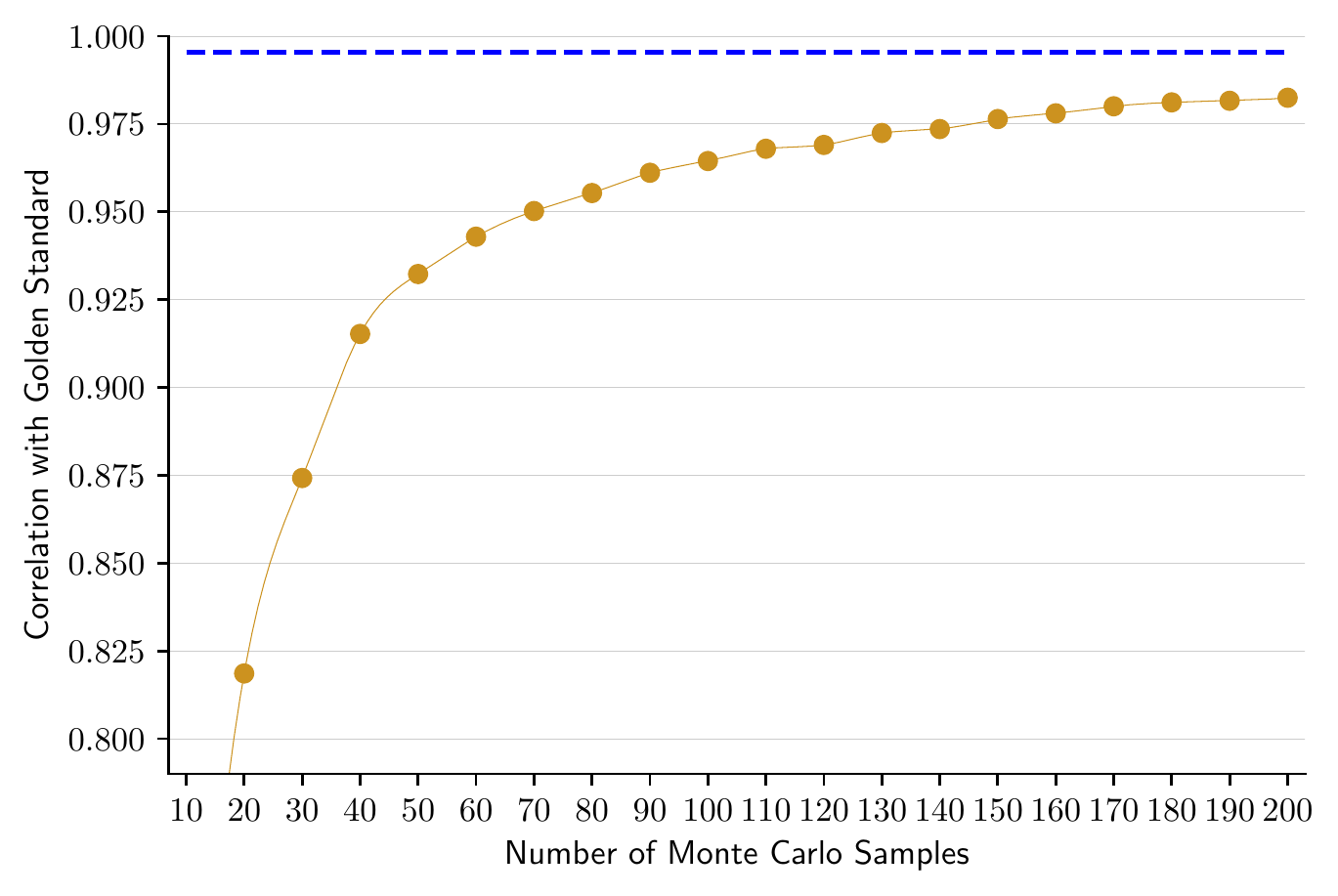}
        \caption{}
        \label{fig:net2acorr}
    \end{subfigure}
    ~
    \begin{subfigure}[b]{0.3\textwidth}
        \includegraphics[width=\textwidth]{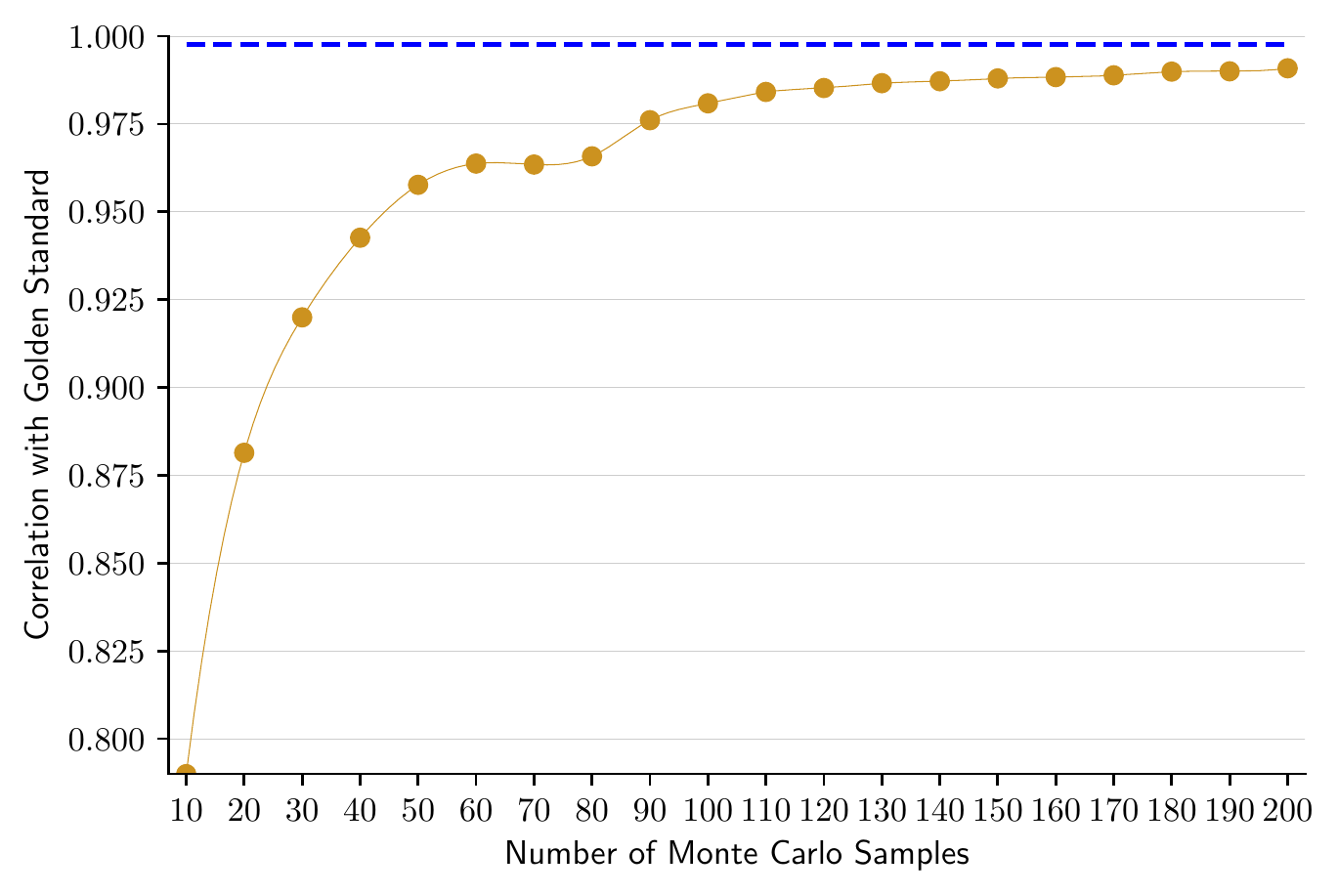}
        \caption{}
        \label{fig:net2bcorr}
    \end{subfigure}
    ~
    \begin{subfigure}[b]{0.3\textwidth}
        \includegraphics[width=\textwidth]{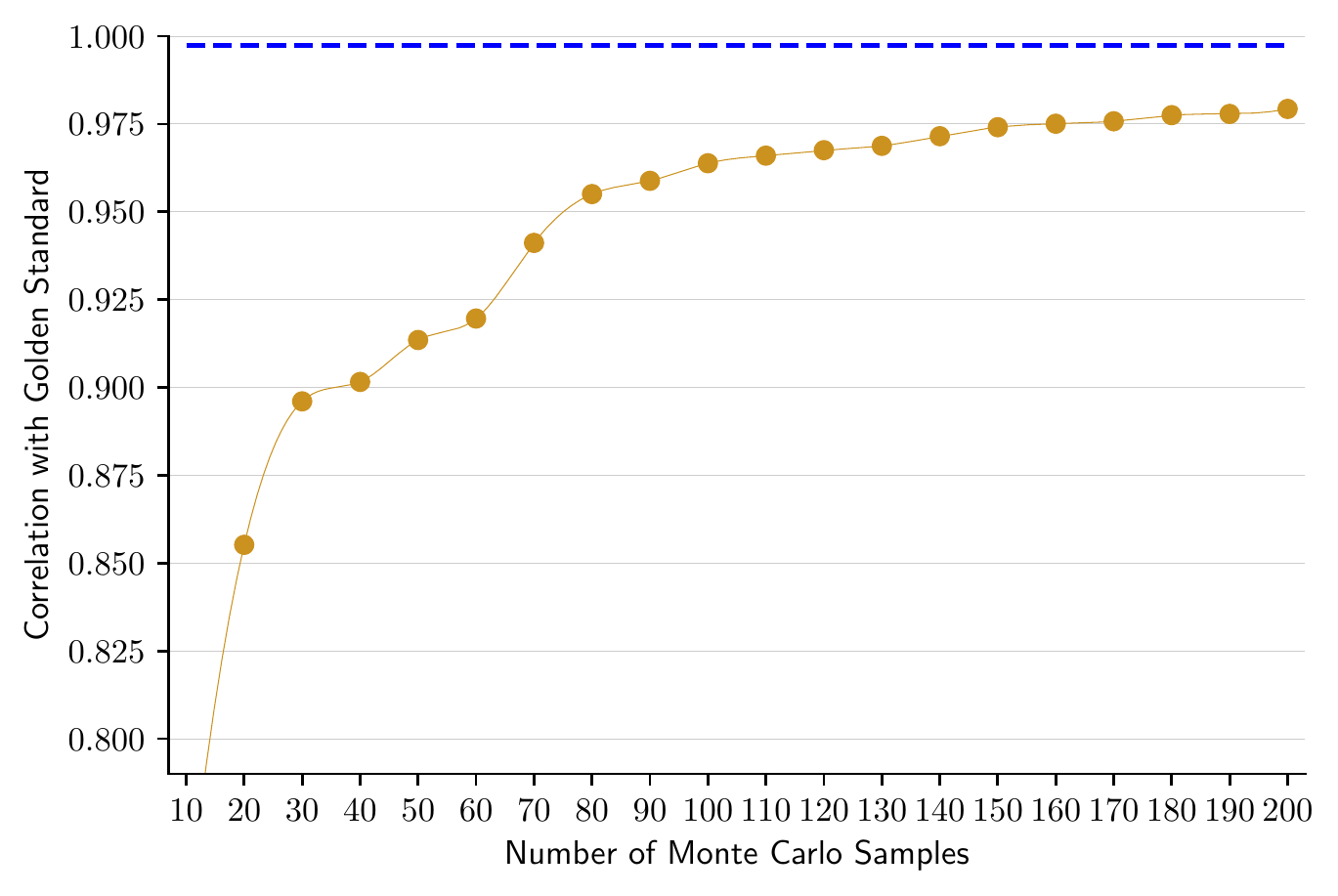}
        \caption{}
        \label{fig:net2ccorr}
    \end{subfigure}
    
    \begin{subfigure}[b]{0.3\textwidth}
        \includegraphics[width=\textwidth]{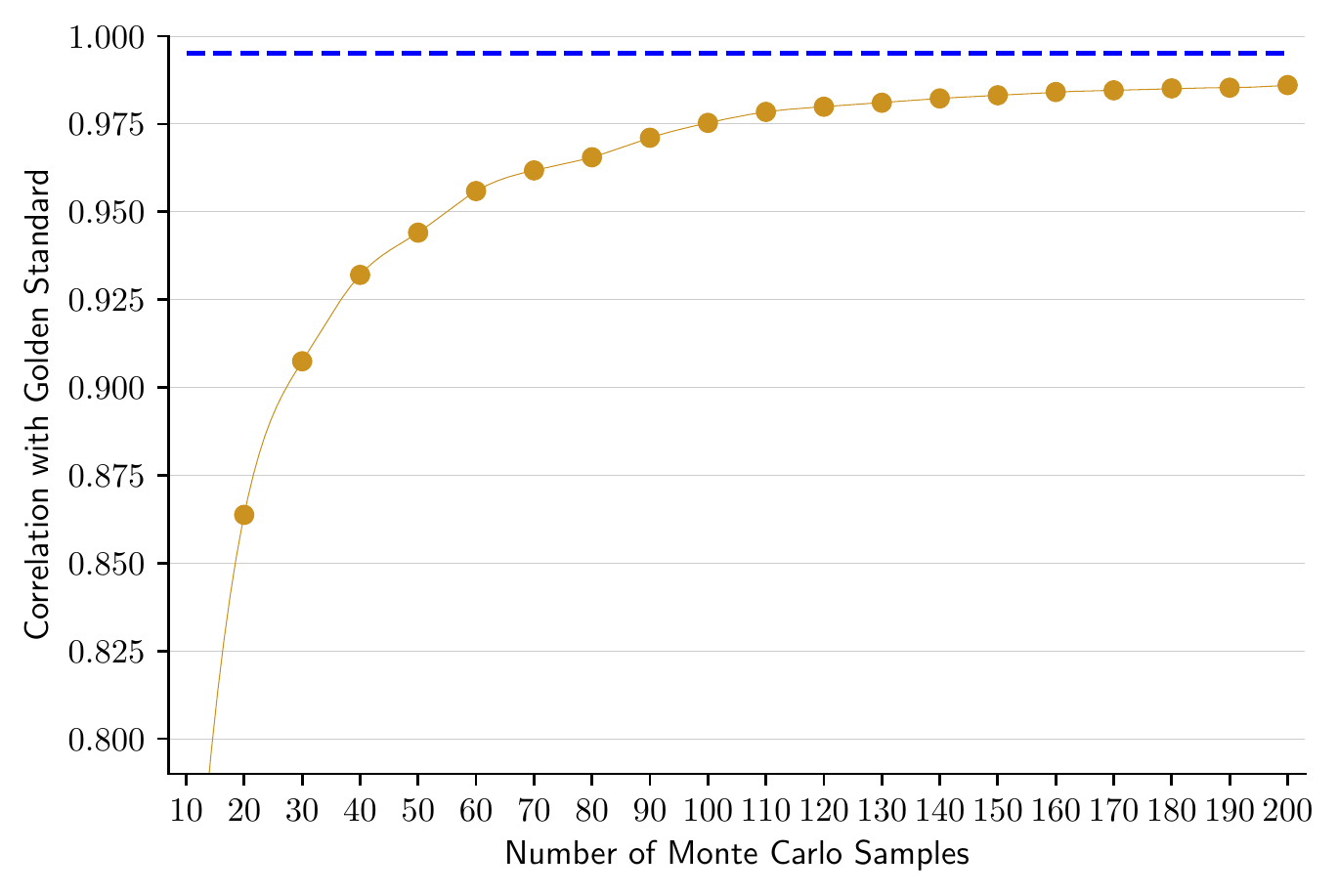}
        \caption{}
        \label{fig:net3acorr}
    \end{subfigure}
    ~
    \begin{subfigure}[b]{0.3\textwidth}
        \includegraphics[width=\textwidth]{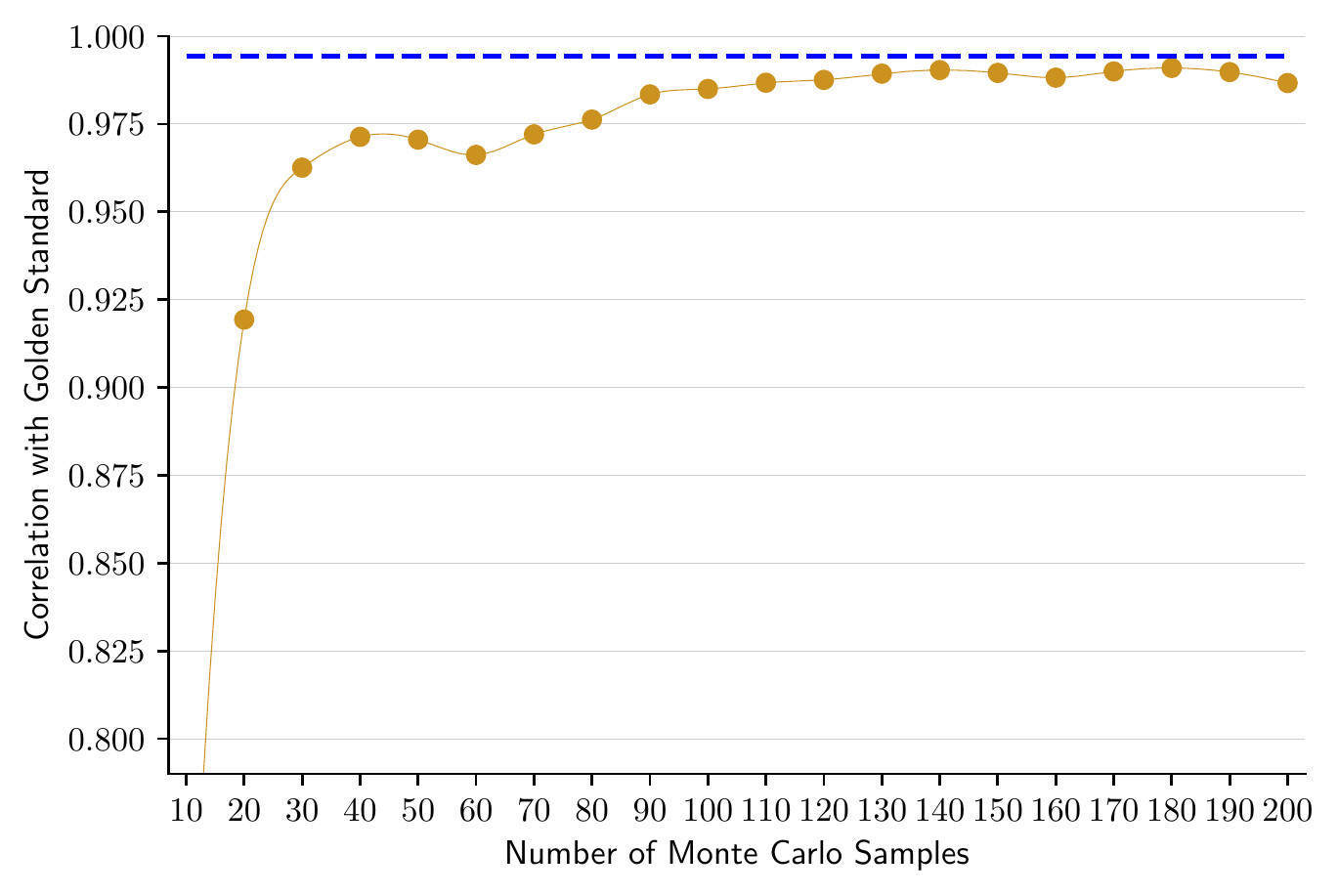}
        \caption{}
        \label{fig:net3bcorr}
    \end{subfigure}
    ~
    \begin{subfigure}[b]{0.3\textwidth}
        \includegraphics[width=\textwidth]{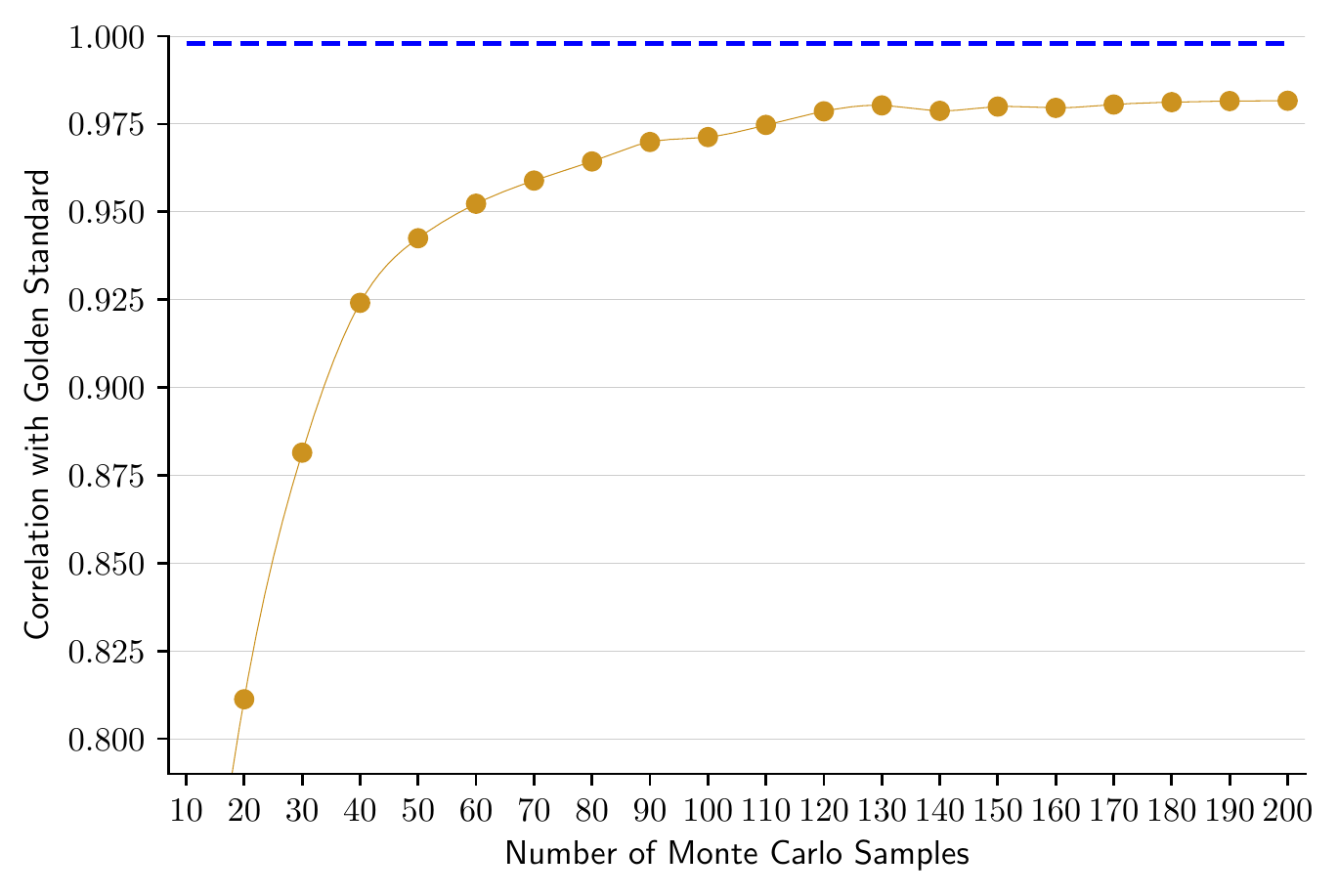}
        \caption{}
        \label{fig:net3ccorr}
    \end{subfigure}
    
    \caption{Correlation of Dirichlet strengths between runs of Monte Carlo approarch varying the number of samples and golden standard (i.e. a Monte Carlo run with 10,000 samples) as well as between the proposed approach and golden standard with cubic interpolation---that is independent of the number of samples used in Monte Carlo---for: (\subref{fig:net1acorr}) Net1 with $N_{ins} = 10$; (\subref{fig:net1bcorr}) Net1 with $N_{ins} = 50$; (\subref{fig:net1ccorr}) Net1 with $N_{ins} = 100$; 
    (\subref{fig:net2acorr}) Net2 with $N_{ins} = 10$; (\subref{fig:net2bcorr}) Net2 with $N_{ins} = 50$; (\subref{fig:net2ccorr}) Net2 with $N_{ins} = 100$;
    (\subref{fig:net3acorr}) Net3 with $N_{ins} = 10$; (\subref{fig:net3bcorr}) Net3 with $N_{ins} = 50$; (\subref{fig:net3ccorr}) Net3 with $N_{ins} = 100$.}
    \label{fig:netsrescorr}
\end{figure*}

As before, 
Table \ref{tab:netsres} provides the root mean square error (RMSE) between the projected probabilities and the ground truth probabilities for all the inferred query variables for $N_{ins}$ = 10, 50, 100, together with the RMSE predicted by taking the square root of the average variances from the inferred marginal beta distributions. Figure \ref{fig:netsres} plots the desired and actual significance levels for the confidence intervals (best closest to the diagonal). Figure \ref{fig:nettimes} depicts the distribution of execution time for running the various algorithms, and Figure \ref{fig:netsrescorr} the correlation of the Dirichlet strength between the golden standard, i.e. a Monte Carlo simulation with 10,000 samples, and both \lbetaprobcov\ and \lMonteCarlo\ varying the number of samples.

Table \ref{tab:netsres} shows that \lbetaprobcov\ shares the best performance with the state-of-the-art \lSBN\ and \lHonolulu\ almost constantly. This is clearly a significant achievement considering that \lSBN\ is the state-of-the-art approach when dealing only with single connected Bayesian Networks with uncertain probabilities, while we can also handle much more complex problems. 
Consistently with Table \ref{tab:smokers}, and also with \cite{DBLP:conf/aaai/CeruttiKKS19}, \lHonolulu\ has lower RMSE than \lJosang\ and it 
seems that \lHonolulu\ overestimates the predicted RMSE and \lJosang\ underestimates it as \lJosang\ predicts smaller error than is realised and vice versa for \lHonolulu.

From visual inspection of Figure \ref{fig:netsres}, it is evident that \lbetaprobcov, \lSBN, and \lMonteCarlo\ all are very close to the diagonal, thus correctly assessing their own epistemic uncertainty. \lHonolulu\ performance is heavily affected by the fact that it computes the conditional distributions at the very end of the process and it relies, in \eqref{eq:variance-division}, on the assumption of independence. \lbetaprobcov, keeping track of the covariance between the various nodes in the circuits, does not suffer from this problem.
This positive result has been achieved without substantial deterioration of the performance in terms of execution time, as displayed in Figure \ref{fig:nettimes}, for which the same commentary of Figure \ref{fig:sftimes} applies. 

Finally, Figure \ref{fig:netsrescorr} depicts the correlation of the Dirichlet strength between the golden standard, i.e. a Monte Carlo simulation with 10,000 samples, and both \lbetaprobcov\ and \lMonteCarlo, this last one varying the number of samples used. Like for Figure \ref{fig:sfcorr}, it is straightforward to see that \lMonteCarlo\ improves the accuracy of its computed epistemic uncertainty when increasing the number of samples considered, approaching the same level of \lbetaprobcov\ when considering more than 200 samples, while \lbetaprobcov\ performs very closely to the optimal value of 1.

\section{Conclusion}
In this paper, we introduce (Section \ref{sec:contribution}) an algorithm for reasoning over a probabilistic circuit whose leaves are labelled with beta-distributed random variables, with the additional piece of information describing which of those are actually independent (Section \ref{sec:preprocessing}). This provides the input to an algorithm that \emph{shadows} the circuit derived for computing the probability of %a set of 
the {\em pieces of evidence} by superimposing a second circuit modified for computing the probability of a given {\em query and %of the set of 
the pieces of evidence}, thus having all the necessary components for computing the probability of a query conditioned %a set of 
on the pieces of evidence (Section \ref{sec:shadowing}). This is essential when evaluating such a shadowed circuit (Section \ref{sec:mainalgorithm}), with the  covariance matrix playing an essential role by keeping track of the dependencies between random variables while they are manipulated within the circuit. We also include discussions on memory management in Section \ref{sec:memory-performance}.

In our extensive experimental analysis (Section \ref{sec:experiment}) we compare   against leading approaches to compute %ing 
uncertain probabilities, notably: (1) Monte Carlo sampling; (2) our previous proposal \cite{DBLP:conf/aaai/CeruttiKKS19} as representative of the family of approaches using a moment matching approach with strong independence assumptions; (3) Subjective Logic \cite{Josang2016-SL}; (4) Subjective Bayesian Network (SBN) \cite{ivanovska.15,kaplan.16.fusion,KAPLAN2018132}; (5) Dempster-Shafer Theory of Evidence \cite{DEMPSTER68,Smets2005}; and (6) credal networks \cite{credal98}. 

We achieve the same or better results of state-of-the-art approaches for dealing with epistemic uncertainty, including highly engineered ones for a narrow domain such as SBN, while being able to handle general probabilistic circuits and with just a modest increase in the computational effort. In fact, this work has inspired us to leverage probabilistic circuits to expand second-order inference for SBN for arbitrary directed acyclic graphs whose variables are multinomials. In work soon to be released~\cite{kaplan2020sbn}, we prove the mathematical equivalence of the updated SBN inference approach to that of \cite{VANALLEN2008483}, but with significantly lower computational burden.  

We focused our attention on probabilistic circuits derived from  \dDNNF s:  work by \cite{Darwiche2011}, and then also by \cite{Kisa2014} has introduced Sentential Decision Diagrams (SDDs) as a new canonical formalism respectively for propositional and for probabilistic circuits. However, as we can read in \cite[p. 819]{Darwiche2011} SDDs is a strict subset of \dDNNF, which is thus the least constrained %t 
type of propositional circuit we can safely rely on according to \cite[Theorem 4]{kimmig2017-algebraic}. However, in future work we will enable our approach to efficiently make use of SDDs. 

In addition, we will also work in the direction of enabling learning with partial observations---incomplete data where the instantiations of each of the propositional variables are not always visible over all training instantiations---on top of its ability of tracking the covariance values between the various random variables for a better estimation of epistemic uncertainty.

\section*{Acknowledgement}

This research was sponsored by the U.S. Army Research Laboratory and the U.K. Ministry of Defence under Agreement Number W911NF-16-3-0001. The views and conclusions contained in this document are those of the authors and should not be interpreted as representing the official policies, either expressed or implied, of the U.S. Army Research Laboratory, the U.S. Government, the U.K. Ministry of Defence or the U.K. Government. The U.S. and U.K. Governments are authorized to reproduce and distribute reprints for Government purposes notwithstanding any copyright notation hereon.

\bibliography{biblio}
\bibliographystyle{plain}

\appendix

\section{aProbLog}
\label{sec:aproblog}

In the last years, several probabilistic variants of Prolog
have been developed, such as 
ICL~\cite{Poole00}, Dyna~\cite{Eisner05}, PRISM~\cite{Sato01} and ProbLog~\cite{deraedt:ijcai07}, with its aProbLog extension \cite{DBLP:conf/aaai/KimmigBR11} to handle arbitrary labels from a semiring.
They all are based on
definite clause logic (pure Prolog) extended with facts labelled with  probability values.
Their meaning is typically derived from
Sato's distribution semantics~\cite{sato:iclp95}, which assigns a probability to every literal. The probability of a Herbrand interpretation, or possible world, is the product of the probabilities of the literals occurring in this world.
The success probability is the probability that a query succeeds in a randomly selected world.

For a set $J$ of ground facts, we define the set of literals $\operatorname{L}(J)$ and the set of interpretations $\mathcal{I}(J)$ as follows:
\begin{align}
\operatorname{L}(J) &= J\cup\{\neg f~|~f\in J \}\\
\mathcal{I}(J) &= \{S~|~S\subseteq \operatorname{L}(J) \wedge \forall l\in J:~l\in S\leftrightarrow \neg l \notin S\}
\end{align}

An algebraic Prolog (aProbLog) program \cite{DBLP:conf/aaai/KimmigBR11} consists of:
\begin{itemize}
\item a \emph{commutative semiring} $\tuple{\mathcal{A},\oplus,\otimes, e^{\oplus},e^{\otimes}}$
\item a finite set of ground \emph{algebraic facts} $\operatorname{F} = \{ f_1, \ldots, f_n\}$
\item a finite set $\operatorname{BK}$ of \emph{background knowledge clauses}
\item a \emph{labeling function} $\rho : \operatorname{L}(\operatorname{F})\rightarrow \mathcal{A}$
\end{itemize}
Background knowledge clauses are definite clauses, but their bodies may contain negative literals for algebraic facts. Their heads may not unify with any algebraic fact.

For instance, in the following aProbLog program
\begin{verbatim}
alarm :- burglary.
0.05 :: burglary.
\end{verbatim}

\noindent
\verb+burglary+ is an algebraic fact with label \verb+0.05+, and \verb+alarm :- burglary+ represents a background knowledge clause, whose intuitive meaning is: in case of burglary, the alarm should go off.

The idea of splitting a logic program in a set of facts and a set of clauses goes back to Sato's distribution semantics~\cite{sato:iclp95}, where it is used to define a probability distribution over interpretations of the entire program in terms of a distribution over the facts.
This is possible because a truth value assignment to the facts in $\operatorname{F}$ uniquely determines the truth values of all other atoms defined in the background knowledge.
In the simplest case, as realised in ProbLog \cite{deraedt:ijcai07,Fierens2015}, this basic distribution considers facts to be independent random variables and thus multiplies their individual probabilities. aProbLog uses the same basic idea, but generalises from the semiring of probabilities to general commutative semirings.
While the distribution semantics is defined for countably infinite sets of facts,
the set of ground algebraic facts in aProbLog must be finite.

In aProbLog, the label of
a complete interpretation $I\in \mathcal{I}(\operatorname{F})$  is defined as the product of the labels of its literals
\begin{equation}
\operatorname{\mathbf{A}}(I)  =  \bigotimes_{l\in I}\rho(l) \label{eq:w_expl}
\end{equation}
and the label of a set of interpretations $S\subseteq \mathcal{I}(\operatorname{F})$ as the sum of the interpretation labels
\begin{equation}
\operatorname{\mathbf{A}}(S)  = \bigoplus_{I\in S}\bigotimes_{l\in I}\rho(l)
\end{equation}
A \emph{query} $q$ is a finite set of algebraic literals and atoms from the Herbrand base,\footnote{I.e., the set of ground atoms that can be constructed from the predicate, functor and constant symbols of the program.} $q\subseteq \operatorname{L}(\operatorname{F}) \cup HB(\operatorname{F}\cup \operatorname{BK})$. We denote the set of interpretations where the query is true by $\mathcal{I}(q)$,
\begin{equation}
\mathcal{I}(q) = \{I~|~I\in \mathcal{I}(\operatorname{F}) \wedge I\cup \operatorname{BK}\models q\}
\end{equation}
The label of query $q$ is defined as the label of $\mathcal{I}(q)$,
\begin{equation}
\operatorname{\mathbf{A}}(q)  =  \operatorname{\mathbf{A}}(\mathcal{I}(q)) = \bigoplus_{I\in \mathcal{I}(q)}\bigotimes_{l\in I}\rho(l).\label{eq:q_int}
\end{equation}
As both operators are commutative and associative, the label is independent of the order of both literals and interpretations.

ProbLog \cite{Fierens2015} is an instance of aProbLog with
\begin{equation}
    \begin{array}{l}
         \mathcal{A} = \mathbb{R}_{\geq 0};\\
         a ~\oplus~ b = a + b;\\
         a ~\otimes~ b = a \cdot b;\\
         e^\oplus = 0;\\
         e^{\otimes} = 1;\\
         \delta(f) \in [0,1];\\
         \delta(\lnot f) = 1 - \delta(f)
    \end{array}
\end{equation}

\section{Subjective Logic Operators of Sum, Multiplication, and Division}
\label{sec:sl-operators}
  Let us recall the following operators as defined in \cite{Josang2016-SL}.
In the following, let $\slop{X} = \sloptuple{X}$ and $\slop{Y} = \sloptuple{Y}$ be two subjective logic opinions.

\subsection{Sum}
The opinion about $X \cup Y$ (\textbf{sum}, $\slop{X} \boxplus_{\text{SL}} \slop{Y}$) is defined as $\slop{X \cup Y} = \sloptuple{X \cup Y}$, where:
\begin{itemize}
    \item $\slbel{X \cup Y} = \slbel{X} + \slbel{Y}$;
    \item $\sldis{X \cup Y} = \frac{\slbase{X} (\sldis{X}-\slbel{Y}) + \slbase{Y} (\sldis{Y} - \slbel{X})}{\slbase{X} + \slbase{Y}}$;
    \item $\slunc{X \cup Y} = \frac{\slbase{X} \slunc{X} + \slbase{Y} \slunc{Y}}{\slbase{X} + \slbase{Y}}$; and
    \item $\slbase{X \cup Y} = \slbase{X} + \slbase{Y}$.
\end{itemize}

\subsection{Product}
The opinion about $X \land Y$ (\textbf{product}, $\slop{X} \boxtimes_{\text{SL}} \slop{Y}$) is defined---under assumption of independence---as $\slop{X \land Y} = \sloptuple{X\land Y}$, where:
\begin{itemize}
    \item $\slbel{X \land Y} = \slbel{X} \slbel{Y} + \frac{(1 - \slbase{X})\slbase{Y} \slbel{X} \slunc{Y} +\slbase{X}(1 - \slbase{Y})\slunc{X}\slbel{Y}}{1 - \slbase{X}\slbase{Y}}$;
    \item $\sldis{X \land Y} = \sldis{X}+{\sldis{Y}} - \sldis{X}\sldis{Y}$;
    \item $\slunc{X \land Y} = \slunc{X}\slunc{Y}+\frac{(1 - \slbase{Y})\slbel{X}\slunc{Y}+(1 - \slbase{X})\slunc{X}\slbel{Y}}{1 - \slbase{X}\slbase{Y}}$; and 
    \item $\slbase{X \land Y} = \slbase{X} \slbase{Y}$.
\end{itemize}

\subsection{Division}
The opinion about the division of $X$ by $Y$, $X \widetilde{\land} Y$ (\textbf{division}, $\slop{X} \boxslash_{\text{SL}} \slop{Y}$) is defined as $\slop{X \widetilde{\land} Y} = \sloptuple{X \widetilde{\land} Y}$ where
\begin{itemize}
    \item $\slbel{X \widetilde{\land} Y}$ = 
             $\frac{\slbase{Y}(\slbel{X}+\slbase{X}\slunc{X})}{(\slbase{Y}-\slbase{X})(\slbel{Y}+\slbase{Y}\slunc{Y})}
             -
             \frac{\slbase{X}(1-\sldis{X})}{(\slbase{Y}-\slbase{X})(1-\sldis{Y})}
             $;
    \item $\sldis{X \widetilde{\land} Y} = \frac{\sldis{X}-\sldis{Y}}{1-\sldis{Y}}$;
    \item $\slunc{X \widetilde{\land} Y} = 
             \frac{\slbase{Y}(1-\sldis{X})}{(\slbase{Y}-\slbase{X})(1-\sldis{Y})}
             -
             \frac{\slbase{Y}(\slbel{X}+\slbase{X}\slunc{X})}{(\slbase{Y}-\slbase{X})(\slbel{Y}+\slbase{Y}\slunc{Y})}
             $; and
    \item $\slbase{X \widetilde{\land} Y} = \frac{\slbase{X}}{\slbase{Y}}$
\end{itemize}

\noindent
subject to: 
\begin{itemize}
    \item $\slbase{X} < \slbase{Y}$; $\sldis{X} \geq \sldis{Y}$; 
    \item $\slbel{X} \geq \frac{\slbase{X}(1-\slbase{Y})(1-\sldis{X})\slbel{Y}}{(1-\slbase{X})\slbase{Y}(1-\sldis{Y})}$; and 
    \item $\slunc{X} \geq \frac{(1-\slbase{Y})(1-\sldis{X})\slunc{Y}}{(1-\slbase{X})(1-\sldis{Y})}$.
\end{itemize}

\section{Independence of posterior distributions when learning from complete observations}
\label{sec:independence-posterior}

Let us instantiate AMC using probabilities as labels (cf. \eqref{eq:semiringprobability}) and let us consider a propositional logic theory over $M$ variables. We can thus re-write \eqref{eq:amc} as:

\begin{equation}
    p(T) = \sum_{I \in \mathcal{M}(T)} \prod_{m = 1}^M p(l_m)
\end{equation}
Hence, the probability of a theory is function of the probabilities of interpretations $p(I \in \mathcal{M}(T))$, where 
\begin{equation} \label{eq:ad1}
    p(I \in \mathcal{M}(T)) = \prod_{m=1}^M p(l_m)
\end{equation}

Let's assume that we want to learn such probabilities from a dataset $\Data = \vt{\vp{\vvx_1, \ldots, \vvx_N}}$, then by (\ref{eq:ad1}) the variables for which we are learning probabilities are independent, hence
%and let's assume independence between the variables for which we are learning probabilities, hence 
\begin{equation}
    p(l_1, \ldots, l_M) = \prod_{m=1}^M p(l_m)
\end{equation}
We can thus re-write the likelihood \eqref{eq:vec-likelihood} as:
\begin{equation}
    \def\arraystretch{2}
    \begin{array}{r c l}
    p(\Data \mid \bm{p_x}) & = &\displaystyle{\prod_{i=1}^{|\Data|} p(\bm{x}_i | \bm{p}_{x_i})}\\
    & = & \displaystyle{\prod_{i=1}^{|\Data|} \prod_{m=1}^M p_{x_m}^{x_{i,m}} (1 - p_{x_m})^{1 - x_{i,m}}}
    \end{array}
\end{equation}
Assuming a uniform prior, and letting $r_m$ be the number of observations for $x_m = 1$ and $s_m$ the number of observations for $x_m = 0$, 
we can thus compute the posterior as:
\begin{equation}
\def\arraystretch{2}
    \begin{array}{r c l}
         p(\bm{p_x} \mid \Data, \bm{\alpha}^0) & \propto & p(\Data \mid \bm{p_x}) \cdot p(\bm{p_x} \mid \bm{\alpha}^0) \\
         & \propto & \displaystyle{\prod_{m=1}^M p_{x_m}^{r_m + \alpha^0_{x_m} - 1} (1 - p_{x_m})^{s_m + \alpha^0_{\overline{x}_m} - 1} }
    \end{array}
\end{equation}
which, in turns, show that the independence is maintained also considering the posterior beta distributions.

\section{Bayesian networks derived from aProbLog programs}
\label{sec:bn}

\begin{figure*}[t]
\centering
    \begin{subfigure}[b]{0.25\textwidth}
        \includegraphics[width=\textwidth]{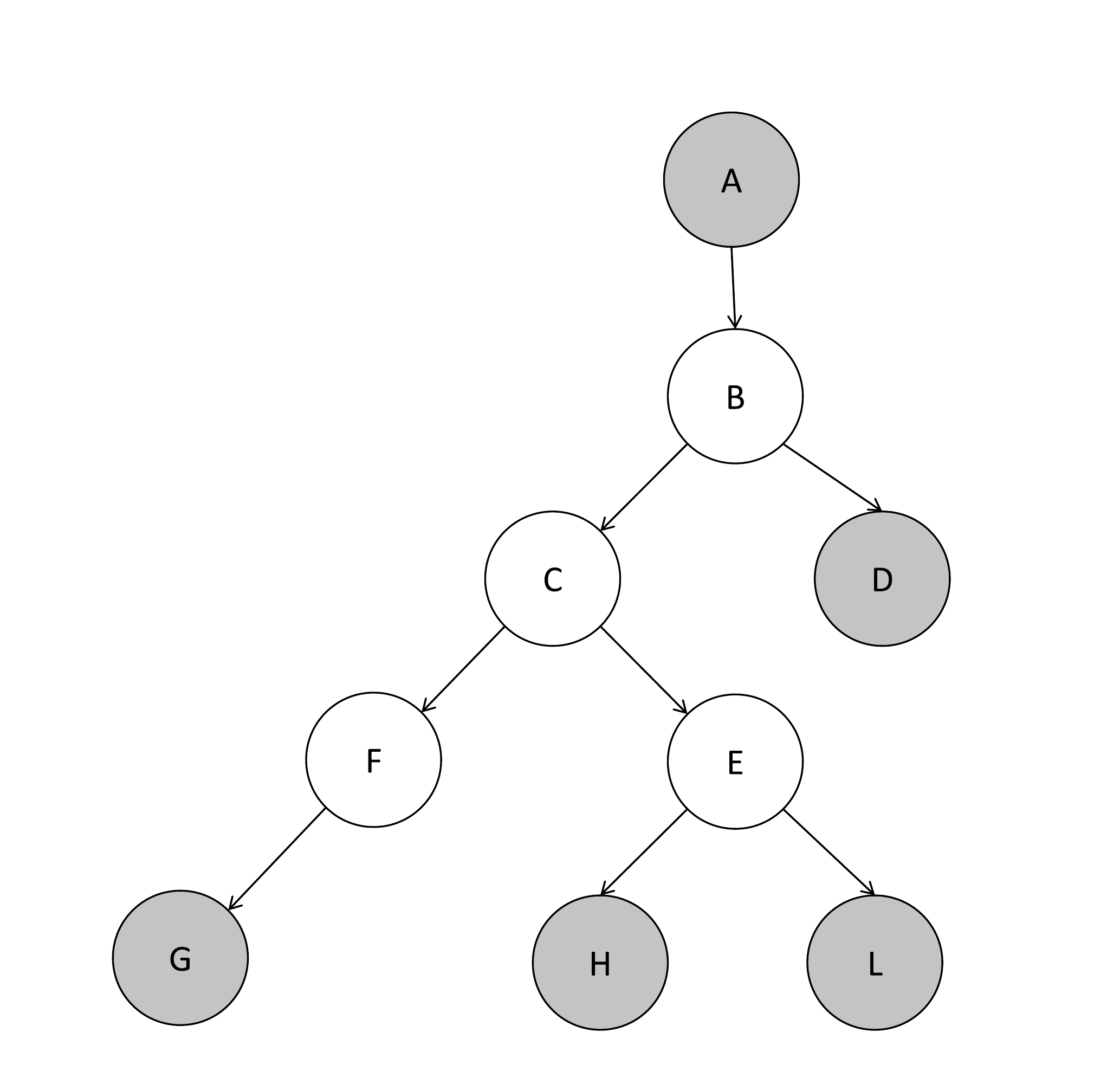}
        \caption{}
        \label{fig:net1}
    \end{subfigure}
    ~
    \begin{subfigure}[b]{0.25\textwidth}
        \includegraphics[width=\textwidth]{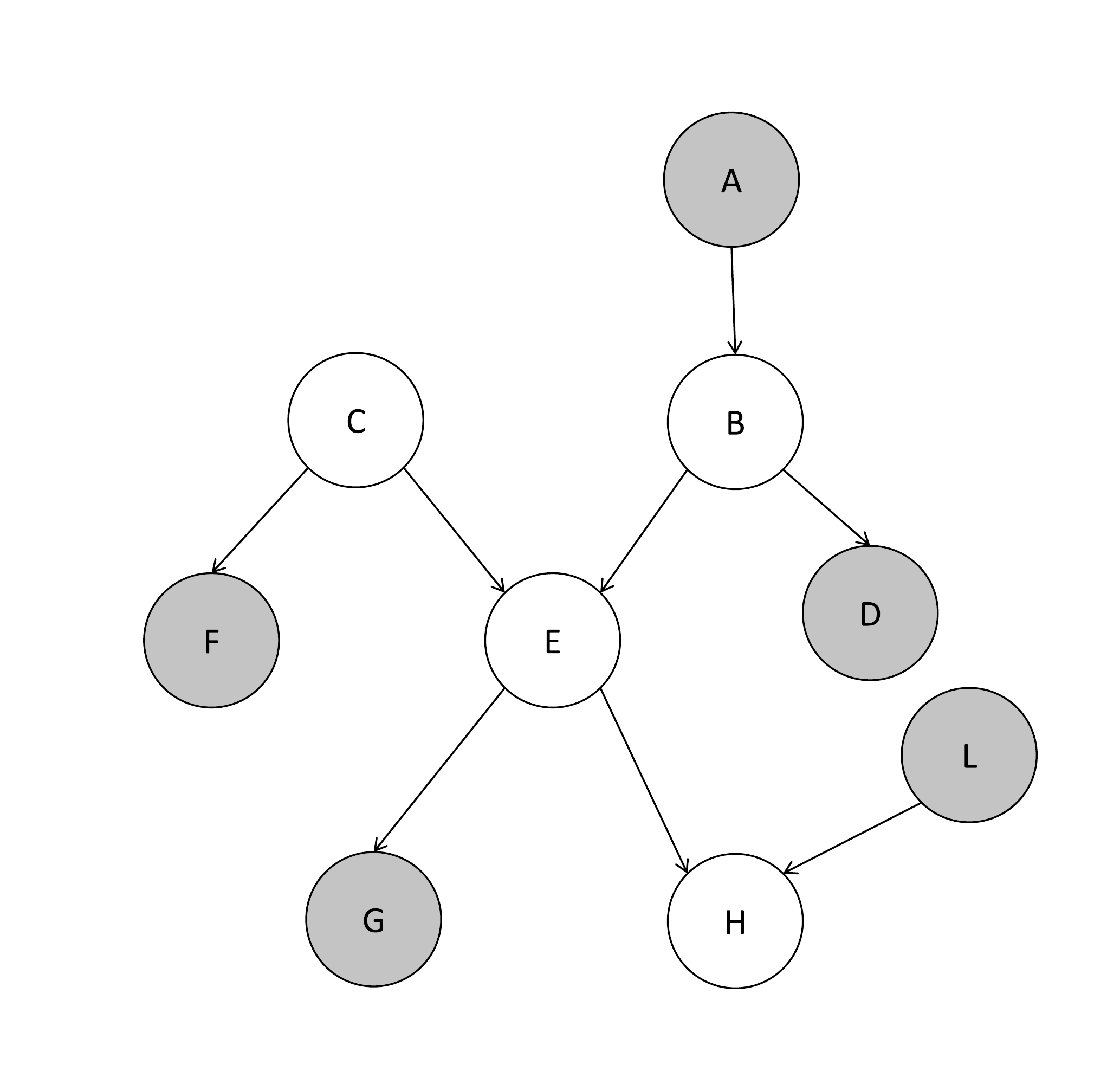}
        \caption{}
        \label{fig:net2}
    \end{subfigure}
    ~
    \begin{subfigure}[b]{0.25\textwidth}
        \includegraphics[width=\textwidth]{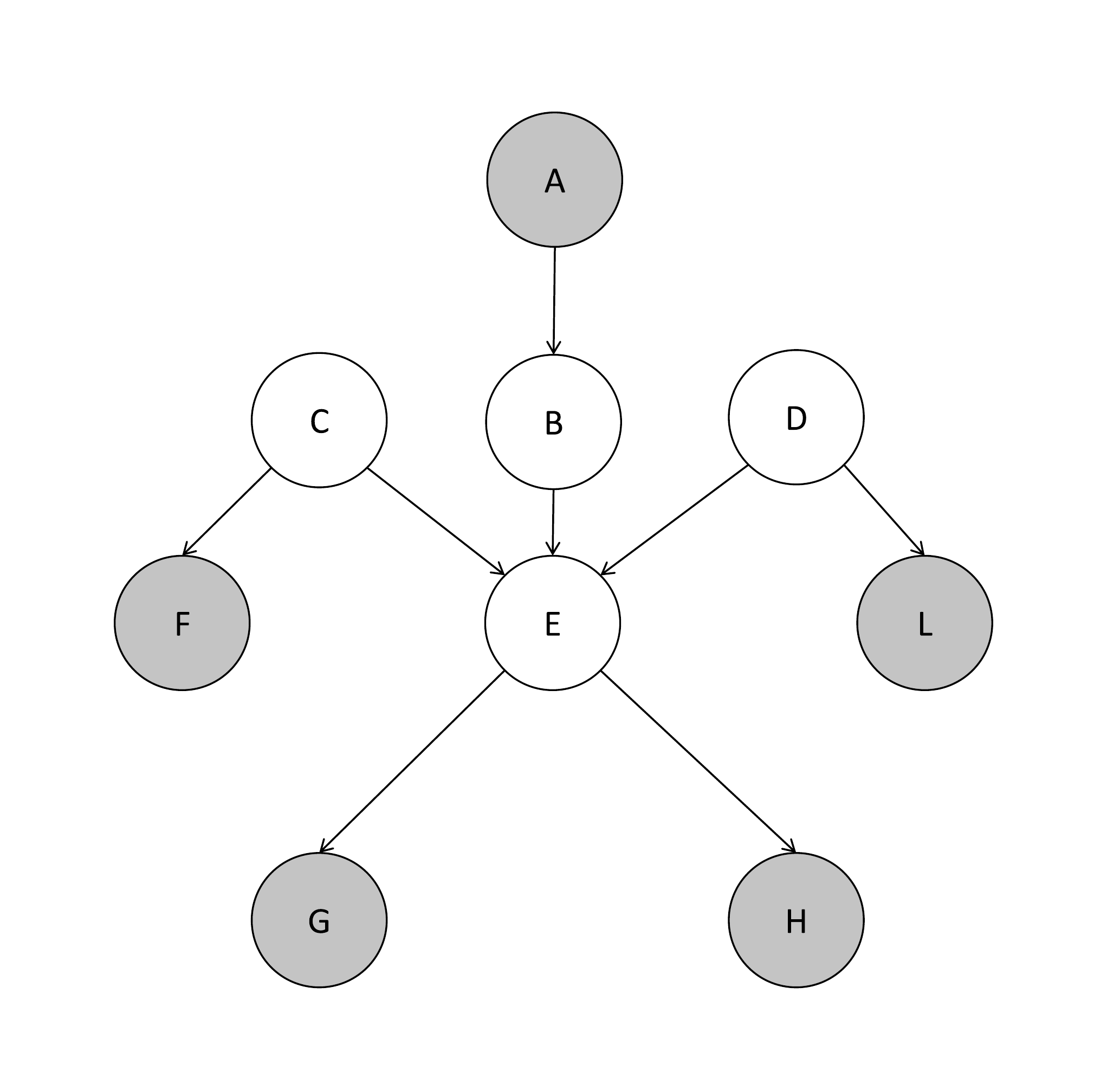}
        \caption{}
        \label{fig:net3}
    \end{subfigure}
    \caption{Network structures tested where the exterior gray variables are directly observed and the remaining are queried: (\subref{fig:net1}) Net1, a tree; (\subref{fig:net2}) Net2, singly connected network with one node having two parents; (\subref{fig:net3}) Net3, singly connected network with one node having three parents.}
    \label{fig:nets}
\end{figure*}

Figure \ref{fig:nets} depicts the Bayesian networks that can be derived from the three circuits considered in the experiments described in Section \ref{sec:bnexp}.

\end{document}